\documentclass[aoas]{imsart}

\RequirePackage{amsthm}
\RequirePackage{amsmath}
\RequirePackage{amsfonts}
\RequirePackage{amssymb} 
\RequirePackage[authoryear]{natbib}
\RequirePackage[colorlinks,citecolor=blue,urlcolor=blue]{hyperref}
\RequirePackage{graphicx}
\usepackage{ulem}
\startlocaldefs
\theoremstyle{plain}

\newtheorem{theorem}{Theorem}[section]
\newtheorem{lemma}[theorem]{Lemma}
\theoremstyle{remark}

\usepackage{psfrag,epsf}

\newcommand{\blind}{0}

\usepackage{color}

\newif\ifworkinprogress
\workinprogresstrue

\ifworkinprogress

\else
  \newcommand{\mq}[1]{}
  \newcommand{\es}[1]{}
  \newcommand{\mr}[1]{}
  \newcommand{\ty}[1]{}
  \newcommand{\pv}[1]{}
\fi

\usepackage{epsfig}

\usepackage{balance}  
\usepackage{algorithmic}
\usepackage{algorithm}
\usepackage{multirow}
\usepackage{subcaption}
\usepackage[font=small]{caption}
\usepackage{multicol}
\usepackage{enumerate}
\usepackage{booktabs}
\usepackage{bbold}
\usepackage{wrapfig}
\usepackage{paralist}

\newcommand{\B}{\boldsymbol}
\newcommand{\M}{\mathbf}
\newcommand{\bff}{{\bf f}}
\newcommand{\bY}{{\bf y}}
\newcommand{\bepsilon}{\boldsymbol \epsilon}
\newcommand{\bv}{{\bf v}}
\newcommand{\bh}{{\bf h}}

\newtheorem{corollary}{Corollary}

\DeclareMathOperator*{\argmin}{argmin}

\usepackage{comment, tcolorbox, changepage, dsfont}
\usepackage{pifont}

\newcommand{\norm}[1]{\left\lVert#1\right\rVert}
\newcommand\eqdef{\mathrel{\overset{\makebox[0pt]{\mbox{\normalfont\tiny\sffamily def}}}{=}}}

\usepackage{outlines}
\usepackage{etoc}

\endlocaldefs

\newcommand{\modelaname}{\texttt{ELAAN}}

\newcommand{\modelbname}{\texttt{ELAAN-I}}

\newcommand{\modelcname}{\texttt{ELAAN-H}}

\begin{document}

\begin{frontmatter}
\title{Predicting Census Survey Response Rates 
with Parsimonious Additive Models and Structured Interactions}
\runtitle{Predicting Census Survey Response Rates}

\begin{aug}
\author[A]{\fnms{Shibal} \snm{Ibrahim}\ead[label=e1,mark]{shibal@mit.edu}},
\author[B]{\fnms{Peter} \snm{Radchenko}\ead[label=e3,mark]{peter.radchenko@sydney.edu.au}},
\author[C]{\fnms{Emanuel} \snm{Ben-David}\ead[label=e4,mark]{emanuel.ben.david@census.gov}},
\and
\author[D]{\fnms{Rahul} \snm{Mazumder}\ead[label=e2,mark]{rahulmaz@mit.edu}}
\address[A]{Department of Electrical Engineering and Computer Science, Massachusetts Institute of Technology,
\printead{e1}}
\address[B]{University of Sydney,
\printead{e3}}
\address[C]{United States Census Bureau,
\printead{e4}}
\address[D]{MIT Sloan School of Management, Massachusetts Institute of Technology,
\printead{e2}}
\end{aug}


\if1\blind
{
    \author{
Shibal Ibrahim\thanks{email: {\texttt{shibal@mit.edu}}, Massachusetts Institute of Technology},~
    Rahul Mazumder\thanks{email: {\texttt{rahulmaz@mit.edu}},~Massachusetts Institute of Technology},~
    Peter Radchenko\thanks{email: {\texttt{peter.radchenko@sydney.edu.au}},~University of Sydney},~ 
    Emanuel Ben-David\thanks{email: {\texttt{emanuel.ben.david@census.gov}},    United States Census Bureau}
    }
    \date{\vspace{-5ex}}  
    \maketitle
} \fi

\if0\blind
{
} \fi

\bigskip

\etocdepthtag.toc{mtchapter}
\etocsettagdepth{mtchapter}{subsection}
\etocsettagdepth{mtappendix}{none}

\begin{abstract}
In this paper, we consider the problem of predicting survey response rates using a family of flexible and interpretable nonparametric models. The study is motivated by the US Census Bureau's well-known ROAM application, which uses a linear regression model trained on the US Census Planning Database data to identify hard-to-survey areas.  A crowdsourcing competition~\citep{Erdman2016} organized more than ten years ago revealed that machine learning methods based on ensembles of regression trees led to the best performance in predicting survey response rates; however, the corresponding models could not be adopted for the intended application due to their black-box nature. We consider nonparametric additive models with a small number of main and pairwise interaction effects using $\ell_0$-based penalization. From a methodological viewpoint, we study our estimator's computational and statistical aspects and discuss variants incorporating strong hierarchical interactions. Our algorithms (open-sourced on GitHub) extend the computational frontiers of existing algorithms for sparse additive models to be able to handle datasets relevant to the application we consider. We discuss and interpret findings from our model on the US Census Planning Database. In addition to being useful from an interpretability standpoint, our models lead to predictions 
comparable to popular black-box machine learning methods based on gradient boosting and feedforward neural networks --  suggesting that it is possible to have models that have the best of both worlds: good model accuracy and interpretability.
\end{abstract}



\end{frontmatter}

\section{Introduction}\label{sec:intro}
\begin{figure}[!t]
        \centering
        \includegraphics[width=0.8\textwidth]{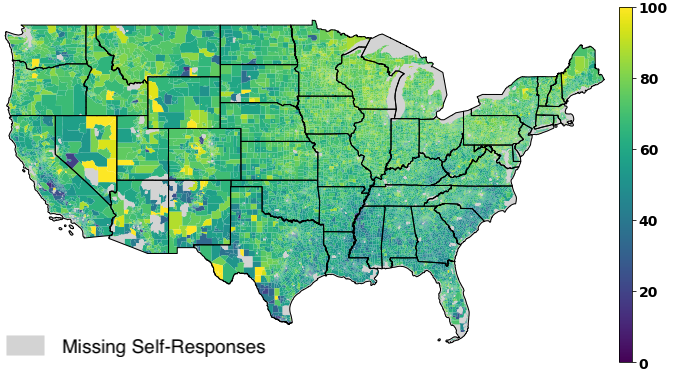}
        \caption{The 2013-2017 American  Community  Survey self-response rates for all tracts in the continental United States. The North, in general, and the Upper Midwest and the Northeast, in particular, have higher self-response rates than the rest of the country. Tracts with lower self-response rates are visible in many states -- in particular, in the South and in the Mountain region.}. 
        \label{fig:Maps-US-b-actual}
\end{figure}

Sample surveys and censuses are primary data sources in social science studies. However, low and often unpredictable response rates in surveys remain a continual source of concern~\citep{Erdman2016,tourangeau_hard--survey_2014,tourangeau2014hard} -- see Figure~\ref{fig:Maps-US-b-actual} for an illustration on the American Community Survey. 
\cite{tourangeau2014hard} discuss many factors that make parts of the population hard to survey -- such factors are often used to improve sampling strategies, questionnaire designs, recruitment methods, and the language in which the interview is conducted, among others.  
\cite{Erdman2016} emphasize the usefulness of having an indicator for hard-to-survey areas to
guide targeted surveying (including oversampling), staff recruitment strategies, and targeted nonresponse followup. For major campaigns such as the decennial US Census, this approach can help guide resource allocation for advertisements and building community partnership activities.
Eliciting responses from non-self-responding households through follow-up operations can be costly and time-consuming.  
The Census Bureau estimates that a single percentage increase in the self-response rate amounts to roughly~$85$ million dollars saved in personal follow-up costs \citep{census_rep08, Bates2010}.  Apart from the cost,  the quality of proxy enumerations and imputations is typically substantially lower than that of self-responses \citep{Mule2010}.

Both the UK Office for National Statistics 
and the US Census Bureau have created measures that help quantify the difficulty in gathering data across different geographic areas.
A report by~\cite{bruce2001hard} from the US Census Bureau introduced a hard-to-count (HTC) score for identifying difficult-to-enumerate segments of the population. The HTC score, based on 12 carefully chosen covariates (for example, housing variables and socio-demographic/economic indicators), has found its use in the planning for the 2010 Census and many other national surveys. The HTC score, however, has some limitations and was later improved to the low-response-score, or LRS \citep{Erdman2016}. The LRS plays a key role in the US Census Bureau's Response Outreach Area Mapper (ROAM) application\footnote{\url{https://www.census.gov/library/visualizations/2017/geo/roam.html}} to identify hard-to-survey areas. 
The LRS appears to have been partly motivated by the 2012 nationwide competition, organized by the Census Bureau in partnership with Kaggle on predicting the mail-return rates.
This competition aimed to solicit models highly predictive of the decennial census response rates, easily replicable, inherently interpretable, and consistent for use in the field at various geographic levels.
Even though the winning models from the competition had good predictive performance, they lacked {\it reproducibility}
  and {\it interpretability}.
 Some of the winning models included covariates that are not publicly available; in particular, they were not chosen from 
 within the US Census Bureau Planning Database.
 Additionally, 
 the winning models included covariates that were good predictors but not actionable\footnote{These include covariates such as the nearest neighboring block group return rates and margins of error for various estimates. Models that incorporate such features are not actionable.}. Interpretability suffered due to a multitude of factors. First, the winning models were based on complex ensemble methods (e.g., random forests and gradient boosting). Second, these methods employed many covariates (the best model had nearly 340), which hurt the model parsimony. 
After careful analysis,~\cite{Erdman2016} proposed a linear model based on 25 covariates -- LRS is the prediction from this model. While this model suffered loss in predictive accuracy compared to the black-box machine learning methods, it was highly interpretable and led to useful actionable insights. 
For example, Erdman and Bates highlight three different block groups (Columbia Heights, Trinidad, Anacostia) in the District of Columbia that have similar LRS predictions, but their 
characteristics vary, indicating a need for different actions to increase self-response. Columbia Heights has a large Hispanic population, suggesting they could benefit from forms and advertising in Spanish. Anacostia has a stable population (with a low relocation rate to the region) and could benefit from community partnerships. On the other hand, Trinidad is characterized as a region in transition with high mobility of the young population. This area could benefit from web-based advertisements for internet-based responses from the tech-savvy younger population.    
 
We aim to predict self-response rates across all Census tracts in the US using operational,  socio-economic, and demographic characteristics from the US Census Planning Database.
Our goal is to propose interpretable statistical models that deliver 
strong predictive performance 
and can potentially complement the currently used LRS metric.
Our hope is that these models lend operational insights into factors influencing survey self-response rates to 
facilitate cost-effectiveness and improved coverage. 

\noindent {\bf Statistical models.} We propose estimators based on Additive Models~\citep{hastie1987generalized,Hastie2001}, or AMs, with smooth nonlinear components that include nonlinear pairwise interactions between covariates. 
Given response $y \in \mathbb{R}$ and feature-vector $\M{x}:=(x_{1}, \ldots, x_{p}) \in \mathbb{R}^p$, we model the conditional mean function as
\begin{equation}\label{eq:GAMs with interactions}
    \mathbb{E}(y | \M{x})  = \sum_{j \in [p]} f_j(x_j) + \sum_{j<k} f_{j,k}(x_j, x_k),
\end{equation}
where~$f_{j}$
and~$f_{j,k}$ are unknown smooth univariate and bivariate functions, 
respectively.
Drawing inspiration from linear model settings~\citep{Bach2012}, we propose a new methodology to estimate the unknown functional components via an optimization problem with structural constraints arising from interpretability considerations. 
To obtain a sparse model with few predictors, that is, with many of the components $\{f_{j}\}$ and $\{f_{j,k}\}$ estimated as exactly zero, we present a novel $\ell_0$-regularized approach\footnote{For examples of $\ell_0$-based approaches and illustrations of their advantages over the $\ell_1$-based counterparts in the context of linear regression, see \cite{bertsimas2015best, mazumder2015discrete, mazumder2017subset,HazimehL0Learn,hazimeh2020sparse} and the references therein.}, penalizing the number of nonzero components in the model.
In addition, we explore a refined notion of interpretability in the context of interaction modeling -- namely, hierarchical sparsity -- inspired by its usage in linear models~\cite [for example,][]{Bien2013,HazimehHS}.
To our knowledge, we present a novel exploration of computational and statistical perspectives of an $\ell_0$ regularized approach for fitting sparse additive models with pairwise interactions.

\noindent{{\bf Applied contributions.}} 
Our empirical analysis of the US Census Bureau Planning Database demonstrates the promise of our proposed approach.
Our estimator improves the prediction of the self-response rates when compared to the linear regression methods discussed in the US Census Bureau report.
The improvements are due to \textit{both} the departure from linearity and the presence of nonlinear interactions. Importantly, the prediction accuracy of our flexible models appears to be at par with (or slightly better than) black-box machine learning methods such as neural networks and gradient-boosted decision trees, which topped the nationwide Kaggle competition organized by the Census Bureau.

Our framework results in simple models and allows for automated variable selection.
Our models are substantially more compact than many competing benchmarks -- we use $8$-$20$ times fewer interaction effects than the corresponding linear models and about half as many covariates as off-the-shelf machine learning methods (e.g., gradient boosting, feedforward neural networks, or explainable boosting machines) or the sparse {\textit{linear}} models with interaction effects. 
Our models also use fewer covariates (by about 30\%) compared to sparse nonparametric additive models (without interactions)---see Section~\ref{sec:expts-case-study} for details. We use our models to gather useful operational insights into the factors that influence response rates in surveys across different segments of the population.
In particular, our models automatically identify interactions between many of the key factors used in prior publications from the Census Bureau \citep{Bates2010} to derive meaningful population clustering. These interaction effects appear to result in improved predictions of survey response rates and 
complement earlier findings reported in~\cite{Bates2010, 2020ICC} on understanding different population segments.   

\noindent {\bf Methodological contributions.} 
We propose a new family of estimators based on nonparametric additive models with sparse interactions under combinatorial constraints.
We establish statistical guarantees for the resulting estimators and present large-scale algorithms for these estimators
extending the current computational landscape for sparse nonparametric additive models with pairwise interactions. Our approach addresses some of the key computational challenges posed by the large-scale setting of the Census dataset, with approx. $10^5$ nonparametric interaction components and approx. $10^5$ observations. The code implementing our algorithms is available at:\\ 
Code: \url{https://github.com/mazumder-lab/elaan}.

\noindent{{\bf Related Work.}} There is an impressive body of methodological and theoretical work on using convex $\ell_1$-based approaches to fit sparse nonparametric AMs without interactions~\cite[see, for example,][and the references therein]{Meier2009, Ravikumar2009,huang1,Zhao2012,yuan2016minimax}. 
Even with main-effects alone, these convex optimization-based methods face computational challenges for the problem-scales we seek to address\footnote{Based on our experience, currently available software (for example, R package \texttt{SAM}) encounters numerical difficulties for instances where~$n$ is on the order of thousands, and~$p$ is on the order of hundreds.}. This possibly limits practitioners from realizing the full potential of nonparametric AMs in large-scale settings arising in various practical applications.  
Regarding statistical properties, the $\ell_0$-based estimators can offer improvements over their $\ell_1$-counterparts, both on the prediction and the model selection fronts.
To this end, we refer the reader to recent work by~\cite{hazimeh2021group} demonstrating the advantages of using the $\ell_0$-regularized framework for 
grouped variable selection (which includes nonparametric AMs) \textit{without} interactions.

In the presence of pairwise interactions, a setting we focus on, the problem of variable selection in nonparametric AMs becomes considerably more challenging. 
The approaches by \citet{Meier2009} and \citet{Ravikumar2009}  consider additive models with main effects but no pairwise interactions. \citet{lin2} introduced COSSO, which is a well-known method to fit model~\eqref{eq:GAMs with interactions}; however, the method appears to be suitable for low-dimensional settings -- for example, the authors consider instances with $n\sim500$, $p\sim10$, and approximately $50$ pairwise interactions\footnote{There appears to be no open-source implementation for COSSO.}.

COSSO penalizes the sum of the Sobolev norms of the functional components in representation~\eqref{eq:GAMs with interactions} and, thus, can produce sparse models. However, the convex COSSO penalty jointly controls sparsity and smoothness, potentially resulting in unwanted shrinkage interfering with component selection. 
In contrast, our proposed approach 
uses separate penalties for smoothness and sparsity, and 
encourages sparsity \textit{directly} via an $\ell_0$-based penalty on the number of components in the model.

More recently, 
there have been approaches that extend the scope of classical nonparametric AMs~\citep{hastie1987generalized}. In 
a series of works~\citet{Lou2013,Nori2019,Yang2021} explore tree-based and neural-network-based approaches for fitting additive models with sparse interactions. 
\citet{Lou2013} propose a two-stage approach using shallow tree-like models for the main and interaction effects. In the first stage, they use gradient boosting to fit the main effects -- the boosting procedure cycles through all features in a round-robin fashion to fit {\textit{all}} main effects. In the second stage, they use a
scheme to select a small subset of pairwise interactions--these interaction effects are fitted with shallow tree-like models via gradient boosting.
\citet{Nori2019} provide an efficient implementation of the above approach as Explainable Boosting Machines (EBM) in the well-known \textit{interpretml} toolkit.
The EBM procedure is not based on optimizing a penalized estimation framework and, to our knowledge, no statistical guarantees for it are known.\\
In another line of work, \citet{Yang2021} proposed GAMI-Net, which uses neural networks to fit main and pairwise interaction effects. 
GAMI-Net uses a multi-stage approach:
(i) fit all main effects, where each main effect is modeled using a multi-layer perceptron (MLP);
(ii) select Top-\textit{k} interactions based on an interaction detection method  \citep{Lou2013} on the residuals;
(iii) fit the Top-\textit{k} interaction effects simultaneously, where each interaction effect is again modeled as  MLP;
(iv) fine-tune all selected main effects and interaction effects together. 
GAMI-Net also prunes some components in steps~(i) and~(iii) based on some ranking measure.
GAMI-Net uses screening methods prior to training to select a collection of main and interaction effects and, as such, does {\it{not}} jointly optimize the sparsity pattern {while} training. 

\citet{Zschech2022} compared various  
additive models and observed that EBM \citep{Nori2019} and GAMI-Net \citep{Yang2021} led to good performance. 
However, both models have some practical limitations. For example, EBM includes all main effects in the model (i.e., there is no feature selection).
Neural-based approaches, such as GAMI-Net, are quite computationally expensive. On the Census dataset that we consider, GAMI-Net takes 12 hours (using an 8-CPU machine) to compute one model; thus, using 1000 tuning parameters leads to a steep 500-day computational cost \footnote{This time is for the architecture that uses a 2-layered NN with 10 neurons for each main/interaction effect.}. 

To the best of our knowledge, our algorithmic framework for sparse nonparametric AMs with interactions is novel and is likely to be of independent interest from a computational methodology perspective.

\noindent {\bf Organization.} Section~\ref{sec: Statistical Methodology} presents the statistical models pursued in this paper. Section~\ref{sec:theory} investigates the statistical properties of our proposed estimators. Section~\ref{sec: Efficient Computations at scale} discusses how to obtain solutions to the corresponding large-scale discrete optimization problems. Section~\ref{sec:nonparm-int-SH} develops an extension of our method that incorporates strong hierarchy constraints. In Section~\ref{sec:simulation-case-study}, we present simulation studies comparing our methods with COSSO, EBM and GAMI-Net. Section~\ref{sec:expts-case-study} discusses numerical results for the US Census Bureau application that motivates this study. 
Additional technical details are provided in the supplementary material \citep{supplement}.

\section{Statistical Models and Methodology} \label{sec: Statistical Methodology}
We now discuss the statistical models we pursue in this work. Section~\ref{sec:additive-nln-pwise-int} gives an overview of AMs with nonlinear main effects and pairwise interactions, along with optimization formulations associated with the estimation procedures.
Section~\ref{sec: Parsimonious interactions} presents our proposed models to incorporate sparsity in the main and interaction effects. 

\subsection{Smooth additive models with pairwise interactions}\label{sec:additive-nln-pwise-int}
Given data $\{(y_{i}, \M{x}_{i})\}_{1}^{n}$, our key objective is to learn a multivariate conditional mean function $f(\M{x}):=\mathbb{E}(y|\M{x})$, where $f:\mathbb{R}^p \mapsto \mathbb{R}$ is an unknown smooth function~\citep{Wahba1990}. It is well known that such functions become difficult to estimate even for moderately high~$p$ due to the curse of 
dimensionality -- therefore, we will focus on a smaller class of models corresponding to additive structures~\citep{hastie1987generalized,stone1986dimensionality}. A popular approach considered in the literature estimates a nonparametric additive model containing main effects only, with $f(\M{x}) = \sum_{j=1}^{p} f_j(x_j)$,
where each~$f_{j}$ is an unknown univariate smooth function of the $j$-th coordinate in~$\M{x}$, namely~$x_{j}$. 
In various applications, however, nonparametric AMs based on main effects alone may not lead to a sufficiently rich representation for predicting the outcome of interest: including interaction terms can lead to better predictive models while remaining interpretable to a practitioner~\citep{hastie1987generalized}. 
AMs with pairwise interactions are useful in applied statistical modeling with various applications in medical sciences and healthcare, e-commerce applications, recommender system problems, and sentiment analysis, among others~\citep{Hastie2015}.

\noindent {\bf The need for interactions.} Exploratory analysis on US Census Planning Database Tract dataset suggests that interaction effects across features can lead to improved predictions of the survey self-response rate. 
For example, there is interaction between the percentages of ``People who do not speak English well'' and ``Renters'' in the area. In areas with a relatively high concentration of poor English speakers (e.g., $\geq 5.4\%$), the self-response rate decreases on an average by~$0.33\%$ for a unit increase in the percentage of renters. On the other hand, when the concentration of poor English speakers is relatively low (e.g., $\leq 1.6\%$), the self-response rate decreases at the rate of~$0.21\%$. Similarly, there is a strong interaction effect between covariates ``Single-unit households'' and ``Household moved in 2010 or later'' in predicting the low self-response rate.
Indeed,~\citet{Erdman2016} note the importance of incorporating interaction effects when predicting the self-response rate, although the paper does not pursue statistical modeling involving interactions. 

An additive model with nonlinear main effects and pairwise interactions extends the traditional AM framework with main effects alone~\citep{Hastie2001, Radchenko2010}, and is given by model~\eqref{eq:GAMs with interactions},
where the unknown components $\{f_j\}$ and $\{f_{j,k}\}$ need to be estimated from the data. This leads to two key challenges. The presence of 
$O(p^2)$-many unknown nonparametric functions results in statistical challenges even for a moderate value of~$p$. Additional regularization (for example, in the form of sparsity in the components) may be necessary to obtain a reliable statistical model with good generalization properties (cf Section~\ref{sec: Parsimonious interactions}). Furthermore, as we mention in Section~\ref{sec:intro}, estimating model~\eqref{eq:GAMs with interactions} leads to severe computational challenges for large problems (for example, those with~$n \approx 10^5$ and~$p^2 \approx 10^5$, similar to the instances we consider in our applications) -- we discuss how we address these challenges in Section~\ref{sec: Efficient Computations at scale}.

\begin{figure}[!tb]
\begin{tabular}{ccc}
      \includegraphics[width=0.3\textwidth]{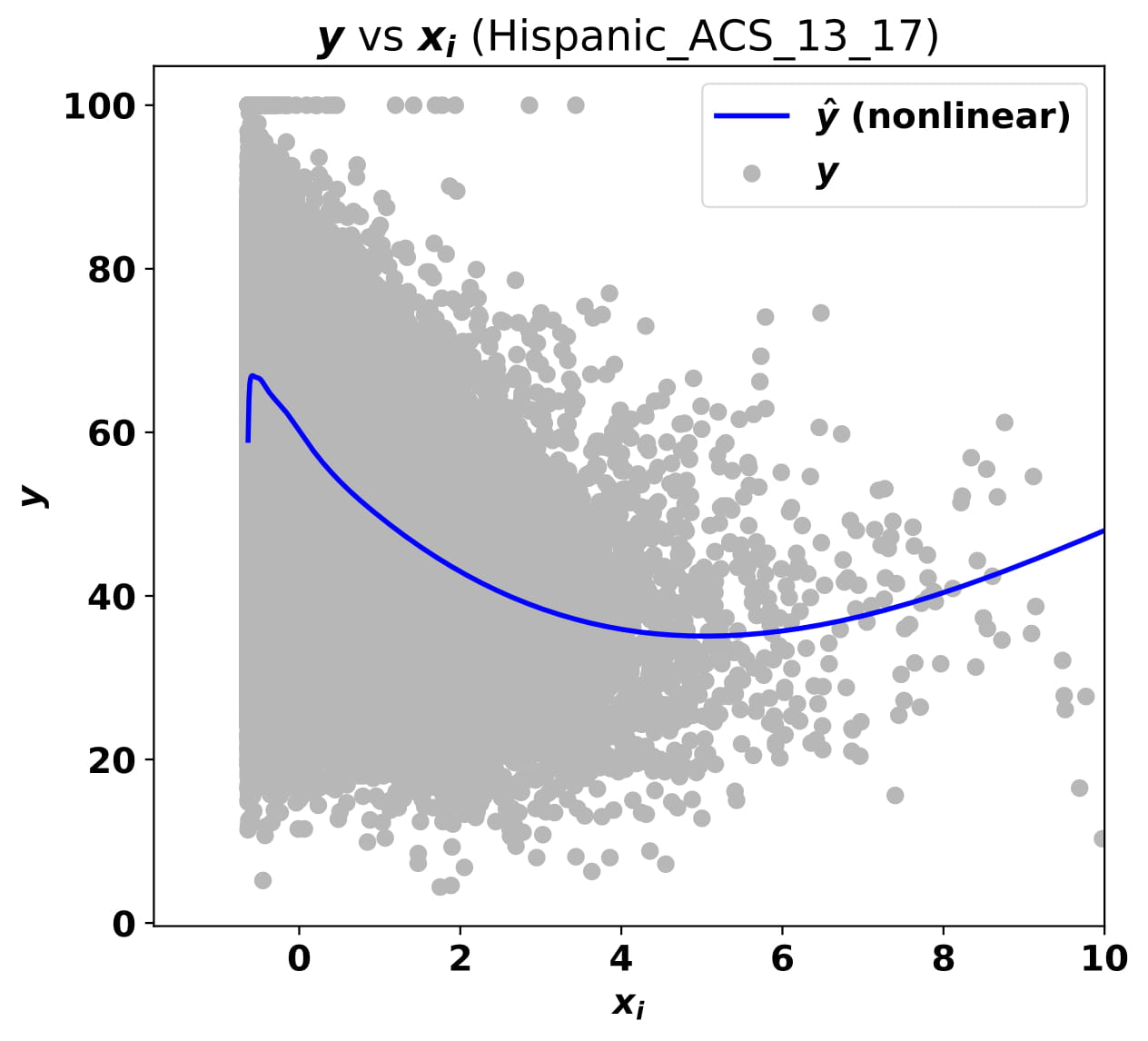}&  \includegraphics[width=0.3\textwidth]{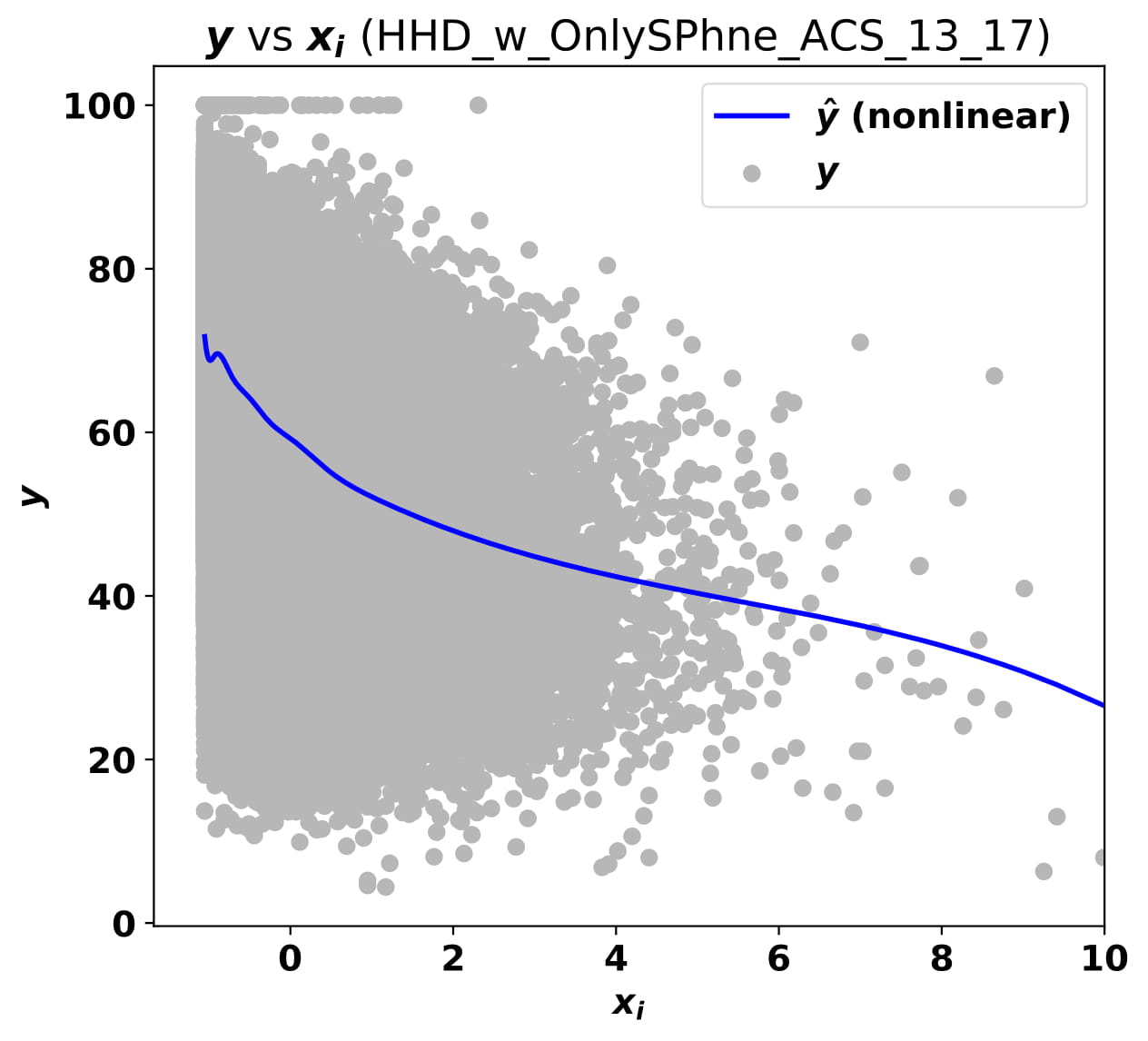}& \includegraphics[width=0.3\textwidth]{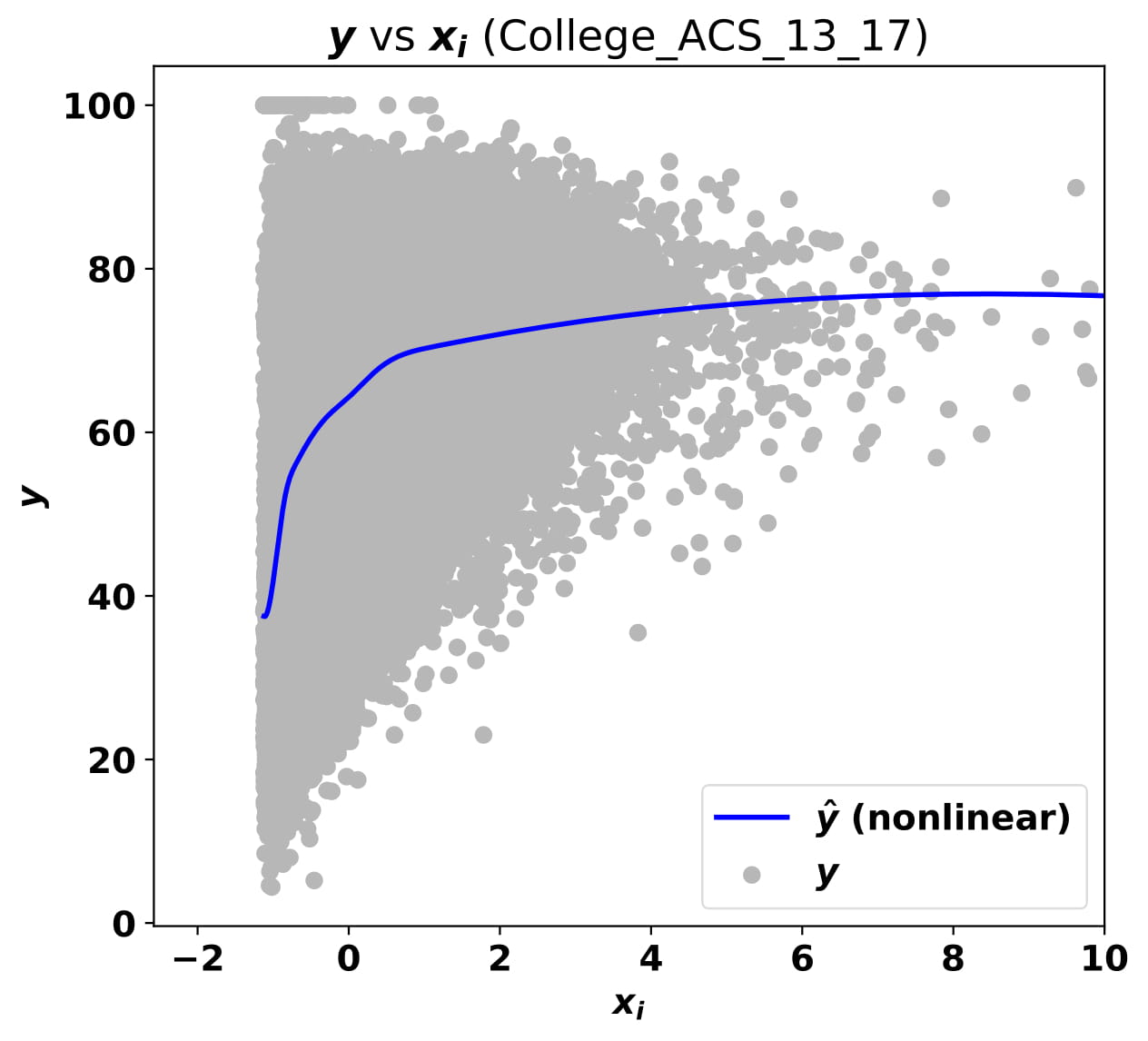}
\end{tabular}
        \caption{\textit{Panels [Left]-[Right] illustrate marginal nonparametric fits for the self-response rate output variable versus three covariates. Each marginal fit, displayed on a scatter plot with a solid blue line, clearly suggests a nonlinear relationship of the output vs the individual covariate (we note that the covariates are standardized.)
         The $x$-axis corresponds to: [Left] ``Persons of Hispanic origin in the ACS''; [Middle] ``Number of households that have only a smartphone and no other computing device''; [Right] ``Persons 25 years and over with college degree or higher in the ACS''.}} 
        \label{fig:Nonlinear predictors}
\end{figure}

We assume that the components~$\{f_j\}$ and~$\{f_{j,k}\}$ are smooth (for example, twice continuously differentiable). For illustration, consider Figure~\ref{fig:Nonlinear predictors}, which shows that the marginal fits for the self-response rate appear to be well-approximated by smooth  nonlinear functions.
We let $\M{f}_{j} = (f_{j}(x_{1j}), \ldots, f_{j}(x_{nj}))$ and $\M{f}_{j,k} = (f_{j,k}(x_{1j}, x_{1k}), \ldots, f_{j,k}(x_{nj}, x_{nk}))$ denote the evaluations of the main effect component $f_{j}$ and the interaction component $f_{j,k}$, respectively, at the~$n$ data points. Writing~$\M{y}$ for the response vector and using squared $\ell_{2}$-loss as the data fidelity term, the task of learning~\eqref{eq:GAMs with interactions} can be expressed as the following optimization problem:
\begin{align} \label{eqn:base-GAM-with interactions}
   \min_{\substack{f_j \in {\mathcal C}_{1}, \forall j  \\
   f_{j,k} \in \mathcal{C}_{2}, \forall j<k}}~\frac{1}{n} \big\|\M{y} -  \sum_{j\in [p]} \M{f}_{j} - \sum_{j<k } \M{f}_{j,k}\big\|_2^2 + \lambda_1 \big[\sum_{j \in [p]} \Omega(f_j) + \sum_{j<k} \Omega(f_{j,k})\big],
\end{align}
where 
${\mathcal C}_1$ and ${\mathcal C}_{2}$ denote the sets of smooth candidate functions,  $\Omega$ is a roughness penalty\footnote{For example, $\Omega(f_j) = \int f_j^{\prime\prime}(x_j)^2 dx_j$ and $\Omega(f_{ij}) = \int_{x_j x_k} (\partial^2 f_{j,k} / \partial x_j^2)^2+(\partial^2 f_{j,k} / \partial x_j\partial x_k)^2  + (\partial^2 f_{j,k} / \partial x_k^2)^2 dx_j dx_k$.}, and $\lambda_1$ is a non-negative regularization parameter controlling the smoothness of the fit. 
We show in Section~\ref{sec: Efficient Computations at scale} that problem~\eqref{eqn:base-GAM-with interactions} can be written as a finite-dimensional quadratic program by using cubic splines to model each of the main and interaction effects.

\subsection{Parsimonious models via $\ell_0$-penalization}\label{sec: Parsimonious interactions} 
Here, we study $\ell_0$-type estimators that limit the number of components in the additive models introduced earlier. 

\subsubsection{Sparse pairwise interactions} 
\begin{figure}[!tb]
    \centering
    \scalebox{0.98}{\begin{tabular}{cc}
    \includegraphics[width=0.49\textwidth]{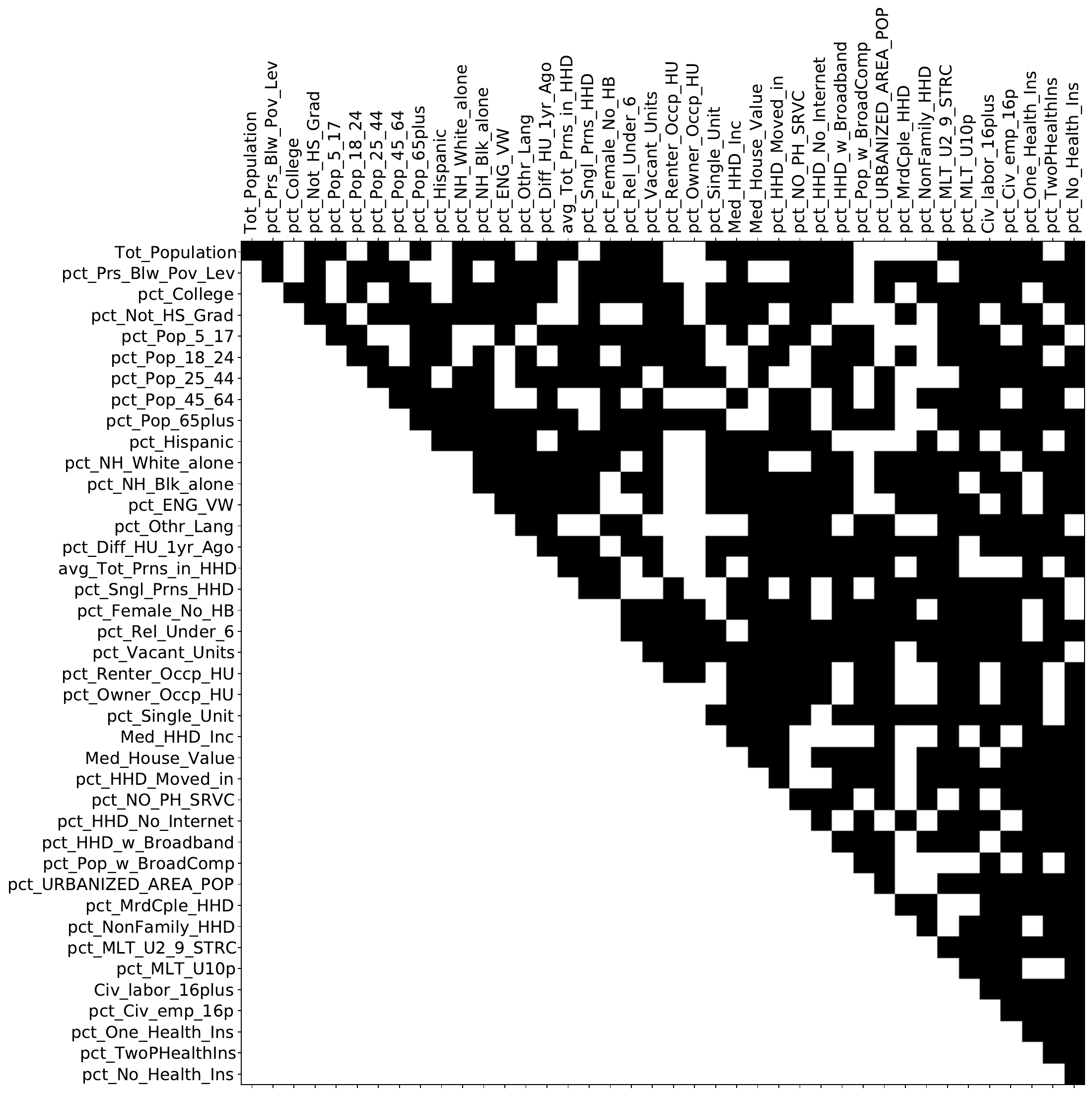}&  \includegraphics[width=0.49\textwidth]{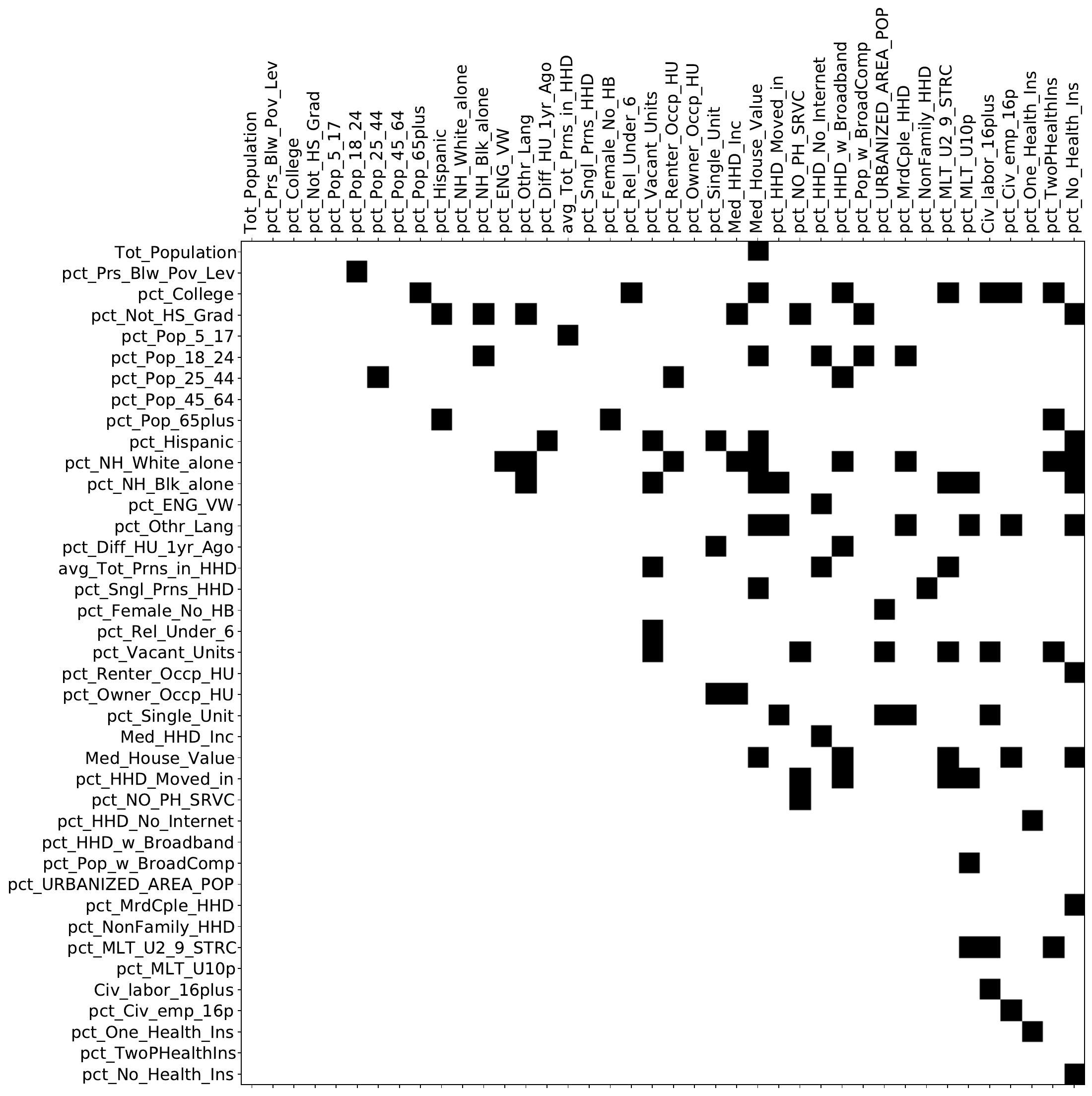}
 \end{tabular}}
    \caption{\textit{Sparsity pattern of the main and interaction effects presented in a $p \times p$ matrix: a black square on the diagonal indicates the presence of a main effect, and an off-diagonal black square indicates the presence of an interaction effect in the joint model.
    [Left] Panel illustrates the sparsity pattern of a Lasso model with main and interaction effects. There are 37 main and 555 interaction effects in the optimal model. 
    [Right] Panel illustrates the sparsity pattern of a nonlinear AM with main and interaction effects, i.e., model~\eqref{eq: GAM with interactions L0 FunForm}. There are 8 main and 92 interaction effects in the optimal model.
    Model~\eqref{eq: GAM with interactions L0 FunForm} has prediction performance similar to the Lasso model, with only 3 main and 33 interaction effects. Nonlinear models lead to much more compact models and, hence, are easier to interpret than linear models with interactions.
    Both models were trained on a 2019 US Census Bureau Planning Database dataset (predicting the tract-level self-response rate) with $p = 40$ covariates and $74,000$ observations.}}
    \label{fig:Spy}
\end{figure}

While a significant body of work has been devoted to studying sparsity in the context of linear models, sparsity in nonlinear models has received relatively less attention.
Interestingly, we observe that the notion of parsimony is linked to the model being used and changes, for example, depending on whether we use a linear interaction model of the form $\mathbb{E}(y|\M{x}) = \sum_{j} x_{j}\beta_{j} + \sum_{j<k} x_{j} x_{k}\beta_{jk}$ or model~\eqref{eq:GAMs with interactions} that has nonlinear components.
For motivation, we consider Figure \ref{fig:Spy}, which presents our findings on a 2019 US Census Bureau Planning Database dataset with $40$ covariates\footnote{These covariates were motivated by our discussions with researchers at the US Census Bureau.}. These features include covariates used for the low-response-score \citep{Erdman2014},  important covariates highlighted in Appendix C of the 2019 US Census Bureau Planning Database Documentation \citep{Bureau2019}, plus some additional covariates capturing internet penetration and urbanization. Specifically, we observe that:
\vspace{4pt}
\begin{compactitem}
    \item  Our proposed AM approach with \textit{nonlinear} main-effects and  interactions\footnote{This corresponds to an estimate available from~\eqref{eq: GAM with interactions L0 FunForm}. We present the model leading to the best prediction performance on the validation set -- see Section~\ref{sec:expts-case-study} for details.} results in a substantially more compact model than a {\textit{linear}} model with interactions.
    \item The above nonparametric additive model leads to better test predictions than its linear model counterpart.
    \item The best  (based on validation tuning with prediction error) model with linear main and interaction effects contains~$37$ main and~$555$ interaction effects. On the other hand, using nonparametric AMs, we need 36 nonlinear main and interaction effects to obtain a similar test prediction performance. The nonlinear model also uses fewer covariates compared to the linear model counterpart.  
\end{compactitem} 
\vspace{4pt}
The above observations suggest that the nonlinear model with sparse interactions has advantages over its linear model counterpart for predicting self-response rates. 
As discussed in Section~\ref{sec:expts-case-study}, the above findings generally carry over to the expanded dataset with more features.

To obtain an additive model with a small number of main and interaction effects, we consider an $\ell_0$-penalized modification of optimization problem~\eqref{eqn:base-GAM-with interactions}. We write 
\begin{equation}
\label{def.Omega}
    \Omega_{\text{gr}}(f)=\sum_{j \in [p]} \Omega(f_j) + \sum_{j<k} \Omega(f_{j,k}) \qquad\text{and}\qquad 
    \M{f}=\sum_{j\in [p]} \M{f}_{j} + \sum_{j<k } \M{f}_{j,k},
\end{equation}
to simplify the expressions and propose an estimator based on the following problem:
\begin{align} \label{eq: GAM with interactions L0 FunForm}
   \min_{\substack{f_j \in {\mathcal C}_{1}, \forall j  \\
   f_{j,k} \in \mathcal{C}_{2}, \forall j<k}}~\frac{1}{n}  \big\|\M{y} -  \M{f}\big\|_2^2 + \lambda_1 \Omega_{\text{gr}}(f)+ \lambda_2 \Big[\sum_{j \in [p]} \mathbb{1} \bigl[\B{f}_j \neq \B 0 \bigr] + \alpha \sum_{j<k} \mathbb{1} \bigl[\B{f}_{j,k} \neq \B 0 \bigr] \Big],
\end{align}
where $\mathbb{1}[\cdot]$ is an indicator function, $\lambda_2\in \mathopen[0,\infty\mathclose)$ controls the number of selected components, and $\alpha \in \mathopen[1,\infty\mathclose)$ controls the tradeoff between the number of main and interaction effects.
We note that the finite-dimensional version of Problem~\eqref{eq: GAM with interactions L0 FunForm} can be formulated as a mixed integer program (MIP)~\citep{wolsey1999integer}, and hence can be solved to optimality with modern commercial solvers (for example, Gurobi, Mosek) for small/moderate scale problems. Tailored nonlinear branch-and-bound techniques~\citep{hazimeh2020sparse} are promising for solving large-scale instances of $\ell_0$-sparse linear regression problems (to optimality). 
Since we intend to compute solutions to~\eqref{eq: GAM with interactions L0 FunForm} for a family of tuning parameters $(\lambda_{1}, \lambda_{2})$ at scale, we consider high-quality approximate solutions (cf Section~\ref{sec: Efficient Computations at scale}). Once a good solution to~\eqref{eq: GAM with interactions L0 FunForm} is available, we can employ MIP techniques to improve the solution and/or certify the quality of the solution, extending the local search techniques presented in~\cite{HazimehL0Learn} for 
$\ell_0$-sparse linear regression problems.  
Our methodological investigation of estimator~\eqref{eq: GAM with interactions L0 FunForm} is of independent interest. 
We refer to this model as \modelbname\footnote{\modelbname~stands for \underline{E}nd-to-end \underline{L}earning \underline{A}pproach for \underline{A}dditive spli\underline{N}es with \underline{I}nteractions.}.

While in this paper, we study a group $\ell_0$ penalty in~\eqref{eq: GAM with interactions L0 FunForm}, one can also consider other non-convex penalty functions: for example, based on the (group) SCAD \citep{Fan2001} and MCP \citep{Zhang2010}. Exploring the statistical and computational aspects of using such penalties would be interesting to pursue and is left as future work.

\noindent {\bf Related Work.} 
In the special case of~\eqref{eq: GAM with interactions L0 FunForm}, when no interactions are present, several convex relaxations of the group $\ell_0$-penalty have been studied~\citep{Ravikumar2009,Zhao2012,Meier2009,Buhlmann2011}\footnote{In terms of existing implementations, R package \texttt{SAM} presents specialized algorithms for a convex relaxation of~\eqref{eq: GAM with interactions L0 FunForm} without interactions. \texttt{SAM}, however,  would not run on the dataset we consider here.}.
\cite{hazimeh2021group} consider $\ell_0$-formulations for the setting without interaction effects and demonstrate the merits of using $\ell_0$-based formulations over convex relaxation-based approaches.
Despite the appeal of estimator~\eqref{eq: GAM with interactions L0 FunForm}, computational challenges appear to be a key limiting factor in exploring this model in practice. In Section~\ref{sec: Efficient Computations at scale}, we present a new algorithm for problem~\eqref{eq: GAM with interactions L0 FunForm}.

\subsection{Statistical Theory}\label{sec:theory}

In this section, we explore the statistical error bounds of our proposed estimator in the deterministic design setting. The supplementary material \citep{supplement} provides technical details, assumptions, and proofs of the established results. \\

We write~$\mathcal{C}_\text{gr}$ for the space of functions $f(\M{x})=\sum_{j\in[p]} f_j(x_j)+\sum_{j<k} f_{j,k}(x_j,x_k)$, where the components belong to $L_2$-Sobolev spaces, and the sparsity level of the representation is bounded. 
We write $G(f)$ for the number of components in the above representation:
\begin{equation*}
  G(f)=\sum_{j \in [p]} \mathbb{1} \bigl[\B{f}_j \neq \B 0 \bigr] + \sum_{j<k} \mathbb{1} \bigl[\B{f}_{j,k} \neq \B 0 \bigr],  
\end{equation*}
recall the definitions of~$\Omega_{\text{gr}}$ and~$\bff$ given in~(\ref{def.Omega}), and let~$\|\cdot\|_n$ denote the Euclidean norm divided by~$\sqrt{n}$. We analyze the solution to optimization problem~\eqref{eq: GAM with interactions L0 FunForm} with $\alpha=1$. After a small change in notation, we focus on the following optimization problem:
\begin{equation}
\label{crit.add1}
\widehat{f}_n=\argmin_{f\in\mathcal{C}_\text{gr}}\;\,\|\bY- \bff\|_n^2+\lambda_n \Omega_{\text{gr}}(f) +\mu_n G(f),
\end{equation}
where~$\lambda_n$ and~$\mu_n$ are nonnegative tuning parameters controlling the amount of the smoothness penalty and the cardinality penalty, respectively.

We assume that the observed data follows the model $\M{y}=\bff^*+\bepsilon$, where $f^*$ is the unknown true regression function of an arbitrary form, and elements of~$\bepsilon$ are independent $N(0,\sigma^2)$. We refer to $\|\widehat\bff_n-\bff^*\|^2_n$ as the prediction error for estimator~$\widehat{f}_n$.
We write~$e$ for the base of the natural logarithm and define $r_n=n^{-1/3}$, noting that $r_n^2$ is the optimal prediction error rate in the bivariate regression setting where $f^*\in\mathcal{C}_\text{gr}$.
We say that a constant is universal if it does not depend on other parameters, such as~$n$ or~$p$. We use the notation~$\lesssim$ to indicate that inequality~$\le$ holds up to a positive universal multiplicative factor.

Theorem~\ref{add.thm} presented below derives a general non-asymptotic oracle prediction error bound, comparing the performance of our estimator to that of sparse approximations to the true regression function. Bounds of this type have been established for sparse nonparametric AMs with main effects; however, we are unaware of similar existing bounds for nonparametric models with pairwise interactions. The existing work has focused mainly on the Lasso-based estimators, which encourage sparsity in the main effects by penalizing the magnitudes of the functional components \cite[see][and the references therein]{Meier2009, tan2019doubly}.
The use of $\ell_1$-based relaxations, in lieu of $\ell_0$-penalization, to induce sparsity in the main effects can lead to unwanted shrinkage, which may interfere with variable selection. The approach in \cite{hazimeh2021group} directly controls the number of main effects, demonstrating the benefits of $\ell_0$-regularization theoretically and empirically.

Our analysis accounts for the interaction effects and focuses on the $\ell_0$-penalized formulation, which poses additional theoretical challenges relative to the $\ell_0$-constrained formulation. To our knowledge, the error bounds established here for sparse nonparametric AMs with pairwise interactions are novel. 

The following result corresponds to the high-dimensional setting where~$p$ is large (we discuss the classical fixed~$p$ case at the end of this section), hence the stated error bound holds with high probability.
\begin{theorem}
\label{add.thm}
Let $\widehat{f}_n$ be defined as in~(\ref{crit.add1}). Then, there exists a universal constant~$c_1$, such that if $\lambda_n\ge c_1\sigma\big[r_n^2+r_n\sqrt{\log (ep)/n}\,\big]$, then
\begin{equation*}
\|\widehat\bff_n-\bff^*\|^2_n + \lambda_n\Omega_{\text{\rm gr}}(\widehat{f}_n)+\mu_nG(\widehat{f}_n)\lesssim \inf_{f\in \mathcal{C}_{\text{gr}}}   \Big[\|\bff-\bff^*\|^2_n + \lambda_n\Omega_{\text{\rm gr}}(f)+\mu_nG(f)\Big] +\sigma^2 \Big[r_n^2+\frac{\log (ep)}{n}\Big]
\end{equation*}
with probability at least $1-1/p$.
\end{theorem}
We note that a popular approach in the literature, even in the linear setting, has been to assume a sparse underlying model for the data. In contrast, Theorem~\ref{add.thm} does not assume a ground truth sparse additive model, allowing for model misspecification and incorporating the approximation error into the bound. This feature is especially appealing for our application of interest where the ground truth is unavailable. The following corollary complements the general result in Theorem~\ref{add.thm} by establishing a non-asymptotic prediction error bound for the proposed approach in the case where the additive model is correctly specified.
\begin{corollary}
\label{add.cor}
Suppose that $f^*\in\mathcal{C}_\text{gr}$. There exist universal constants~$c_1$ and~$c_2$, such that if $\lambda_n\ge c_1\sigma\big[r_n^2+r_n\sqrt{\log (ep)/n}\big]$ and $\mu_n\ge c_2\sigma^2 \big[r_n^2+\log(ep)/n\big] + c_2\lambda_n\Omega_{\text{\rm gr}}(f^*)$, then
\begin{equation}
\label{np.or.bnd.cor}
\|\widehat\bff_n-\bff^*\|^2_n \lesssim \sigma^2 \Big[r_n^2+\frac{\log ({e}p)}{n}\Big] + \lambda_n\Omega_{\text{\rm gr}}(f^*) + \mu_nG(f^*)\quad\text{and}\quad
G(\widehat{f}_n)\le G(f^*)
\end{equation}
with probability at least $1-1/p$.
\end{corollary}

We note that inequality~\eqref{np.or.bnd.cor} yields the following prediction error bound:
\begin{equation}
\label{error.rate}
\|\widehat\bff_n-\bff^*\|^2_n \lesssim \sigma^2 \Big[n^{-2/3}+\frac{\log(ep)}{n}\Big],
\end{equation}
which holds with probability at least $1-1/p$ for an appropriate choice of the tuning parameters.
For clarity of presentation, we do not specify the multiplicative constants in inequalities~\eqref{np.or.bnd.cor} and~\eqref{error.rate}, focusing instead on the corresponding error rate. In contrast to the Lasso-based estimators for nonparametric AMs \cite[e.g.,][]{Meier2009,tan2019doubly}, the error rate in~\eqref{error.rate} holds without imposing assumptions on the design. The accompanying sparsity bound $G(\widehat{f}_n)\le G(f^*)$ is exact, and hence stronger than the corresponding bounds for the Lasso-based estimators \cite[e.g.,][]{lounici1}. The latter bounds hold up to a multiplicative constant, which depends on the design and can be significantly greater than one.
When~$\log(p)\lesssim n^{1/3}$, the prediction error rate in~(\ref{error.rate}) matches the optimal bivariate rate of~$n^{-2/3}$.

Theorem~\ref{add.thm} and Corollary~\ref{add.cor} are stated for the high-dimensional setting, where~$p$ is large. We show in the proof of Corollary~\ref{add.cor} that in the classical asymptotic setting, where $p$ is fixed and $n$ tends to infinity, an appropriate choice of~$\lambda_n$ and~$\mu_n$ leads to the optimal bivariate rate of convergence, $\|\widehat\bff_n-\bff^*\|^2_n=O_p\big(n^{-2/3}\big)$, and the exact sparsity bound $G(\widehat{f}_n)\le G(f^*)$ that holds with probability tending to one.

The established error bounds also have implications for model selection. Let~$\mathcal{I}(f)$ denote the index set specifying the main and interaction effects for $f\in\mathcal{C}_\text{gr}$. Suppose that $f^*\in\mathcal{C}_\text{gr}$ and the true regression model is asymptotically identifiable in the sense that  $\|\bff-\bff^*\|_n$ is bounded away from zero for all sufficiently large~$n$ uniformly over the class $\{f\in\mathcal{C}_\text{gr}, \, G(f)\le G(f^*), \, \mathcal{I}(f)\ne\mathcal{I}(f^*)\}$. If, in addition, $p=o(e^n)$ as~$n$ goes to infinity, then $\mathcal{I}(\widehat{f}_n)=\mathcal{I}(f^*)$ with probability tending to one, i.e., the proposed approach is model selection consistent.

\subsection{Efficient Computations at Scale} \label{sec: Efficient Computations at scale}

We present specialized algorithms for obtaining good solutions to Problem~\eqref{eq: GAM with interactions L0 FunForm}. 
Our approach scales to the problem-size of the Census application, with $n \approx 10^5$ and $p \approx 500$, which poses formidable computational challenges due to the presence of approximately $10^5$ interaction effects\footnote{Note that, using 25 knots for every component, this leads to estimating around 2.5 million basis coefficients.}. 

To obtain good solutions at scale, we use techniques inspired by first-order methods in continuous optimization~\citep{nesterov2003introductory} and careful exploitation of the problem structure. A high-level summary is presented below, with the details relegated to the Supplement \citep{supplement}.

\noindent {\bf State-of-the-art approaches.} To appreciate the computational challenges of sparse nonlinear AMs with interactions, and nonparametric AMs in general, we provide a few examples of the problem instances that can be handled by current state-of-the-art algorithms with publicly available implementations. Current implementations based on R package \texttt{SAM}~\citep{Zhao2012}, the stepwise GAM function in R package \texttt{step.gam} 
(which performs greedy variable selection), and Python package \texttt{pyGAM}, take on the order of days to run and/or face numerical difficulties for instances with $n \approx 10^5$ to obtain a single solution without interactions. \texttt{pyGAM} is not designed to do variable selection. \cite{Wood2017} present an interesting approach for AMs that scales to large $n$ settings (see R package \texttt{mgcv}) but doesn't appear to perform automated variable selection in the presence of a large number of features  -- \citet{Wood2017} report instances containing fewer than $20$ pre-specified main and interaction effects.
EBM (\citealp{Nori2019}) is a state-of-the-art computational approach for AMs, but does not allow for feature selection as all the main effects are included in the estimated model. Additionally, the variant of EBM that selects interaction effects results in many interaction effects, leading to sub-optimal support recovery -- see Sections  \ref{sec:simulation-case-study} and \ref{sec:expts-case-study} for an illustration. The GAMI-Net method (\citealp{Yang2021}) is computationally expensive. Additionally, since it uses a multi-layered NN for every interaction effect it requires $10-100\times$ more parameters when compared with the approaches we consider.

\noindent{\bf Algorithms for sparse nonlinear interactions: Problem~\eqref{eq: GAM with interactions L0 FunForm}.}
We represent the main and interaction effects as linear combinations of cubic spline basis functions (see the Supplement \citep{supplement} for the specific details). In particular, we let $\M{f}_{j} = \M{B}_{j} \B{\beta}_{j}$, where $\M{B}_{j} \in \mathbb{R}^{n \times K_j}$ is the model matrix and $\B{\beta}_{j} \in \mathbb{R}^{K_{j}}$ is the vector of coefficients for each main effect component. Similarly, we let $\M{f}_{j,k} = \M{B}_{j,k} \B{\theta}_{j,k}$, where $\M{B}_{j,k} \in \mathbb{R}^{n \times K_{j,k}}$ is the model matrix and $\B{\theta}_{j,k} \in \mathbb{R}^{K_{j,k}}$ is the vector of coefficients for each interaction effect component.
Writing~$\B{\beta}$ for the vector obtained by stacking together the coefficients $\B{\beta}_j$, $j\in[p]$ for the main-effects, and defining the vector~$\B{\theta}$ for the interaction effects analogously, we express the objective function in~\eqref{eqn:base-GAM-with interactions} as follows: 
\begin{equation*}
\small
   g_{\lambda_1}( \B{\beta}, \B{\theta}) \;\eqdef\; \frac{1}{n} \Big\|\B{y} - \big[\sum_{j \in [p]} \B{B}_j \B{\beta}_j + \sum_{j<k} \B{B}_{j,k} \B{\theta}_{j,k} \big]\Big\|_2^2 + \lambda_1 \big[\sum_{j \in [p]} \B{\beta}_j^T \B{S}_j \B{\beta}_j + \sum_{j<k} \B{\theta}_{j,k}^T \B{S}_{j,k} \B{\theta}_{j,k} \big].
\end{equation*}
Here, $\B{S}_{j} = \B{D}_j^T \B{D}_j$ and $\B{S}_{j,k} = (\B{D}_j^T \B{D}_j)\otimes \B{I}_k + \B{I}_j \otimes (\B{D}_k^T \B{D}_k)$ are the smoothness penalty matrices for the main effects and the interaction components, respectively (further details are provided in the Supplement \citep{supplement}). For convenience, we use the same smoothness penalty~$\lambda_{1}$ for both the main and the interaction effects, though in general, they may be taken to be different. The above representation leads to the following form of the optimization problem~\eqref{eq: GAM with interactions L0 FunForm}:
\begin{align} \label{eq: GAM with interactions L0}
    \min_{\B{\beta}, \B{\theta}} G(\B \beta, \B \theta) \;\eqdef\; g_{\lambda_1}(\B{\beta}, \B{\theta}) + \lambda_2 \Big[\sum_{j \in [p]} \mathbb{1} \bigl[\B{\beta}_j \neq \B 0 \bigr] + \alpha \sum_{j<k} \mathbb{1} \bigl[\B{\theta}_{j,k} \neq \B 0 \bigr] \Big].
\end{align}
The indicator functions in the above equation are applied to vectors of basis coefficients corresponding to particular main or interaction effects. Hence, for example, when $\mathbb{1} \bigl[\B{\beta}_j \neq \B 0 \bigr]$ equals zero, the entire main effect~$\M{f}_{j}$ is also zero.
As mentioned earlier,~\eqref{eq: GAM with interactions L0} can be expressed as a MIP and solved for small-to-moderate scale problems using modern MIP solvers. Developing specialized branch-and-bound based solvers for Problem~\eqref{eq: GAM with interactions L0} along the lines of~\cite{hazimeh2022sparse,hazimeh2021group} is beyond the scope of this work. 
We present fast approximate algorithms to obtain good solutions to Problem~\eqref{eq: GAM with interactions L0} for a family of tuning parameters. We note that the objective in Problem~\eqref{eq: GAM with interactions L0} is a sum of a smooth convex loss function and a discontinuous regularizer separable across the components $\{\B\beta_{j}\}$ and $\{\B\theta_{i,k}\}$. Motivated by the strong empirical performance of cyclical coordinate descent (CD) methods~\citep{wright2015coordinate} in $\ell_0$-penalized linear regression~\citep{HazimehL0Learn}, we explore block CD methods to obtain fast approximate solutions for the nonparametric setting with interactions~\eqref{eq: GAM with interactions L0}. For convergence guarantees of this procedure, see~\cite{HazimehL0Learn} and references therein. 

In our block CD method, the blocks correspond to the basis coefficients for either the main effects $\{\B\beta_{j}\}$ or the interaction effects $\{\B\theta_{j,k}\}$. Given an initialization 
$(\B \beta_1^{(0)}, \cdots, \B \beta_{p}^{(0)}, \B \theta_{1,2}^{(0)}, \\ \cdots, \B \theta_{p-1,p}^{(0)})$, at every cycle, we sequentially sweep across the main effects and the interaction effects.  In particular, if we denote the solution after $t$ cycles by $(\B \beta_1^{(t)}, \cdots, \B \beta_{p}^{(t)}, \B \theta_{1,2}^{(t)}, \cdots, \B \theta_{p-1,p}^{(t)})$, then the block of coefficients for $j$-th main effect at cycle $t+1$, namely, $\B \beta_{j}^{(t+1)}$, is obtained by optimizing~\eqref{eq: GAM with interactions L0} with respect to~$\B \beta_{j}$, with other variables held fixed:
\begin{align}
\label{eq:bcd-main}
    \B \beta_{j}^{(t+1)} \in \underset{\B{\beta}_j \in \mathbb{R}^{K_j}}{\argmin}~G(\B \beta_1^{(t+1)}, \cdots, \B \beta_{j-1}^{(t+1)}, \B \beta_j^{}, \B \beta_{j+1}^{(t)}, \cdots, \B \beta_{p}^{(t)}, ~~ \B \theta_{1,2}^{(t)}, \cdots, \B \theta_{p-1,p}^{(t)}).
\end{align}
We update $\B \theta_{j,k}^{(t+1)}$, containing the coefficients for the $(j,k)$-th interaction at cycle $t+1$, using
\begin{align}
\label{eq:bcd-interaction}
    \B \theta_{j,k}^{(t+1)} \in \underset{\B{\theta}_{j,k} \in \mathbb{R}^{K_{j,k}}}{\argmin}~G(\B \beta_1^{(t+1)}, \cdots, \B \beta_{p}^{(t+1)}, ~~ \B \theta_{1,2}^{(t+1)}, \cdots, \B \theta_{(j,k)-1}^{(t+1)}, \B \theta_{j,k}^{}, \B \theta_{(j,k)+1}^{(t)} \cdots, \B \theta_{p-1,p}^{(t)}).
\end{align}
The block minimization problems~\eqref{eq:bcd-main} and~\eqref{eq:bcd-interaction} can be solved in closed-form as discussed in the Supplement \citep{supplement}. These block CD updates need to be paired with several computational devices in the form of active set updates, cached matrix factorizations,  and warm-starts, among others. We draw inspiration from similar strategies used in CD-based procedures for sparse linear  regression~\citep{HazimehL0Learn,Friedman2010}, and extend them to our problem. 
See the Supplement \citep{supplement} for more details. 

CD methods have old roots in optimization: see for example, the review paper~\cite{wright2015coordinate}. CD methods  have close links with the well known Gauss-Seidel method (for solving linear/nonlinear systems of equations)---similar methods have been extensively used in additive 
models~\citep{Bhastie5}, where they are referred to as  \textit{backfitting}.  
CD schemes (including their block variants) arising from the optimization literature can be efficiently applied to optimization problems involving sparsity constraints: such problems arise in fitting additive models with sparsity-inducing penalties~\citep{Ravikumar2009,Hastie2015,hazimeh2021group}.  
We note that there are other successful additive model fitting approaches, for example, smooth backfitting \citep{mammen1999existence,mammen2006simple}, which have excellent theoretical properties.
These works focus on the additive model setup without sparsity---extending these approaches to our setting of large-scale sparse additive models in a computationally efficient fashion is an interesting direction for future research.

\section{Incorporating strong hierarchy constraints}\label{sec:nonparm-int-SH}
This section discusses the model with sparse interactions under strong hierarchy. 

Problem~\eqref{eq: GAM with interactions L0} limits the total number of main and interaction effects and works well in our experiments (see Section~\ref{sec:expts-case-study} for the details) in terms of obtaining a sparse model with good predictive performance. In terms of variable selection properties, however,~\eqref{eq: GAM with interactions L0} can lead to the inclusion of an interaction effect, say, $\{(j,k)\}$ where at least one of the corresponding main-effects $\{j\}$ or $\{k\}$ is excluded from the model. This may be somewhat problematic from an interpretation viewpoint---it may be desirable to enforce additional constraints in~\eqref{eq: GAM with interactions L0}, such as the \textit{hierarchy constraints}~\citep{mccullagh1989generalized,Bien2013}. In this paper, we consider the \textit{strong hierarchy} constraint, where an interaction effect $\{(j,k)\}$ is included in the model only if both the corresponding main effects are also included. In addition to improved interpretation, strong hierarchy can reduce the \textit{effective} number of features in the model, subsequently reducing the operational costs associated with data collection~\citep{Bien2013,HazimehHS}. 

For example, based on our analysis, covariates ``Males'' and ``Females'' appear to be unnecessary for self-response prediction when the strong hierarchy constraint is imposed via model~\eqref{eq: GAM with interactions L0 with hierarchy}, while a related covariate ``Households with female without spouse'' remains important. In contrast, the LRS uses both ``Males'' and ``Households with females without spouse''~\citep{Erdman2016} -- both features appear to be important perhaps due to the use of a linear regression model.
Similarly, covariates pertaining to civilian employment status across different segments/age-groups of the population are excluded from our nonlinear AM
when we learn the (sparse) main and interaction effects under a strong hierarchy.

To enforce strong hierarchy into model~\eqref{eq: GAM with interactions L0}, we consider the following estimator:
\begin{subequations} \label{eq: GAM with interactions L0 with hierarchy}
\begin{align}
    \min_{\B{\beta}, \B{\theta}}~~ &~~ \, g_{\lambda_1}(\B{\beta}, \B{\theta}) +\lambda_2 \big(\sum_{j \in [p]} \mathbb{1} [\B{\beta}_j \neq \B 0 ] + \alpha \sum_{j<k} \mathbb{1} [\B{\theta}_{j,k} \neq \B 0 ] \big)  \label{hier-eqn-line1}\\
    \text{s.t.} ~~&~~ \,\B{\theta}_{j,k} \neq \B 0 ~~ \Longrightarrow~~ \B{\beta}_{j} \neq \B 0 ~~\&~~ \B{\beta}_{k} \neq \B 0~~~~\forall j < k,~j \in [p],~k\in [p].  \label{hier-eqn-line2}
    \end{align}
\end{subequations}
We note that~\eqref{eq: GAM with interactions L0 with hierarchy} differs from~\eqref{eq: GAM with interactions L0} in the additional strong hierarchy constraint appearing in~\eqref{hier-eqn-line2}. By using binary variables to model sparsity in the main/interaction effects and to encode the hierarchy constraint~\eqref{hier-eqn-line2}, Problem~\eqref{eq: GAM with interactions L0 with hierarchy} can be expressed as the following MIP: 
\begin{subequations}\label{eq: GAM with interactions L0 with hierarchy MIP}
\begin{align}
    \min_{\B{\beta}, \B{\theta}, \B{z}}~~& g_{\lambda_1}(\B{\beta}, \B{\theta}) +\lambda_2 \big(\sum_{j\in [p]} {z_j} + \alpha \sum_{j<k} z_{j,k} \big)  \\
    \text{s.t.}~~& z_j, z_{j,k} \in \{0,1\},~~\norm{\B{\beta}_j}_2^{2} \leq M z_j,~~\norm{\B{\theta}_{j,k}}_2^{2} \leq M z_{j,k}~~~\forall j < k,~j,k \in [p], \\   
   & z_{j,k} \leq z_j,~~z_{j,k} \leq z_k,~~~\forall j<k, \label{eq:hier-eqn-mip-line3}
\end{align}
\end{subequations}
where the BigM parameter~$M$ is a sufficiently large finite constant such that an optimal solution to~\eqref{eq: GAM with interactions L0 with hierarchy MIP} satisfies $\max_{j} \| \B\beta_{j} \|_{2}^{2} \leq M$ and $\max_{j,k}  \| \B\theta_{j,k}\|_{2}^{2} \leq M$.
Binary variable $z_{j}$ (and $z_{j,k}$) indicates whether the corresponding main effect $\B\beta_{j}$ (respectively, interaction effect $\B\theta_{j,k}$) is zero or not; the constraint appearing in~\eqref{eq:hier-eqn-mip-line3} enforces the hierarchy constraint in~\eqref{hier-eqn-line2}.

To our knowledge, we present a novel computational model to study estimator~\eqref{eq: GAM with interactions L0 with hierarchy MIP}. The methodology presented here is of independent interest in the context of structured nonparametric learning with interactions.
We refer to this model as \modelcname\footnote{\modelcname~stands for \underline{E}nd-to-end \underline{L}earning \underline{A}pproach for \underline{A}dditive spli\underline{N}es with \underline{H}ierarchy.}.
In the Supplement \citep{supplement}, we propose algorithms to obtain good solutions to Problem~\eqref{eq: GAM with interactions L0 with hierarchy MIP}.

\noindent {\bf Related Work.}  Methodology for strong hierarchy in linear models has been studied in the statistics/machine learning literature~\citep{Bien2013, Lim2015, Yan2017, HazimehHS} -- these works focus on the linear model setting, a special case of the nonlinear setting we consider here.
\cite{Radchenko2010}~consider hierarchy constraints in the nonparametric setting via convex optimization schemes. To our knowledge, current techniques are unable to scale to the functional learning instances we consider in our paper.

As mentioned earlier, the EBM approach (\citealp{Nori2019}) does not support hierarchy constraints. 
GAMI-Net~\citep{Yang2021} follows a two-stage approach to fit nonparametric additive models with sparse (weak) hierarchical interactions. GAMI-Net performs a screening step after estimating the main effects with neural-network blocks and only consider interaction effects that satisfy the weak hierarchy principle amongst the screened main effect -- an interaction effect can appear in the model if one of the main effects is in the model. As mentioned earlier, GAMI-Net is not based on a sparsity-inducing penalized optimization procedure; and can be computationally much more expensive than our estimators.

\section{Simulations}\label{sec:simulation-case-study}

In this section we study the empirical performance (estimation and prediction) of the \modelbname~and \modelcname~estimators on synthetic datasets.

\subsection{Sparse additive model with interactions}
\label{sec:simulation-1}
Motivated by~\cite{li2}, we consider a problem with $p=10$ features where the true underlying model is additive in a small number of main and interaction effects: 
\begin{align}
    f^*(\M{x}) &= g_1\left(x_1\right) + g_2\left(x_2\right) + g_3\left(x_3\right) + g_4\left(x_4\right) \label{eq:synthetic}\\
    &+g_1\left(x_3x_4\right) + g_2\left(\frac{x_1+x_3}{2}\right) + g_3\left(x_1x_2\right), \nonumber
\end{align}
where functions $g_1(t)=t$, $g_2(t)=(2t-1)^2$, $g_3(t)=\frac{\text{sin}(2 \pi t)}{2-\text{sin}(2 \pi t)}$, and $g_4(t)=0.1\text{sin}(2 \pi t)+0.2\text{cos}(2 \pi t)+0.3\text{sin}^2(2 \pi t)+0.4\text{cos}^3(2 \pi t)+0.5\text{sin}^3(2 \pi t)$ are defined on $[0,1]$.
Note that the covariates  $x_5, \cdots, x_{10}$ do not contribute to the response. Each of the covariates $x_1,\ldots,x_{10}$ are independently drawn from the uniform distribution $\mathcal{U}(0,1)$.  
We generate the responses as $y=f^*(\M{x})+\epsilon$, where the errors $\epsilon$ are drawn from a Gaussian distribution $\mathcal{N}(0,0.2546^2)$.

We measure prediction accuracy using the integrated squared error, $\text{ISE} = \mathbb{E}_{\M{x}}[(\hat{f}(\M{x})- f^*(\M{x}))^2]$, estimated by Monte Carlo integration using $10,000$ test observations from the same distribution as the training ones, following the procedure in \cite{li2}. We vary the number of training observations from $100$ to $400$ and use  $100$ replications for each simulation setting.
We compare our estimators to well-known benchmarks EBM \citep{Nori2019}, GAMI-Net \citep{Yang2021}, COSSO \citep{lin2}, and MARS\footnote{MARS is a stepwise forward–backward procedure for building functional ANOVA models.}~\citep{Friedman1991}. 

Table \ref{tab:synthetic-i} presents the average test ISE and their standard errors (based on the replications). The tuning parameters for \modelbname, \modelcname, EBM and GAMI-Net are selected via  5-fold cross-validation. The results for MARS and COSSO are taken from Table 6 in \cite{li2}. Table~\ref{tab:synthetic-i} demonstrates that both of our estimators, \modelbname~and \modelcname, appear to work well compared to competitors.
Due to space constraints, additional simulations comparing our approach to EBM and with similar conclusions are provided in the Supplement \citep{supplement}.

\begin{wraptable}[22]{r}{0.5\textwidth}
\centering
    \caption{\textit{Integrated Squared Errors for \modelbname,  \modelcname, MARS, COSSO, EBM and GAMI-Net.}}
    \label{tab:synthetic-i}
    \resizebox{0.5\textwidth}{!}{
        \setlength{\tabcolsep}{10.0pt}
        \begin{tabular}{|l|c|c|c|}
        \hline
        \multicolumn{1}{|r|}{$\mathbf{N_{train}}$} &  &  &  \\ \cline{1-1}
        \textbf{Model} & \multirow{-2}{*}{\textbf{100}} & \multirow{-2}{*}{\textbf{200}} & \multirow{-2}{*}{\textbf{400}} \\ \hline
        MARS & \multicolumn{1}{l|}{$0.239\pm0.008$} & \multicolumn{1}{l|}{$0.109\pm0.003$} & \multicolumn{1}{l|}{$0.084\pm0.001$} \\
        COSSO & $0.378\pm0.005$ & $0.094\pm0.004$ & $0.043\pm 0.001$ \\
        EBM & $0.274\pm0.004$ & $0.170\pm0.002$ & $0.100\pm 0.001$ \\
        GAMI-Net & $0.281\pm0.007$ & $0.114\pm0.004$ & $0.063 \pm0.003$ \\
        \modelbname & $\mathbf{0.220}\pm0.013$ & $\mathbf{0.077}\pm0.002$ & $\mathbf{0.035}\pm0.001$ \\
        \modelcname & $\mathbf{0.180}\pm0.008$ & $\mathbf{0.081}\pm0.004$ & $\mathbf{0.038}\pm0.001$ \\ \hline
        \end{tabular}
    }
    \caption{\textit{Support recovery metric (F1-score) for the main effects and the interaction effects.}}
    \label{tab:synthetic-i-supp}
    \resizebox{0.5\textwidth}{!}{
        \setlength{\tabcolsep}{10.0pt}
        \begin{tabular}{|c|c|c|c|}
        \hline
        \multicolumn{1}{|r|}{$\mathbf{N_{train}}$} & \multicolumn{1}{l|}{\textbf{Model}} & \multicolumn{1}{c|}{\textbf{F1 (main)}} & \multicolumn{1}{c|}{\textbf{F1 (Interactions)}} \\ \hline
         & EBM & $57.14\pm0.00$ & $51.47\pm1.89$ \\
         & GAMI-Net & $61.77\pm0.79$ & $22.31\pm1.27$ \\
         & \modelbname & $85.35\pm1.24$ & $40.30\pm2.78$ \\
        \multirow{-3}{*}{100} & \modelcname & $\mathbf{93.46}\pm1.14$ & $\mathbf{66.61}\pm2.45$ \\ \hline
         & EBM & $57.14\pm0.00$ & $70.43\pm2.07$ \\
         & GAMI-Net & $59.23\pm0.41$ & $33.01\pm1.51$ \\
         & \modelbname & $90.55\pm1.02$ & $74.83\pm1.58$ \\
        \multirow{-3}{*}{200} & \modelcname & $\mathbf{98.52}\pm0.43$ & $\mathbf{82.43}\pm1.38$ \\ \hline
         & EBM & $57.14\pm0.00$ & $83.47\pm1.86$ \\
         & GAMI-Net & $58.11\pm0.30$ & $39.24\pm2.16$ \\
         & \modelbname & $97.43\pm0.62$ & $87.34\pm1.17$ \\
        \multirow{-3}{*}{400} & \modelcname & $\mathbf{99.31}\pm0.41$ & $\mathbf{90.17}\pm1.23$ \\ \hline
        \end{tabular}
    }
\end{wraptable}
Next, we study the support recovery performances of \modelbname, \modelcname, EBM and GAMI-Net. We evaluate models using the discrepancy between the true and estimated support and separately report the average F1-score (defined in the Supplement \citep{supplement}) for the main and the interaction effects. The support recovery metrics are shown in Table~\ref{tab:synthetic-i-supp}. We observe that, overall, both \modelbname~and \modelcname~appear to work better compared to EBM and GAMI-Net in F1-score. Moreover, \modelcname~is seen to be the best-performing method across all the simulation settings. In summary, our approaches work well in terms of variable selection.  

\begin{wraptable}[11]{r}{0.7\textwidth}
\caption{\textit{Relative frequency of the true main and interaction effects appearing in the \modelbname~ and \modelcname~ models.}}
\label{tab:synthetic-i-freq}
\resizebox{0.7\textwidth}{!}{
\setlength{\tabcolsep}{2.0pt}
\begin{tabular}{|l|cc|cc|cc|cc|}
\hline
\multicolumn{1}{|r|}{$\mathbf{N_{train}}$} & \multicolumn{2}{c|}{\textbf{100}} & \multicolumn{2}{c|}{\textbf{200}} & \multicolumn{2}{c|}{\textbf{400}} & \multicolumn{2}{c|}{\bf 1000} \\ \hline
Components & {\modelbname} & \modelcname & \modelbname & \modelcname & \modelbname & \modelcname & \modelbname & \modelcname \\ \hline
$x^{(1)}$ & 87\% & \textbf{97\%} & 79\% & \textbf{100\%} & 89\% & \textbf{100\%} & 90\% & \textbf{100\%} \\ \hline
$x^{(2)}$ & 67\% & \textbf{93\%} & 88\% & \textbf{100\%} & 99\% & \textbf{100\%} & 100\% & 100\% \\ \hline
$x^{(3)}$ & \textbf{100\%} & 95\% & \textbf{100\%} & 98\% & 100\% & 100\% & 100\% & 100\% \\ \hline
$x^{(4)}$ & 99\% & 99\% & \textbf{100\%} & 99\% & \textbf{100\%} & 99\% & 100\% & 100\% \\ \hline
{$[x^{(1)},x^{(2)}]$} & 69\% & \textbf{100\%} & 100\% & 100\% & 100\% & 100\% & 100\% & 100\% \\ \hline
{$[x^{(1)},x^{(3)}]$} & 18\% & \textbf{100\%} & 74\% & \textbf{94\%} & 97\% & \textbf{99\%} & 100\% & 100\% \\ \hline
{$[x^{(3)},x^{(4)}]$} & {0\%} &  \textbf{96\%} & 14\% & \textbf{33\%} & 42\% & \textbf{59\%} & \textbf{99\%} & 96\% \\ \hline
\end{tabular}}
\end{wraptable}
Table \ref{tab:synthetic-i-freq} presents additional variable selection details of the two best-performing approaches: \modelbname~and \modelcname. For each true main and interaction effect,  we report the relative frequency of their appearance in the estimated support across all replications.  The results in Tables~\ref{tab:synthetic-i-supp} and~\ref{tab:synthetic-i-freq}
suggest that \modelcname~has an edge over \modelbname~in terms of the variable selection when the true model is hierarchical. 
In practice, when we do not know if the underlying truth obeys a strong hierarchy, \modelcname~tends to result in more compact models, as we demonstrate in the Census application in Section \ref{sec:expts-case-study}. 

\subsection{Large-scale setting with correlated features}
\label{sec:simulation-2}
Next, we evaluate our proposed methods on large-scale synthetic data with $p=500$ correlated features.
We draw $(x_1,...,x_p)$ from a multivariate normal distribution $\mathcal{N}(0, \B \Sigma)$, where $\Sigma_{ij}=\sigma^{|i-j|}$ with $\sigma\in[0,1]$.
We generate the responses as $y=f^*(\M{x})+\epsilon$, where $\epsilon\sim\mathcal{N}(0,0.25)$ and
\begin{align}
\small
 f^*(\M{x}) &  = h_1\left(x_{26}\right) + h_2\left(x_{76}\right) + h_3\left(x_{126}\right) + h_4\left(x_{176}\right) + h_5\left(x_{226}\right) + h_6\left(x_{276}\right) 
\label{eq:large-synthetic}\nonumber \\
    & 
 + h_1\left(x_{326}\right) + h_2\left(x_{376}\right) + h_3\left(x_{426}\right) + h_4\left(x_{476}\right) 
\nonumber \\
    & 
 + h_1\left(x_{26}\right)h_2\left(x_{76}\right) +h_1\left(x_{26}\right)h_3\left(x_{126}\right)+ h_4\left(0.5(x_{126}+x_{176})\right) + h_4\left(x_{176}\right)h_5\left(x_{226}\right) 
\nonumber \\
    & 
 + h_4\left(x_{176}\right)h_6\left(x_{276}\right)  + h_5\left(x_{326}x_{376}\right) + h_6\left(x_{426}x_{476}\right) + h_4\left(x_{276}x_{476}\right), 
\nonumber
\end{align}
with $h_1(t)=0.5t$, $h_2(t)=1.25\text{sin}(t)$, $h_3(t)=0.3\text{exp}(t)$,  $h_4(t)=0.5t^2$, $h_5(t)=0.9\text{cos}(t)$, and $h_6(t)=1/(1+\text{exp}(-t))$.

\begin{wraptable}[17]{r}{0.6\textwidth}
\centering
\caption{\textit{Integrated Squared Errors and support recovery metrics for EBM, GAMI-Net, \modelbname~and \modelcname~on large-scale synthetic data with different correlation strengths ($\sigma$).}}
        \label{tab:synthetic-large-supp}
        \resizebox{0.6\textwidth}{!}{
        \setlength{\tabcolsep}{5.0pt}
        \begin{tabular}{|c|c|c|c|c|c|}
        \hline
        \multicolumn{1}{|r|}{} & \multicolumn{1}{l|}{} & \multicolumn{1}{c|}{\textbf{Prediction Error}} & \multicolumn{3}{c|}{\textbf{Support Recovery}} \\ \cline{3-6}
        \multicolumn{1}{|c|}{$\sigma$} & \multicolumn{1}{l|}{\textbf{Model}} & \multicolumn{1}{c|}{\textbf{ISE $(\times 10^{-2})$}} & \multicolumn{1}{c|}{\textbf{F1 (features)}} & \multicolumn{1}{c|}{\textbf{F1 (main)}} & \multicolumn{1}{c|}{\textbf{F1 (Interactions)}} \\ \hline
         \multirow{4}{*}{0.1} & \multicolumn{1}{l|}{EBM} & $379.1\pm5.3$ & $~~~~3.92\pm0.00$ & $~~3.92\pm0.00$ & $26.41\pm3.26$ \\
         & \multicolumn{1}{l|}{GAMI-Net} & $~~75.6\pm3.6$ & $~~20.84\pm3.34$ & $\textbf{99.11}\pm0.67$ &  $~6.66\pm1.30$ \\
         & \multicolumn{1}{l|}{\modelbname} & $~~~~8.9\pm0.9$ & $~~\textbf{98.95}\pm0.69$ & $67.75\pm4.77$ & $\textbf{95.17}\pm1.17$ \\ 
         & \multicolumn{1}{l|}{\modelcname} & $~~~~\textbf{7.4}\pm0.8$ & $~~98.72\pm0.59$ & $98.72\pm0.59$ & $93.78\pm1.65$ \\ \hline
         \multirow{4}{*}{0.3} & \multicolumn{1}{l|}{EBM} & $381.4\pm4.24$ & $~~~~3.92\pm0.00$ & $~~3.92\pm0.00$ & $28.26\pm3.07$ \\
         & \multicolumn{1}{l|}{GAMI-Net} & $~~80.6\pm5.3$ & $~~23.86\pm4.78$ & $\textbf{100.00}\pm0.36$ & $~9.65\pm2.73$ \\
         & \multicolumn{1}{l|}{\modelbname} & $~~11.4\pm1.6$ & $~~\textbf{99.24}\pm0.35$ & $65.90\pm5.07$ & $\textbf{95.10}\pm1.08$ \\ 
         & \multicolumn{1}{l|}{\modelcname} & $~~\textbf{10.0}\pm1.6$ & $~~98.16\pm1.80$ & $98.16\pm1.80$ & $93.48\pm2.41$ \\ \hline
         \multirow{4}{*}{0.5} & \multicolumn{1}{l|}{EBM} & $386.4\pm5.9$ & $~~~~3.92\pm0.00$ & $~~3.92\pm0.00$ & $26.00\pm3.40$ \\
         & \multicolumn{1}{l|}{GAMI-Net} & $~~73.8\pm3.3$ & $~~20.97\pm3.10$ & $\textbf{99.10}\pm0.56$ & $~7.87\pm1.96$ \\
         & \multicolumn{1}{l|}{\modelbname} & $~~12.4\pm1.7$ & $\textbf{100.00}\pm0.00$ & $60.04\pm5.75$ & $\textbf{95.25}\pm1.00$ \\ 
         & \multicolumn{1}{l|}{\modelcname} & $~~\textbf{10.6}\pm1.8$ & $~~98.87\pm0.54$ & $98.87\pm0.54$ & $94.32\pm1.05$ \\ \hline
         \multirow{4}{*}{0.7} & EBM & $375.7\pm3.3$ & $~~~~3.92\pm0.00$ & $~~3.92\pm0.00$ & $17.20\pm2.06$ \\
         & \multicolumn{1}{l|}{GAMI-Net} & $~~75.1\pm3.2$ & $~~20.04\pm3.18$ & $\textbf{99.28}\pm0.54$ & $~6.33\pm1.11$ \\
         & \multicolumn{1}{l|}{\modelbname} & $~~~~\textbf{9.3}\pm0.8$ & $~~\textbf{99.43}\pm0.31$ & $77.38\pm4.71$ & $\textbf{97.12}\pm0.98$ \\ 
         & \multicolumn{1}{l|}{\modelcname} & $~~10.6\pm1.8$ & $~~97.57\pm0.68$ & $97.57\pm0.68$ & $91.31\pm1.52$ \\ \hline
        \end{tabular}}
\end{wraptable}
We produce 10,000 training observations and evaluate the prediction performance as before, using ISE on a test set of size 10,000. 
We compare our estimators to EBM \citep{Nori2019} and GAMI-Net \citep{Yang2021}. 
Table \ref{tab:synthetic-large-supp} presents the average prediction errors and support recovery metrics over 25 simulation replications, along with the corresponding standard errors. The tuning procedure for \modelbname, \modelcname, EBM and GAMI-Net is described in the Supplement \citep{supplement}. 
Table~\ref{tab:synthetic-large-supp} demonstrates that our proposed models 
exhibit strong performance compared to competitors, showing a 7-11 fold reduction in the prediction error.

\section{Case study: Predicting the Census Survey Self-Response Rate}\label{sec:expts-case-study}
We now present our findings on the Census application. As mentioned in Section~\ref{sec:intro}, obtaining interpretable models with good predictive capabilities is important in this application -- such models can help inform the planning of outreach campaigns and guide stakeholders in deciding the allocation of spending for different communications channels (for example, TV, radio, digital), advertising messages with appropriately tailored content, and the timing of spending during the campaigns~\citep{2020ICC}. 
Due to their opaque nature, predictive models that 
topped the nationwide competition were not actionable for this application. On the other hand, the linear models that drive the LRS have limited predictive power. 
We explore how our proposed approach balances simplicity and good prediction performance.

The data we  use for this study is publicly available in the US Census Planning Database, which provides a range of demographic, socioeconomic, housing, and Census operational data \citep{Bureau2019}. The data in the Planning Database includes covariates from the 2010 Census and the 2013-2017 American Community Survey (ACS), aggregated at both the Census tract level and the block level.
We use tract-level data, with approximately $74,000$ observations and $500$ covariates. The response is the ACS self-response rate.  
We exclude the following covariates from our model: spatial covariates (``State'', ``County'', ``Tract'', ``Flag'', ``AIAN Land'') and variables that serve as a proxy to the response (for example, ``Low response score'', ``Number of housing units that returned first forms'', ``Replacement forms'' or ``Bilingual forms'' in Census 2010).
We also remove the margin of error variables corresponding to the ACS. After excluding these variables, we are left with $p=295$ covariates\footnote{To clarify, the results in this section are based on these $295$ features, though our algorithms scale to $p=500$.}.

\subsection{Experimental setup}

We randomly split the data into $58$K for training, $7.2$K for validation (used for selecting tuning parameters), and $7.2$K observations for testing. We repeat this procedure $20$ times with different random splits of the data and report the average numbers on the test sets.   
The features were standardized to have zero mean and unit variances. We use the squared $\ell_{2}$ loss for training and evaluate the performance of the models in terms of root mean square (RMSE). We also study the variables selected by our algorithm: for example, the number of features retained and the associated interpretations they offer. 

As noted in~\cite{Erdman2016}, the current approach for predicting the self-response rates--i.e., the LRS--is based on a linear regression model with around $25$ {\textit{hand-selected}} features. Erdman and Bates also use tailored variable transformations on some features to incorporate nonlinear effects. 
When the number of features, both main and pairwise interaction effects, increase, it is desirable to have an 
automated procedure such as the one we propose.  
As we use nonparametric models, which seek to learn nonlinearities in the main and interaction effects, manual feature engineering may not be necessary.

\noindent {\bf Benchmark Methods.} We evaluate and compare our models against existing linear and nonparametric approaches. 
We consider the following linear models with main effects only (i.e., without interactions):
{\bf (i)} Ridge regression; {\bf (ii)} Lasso regression~\citep{Hastie2015}; 
{\bf (iii)} $\ell_0$-penalized regression with additional ridge regularization (denoted as $\ell_0-\ell_2$), implemented via \texttt{L0Learn}~\citep{HazimehL0Learn}.
In addition, we consider the following linear models with interaction effects: 
{\bf (iv)}~Lasso with main effects and all pairwise linear interactions, 
{\bf (v)}~\texttt{hierScale}: a convex optimization framework for learning sparse main effects and interactions under a strong hierarchy constraint~\citep{HazimehHS}. 
For (i), (ii), and (iv), we use Python's \texttt{scikit-learn} library~\citep{Pedregosa2011}. 

In addition to the competing methods discussed above based on linear models, we consider nonparametric additive models with/without interactions. For AMs with main effects only, we compare with {\bf(vi)} Additive Models with Lasso-based selection, which we fit using Python's \texttt{scikit-learn} library~\citep{Pedregosa2011}. For additive models with interactions, we consider {\bf(vii)} EBM: A boosting approach that uses trees for main and interaction effects \citep{Nori2019}. {\bf(viii)} GAMI-Net: a neural network-based additive model with structured interactions \citep{Yang2021}. 
For EBM, we tune the learning rate in the range $[10^{-4}, 10^{-1}]$ and the number of interactions in $[10,500]$ for 1000 trials.
For GAMI-Net, we use a $200$ batch-size, $0.001$ learning rate, $500$ interaction effects, $500$ epochs (for each stage) and $0.001$ loss threshold; we set the NN basis to 2-layered network with 10-neurons.\footnote{The default NN basis of 5-layers with 40 neurons per-layer had long runtimes (3-days on a CPU), so a referee suggested the alternative hyperparameters. See the code in supplementary files \citep{supplement} for more details.}    

In addition to the linear and additive model benchmarks mentioned above, we also compare with state-of-the-art black-box machine learning methods such as {\bf (ix)} Gradient boosted decision trees (GBDT) from XGBoost \citep{Chen2016}; and {\bf (x)} Neural networks: multilayer perceptron (MLP).
The tuning parameters in both the models (ix) and (x) are selected using the Python hyperparameter optimization package \texttt{hyperopt} \citep{Bergstra2015}. GBDT is tuned with respect to the maximum depth $[1-10]$, number of estimators $[10-200]$, learning rate $[10^{-4},1]$. 
Neural networks are tuned with respect to the number of dense layers $\{2,3,4,5,6,7\}$, number of hidden units $\{64,128,256,512\}$, dropout rate $\{0.1,0.2,0.3\}$, learning rates $\{0.1,0.01,0.001,0.0001\}$ for Adam optimizer \citep{Kingma2015}, batch sizes $\{64,128\}$ and epochs $\{25,50,75,100\}$. The number of tuning parameters for all the nonparametric models is capped at $1000$.

\noindent {\bf Proposed Models.} 
Among our proposed estimators, we consider:
{\bf (a)}~\modelaname:~$\ell_0$-penalized AMs with nonlinear main effects. We also consider the following AMs with main and pairwise interaction effects: 
{\bf(b)}~\modelbname:~$\ell_0$-penalized AMs with main and interaction effects, i.e., estimator~\eqref{eq: GAM with interactions L0}; and 
{\bf(c)}~\modelcname:~sparse hierarchical interactions, i.e., estimator~\eqref{eq: GAM with interactions L0 with hierarchy MIP}.

All our algorithms for estimators (a)--(c) are implemented in Python. 
We use cubic B-splines with~$10$ knots for the main effects. For the interaction effects, we use tensor spline bases of degree~$3$ with 5 knots in each coordinate, leading to a total of $5 \times 5=25$ knots\footnote{We study the effect of number of knots on out-of-sample generalization on synthetic data (see the Supplement \citep{supplement}).}. Because the problem at hand has~$295$ covariates and $43,365$ possible pairwise interactions, we need to be careful with implementation aspects while generating spline-transformed representations for all the interaction effects, which can be memory intensive.

For \modelaname~and \modelbname~we perform a tuning procedure with warm-starts over a 2D grid of parameters ($\lambda_1$, $\lambda_2$) -- for details, see Section S1.4.3 of the Supplement \citep{supplement}. 
For both \modelbname~and \modelcname, we set $\alpha=1$ so that the main and the interaction effects have the same $\ell_0$-penalty parameter.
For \modelcname, we fix the smoothing parameter~$\lambda_1$ to the optimal value available from \modelbname.
Parameter~$\lambda_2$ (as well as parameter~$\tau$ defined in Section S3 of the Supplement \citep{supplement}) is chosen based on validation tuning. For all our methods (\modelaname, \modelbname, \modelcname), we cap the number of tuning-parameter values at 1000.

\subsection{Comparing methods: prediction, sparsity and model structure}\label{subsec:performance}

\begin{table}[!t]
   \caption{\textit{Comparisons of our methods with several benchmark models as discussed in the text. We display the average test RMSE for the different models, along with the corresponding number of covariates, and a number of effects (main and interactions).
   The best metrics are highlighted in bold. Numbers after $\pm$ provide the standard errors. Asterisk(*) indicates statistical significance (p-value<0.05) over the best existing model, using a one-sided paired t-test. Models (ix), (x) are black-box models using higher order interactions, hence \#Main and \#Interactions are left as `-' (dash).}}
   \label{tab:Out-of-sample MAE performance}
   \small 
   \centering 
    \resizebox{\textwidth}{!}{\begin{tabular}{|l|l|c|c|c|c|}
    \hline
    \textbf{Type} & \textbf{Model} & \textbf{RMSE}& \textbf{\#Covariates} & \textbf{\#Main} & \textbf{\#Interactions}\\ \hline
    \multirow{3}{*}{Linear Models (LM)} & (i) Ridge & $~6.750 \pm 0.013 $ & $295\pm~0$ & $295\pm~0$ & - \\ 
     & (ii) Lasso & $~6.741 \pm 0.013$ & $226\pm~7$ & $226\pm~7$ & - \\ 
     & (iii) L0Learn ($\ell_0-\ell_2$) & $~6.752 \pm 0.012$ & $145\pm~3$ & $145\pm~3$ & - \\ \hline
     \multirow{2}{*}{\begin{tabular}[c]{@{}l@{}} Linear Models \\ with Interactions (LMI)\end{tabular}} & (iv) LMI with Lasso & $~6.514\pm0.019$ & $262\pm~2$ & $~~75\pm~2$ & $1592\pm~~79$ \\ 
     & (v) LMI with Strong Hierarchy (hierScale)  & $~6.528 \pm 0.018$ & $276\pm~2$ & $276\pm~3$  & $4107\pm179$ \\ \hline
    \multirow{2}{*}{\begin{tabular}[c]{@{}l@{}}\\ 
    Additive Models (AM) \end{tabular}} & (vi) AM with Lasso & $~6.548 \pm 0.015$ & $285\pm~1$ & $285\pm~1$ & - \\ 
    & (a) 
    \modelaname
    & $~6.566 \pm 0.014$ & $184\pm11$ & $184\pm11$ & - \\ \hline
     \multirow{4}{*}{\begin{tabular}[c]{@{}l@{}} Additive Models \\ with Interactions (AMI)\end{tabular}} & (vii) EBM & $~6.475 \pm 0.014$ & $295\pm~0$ & $295\pm~0$ & $~492\pm~~~1$\\ 
     & (viii)  GAMI-Net & $~6.521 \pm 0.013$ & $232\pm~5$ & $202\pm~6$ & $~123\pm~~10$\\    
     & (b) 
     \modelbname
     & ${}^*\mathbf{6.442} \pm 0.019$ & $154\pm~8$ & $~~\textbf{33}\pm~5$ & $~201\pm~~19$ \\ 
     & (c) 
     \modelcname
     & ${}^*\mathbf{6.425} \pm 0.019$ & $\mathbf{133} \pm 6$ & $133 \pm 6$ & $~255\pm~13$\\ \hline
    \multirow{2}{*}{\begin{tabular}[c]{@{}l@{}}Nonparametric\\  (Non-interpretable)\end{tabular}} & (ix) GBDT& $~6.481 \pm 0.016$ & $278\pm~0$ & - & - \\ 
     & (x) Neural Networks (MLP) & $~6.505\pm0.016$ & $295\pm~0$ & - & - \\ \hline
    \end{tabular}}
\end{table}
We discuss the performance of different estimators in terms of prediction error and model parsimony. 

\noindent {\bf Additive models vs black-box ML methods.}
Table~\ref{tab:Out-of-sample MAE performance} reports the prediction errors (RMSE) on the test set along with the number of nonzero features in the model.
Importantly, we observe that the test performances of AMs with sparse interactions (both with and without the hierarchy constraints) are better than that of the best black box predictive ML models (ix), (x) that are based on GBDT, Neural Networks. Our model \modelcname~delivers the best RMSE value and is closely followed by \modelbname.

\noindent {\bf \modelbname/\modelcname~ vs state-of-the-art for AMs with sparse interactions.}
Among the methods we compared, EBM and GAMI-Net were the only methods that could compute AMs with sparse interactions for the Census dataset. Interestingly, the test performance of both \modelbname~and \modelcname~is better than that of EBM and GAMI-Net.
Interpretability (i.e., model sparsity and/or hierarchy) is another key factor differentiating the leading prediction methods.
Our models select a smaller number of additive components than EBM and GAMI-Net. For example, \modelbname~selects a total of 234 additive components. In contrast, EBM and GAMI-Net select 787 and 325 additive components, respectively.
We also observe that our models are almost twice as compact in terms of the number of selected covariates.

While Table~\ref{tab:Out-of-sample MAE performance} presents a summary of the best models chosen based on the validation set prediction performance, a practitioner may also find it useful to study a family of models prioritizing models with fewer features (perhaps at the cost of a marginal deterioration in predictive performance) -- see Section~\ref{subs:interpret} for further discussion.

We briefly discuss computation times for different methods. EBM takes around one hour to fit a model with 500 pairwise interactions (for a single tuning parameter). 
Computing a GAMI-Net solution (for one tuning parameter corresponding to 500 interaction effects with a 2-layered NN basis per effect) takes around $12$~hours\footnote{The default implementation with a 5-layered NN basis per effect took 3 days on a CPU.} using 8 CPUs and 3 hours using a single V100 Tesla GPU.
In contrast, our approaches are much faster: \modelbname~takes on average $1.8$~mins to compute a solution for a fixed tuning parameter\footnote{We use warm-start continuation across $\lambda_2$-values for a fixed $\lambda_1$: it takes approximately 3 hours to compute 100 solutions where the maximum number of pairwise interactions is set to 500.}. 
When \modelcname~is run on a reduced subset of $~800$ pairwise interactions, it takes $1.5$~hours to compute a path of 100 solutions.
These numbers are reported on an 8-CPU 128GB RAM device.

\noindent {\bf Nonlinear models vs linear models.} Table~\ref{tab:Out-of-sample MAE performance} suggests that
nonlinear/nonparametric models have better predictive performance than their linear counterparts. Also, linear models with sparse main and interaction effects appear to have an edge over sparse linear models without interactions. 
Another appealing aspect of sparse nonparametric AMs compared to linear models, is in model parsimony -- we alluded to this aspect in Figure~\ref{fig:Spy} while using a reduced number of features. 
The number of nonzero effects is significantly lower for the nonlinear AMs. For example, $\ell_0$-sparse nonparametric AMs with no interactions (i.e., \modelaname) can achieve the same level of predictive performance as its linear model counterpart (i.e., Model~(iii)) with significantly fewer covariates: $40$ versus $145$. Similarly, if we compare \texttt{hierScale} (i.e., linear main and interaction effects with strong hierarchical restrictions) to its nonparametric counterpart i.e., \modelcname, the number of main effects (and also, covariates) reduces from $276$ to $133$, and the number of interaction effects reduces from $4,107$ to $255$. Interestingly, all the covariates selected by the nonparametric AMs with interaction models (with/without strong hierarchy) are contained in the set of covariates selected by the sparse linear models with both main and interaction effects.

\noindent {\bf Additive nonlinear models: interactions vs no interactions.} Table~\ref{tab:Out-of-sample MAE performance} shows that ELAAN (AMs with no interactions) has more covariates than AMs with interactions, both \modelbname~and \modelcname. 
By including interactions that obey the hierarchy principle, we select a smaller number of covariates: \modelbname~has 154 covariates vs \modelcname~has 133.
The difference between \modelaname~vs \modelbname~is possibly because many nonlinear main effects attempt to explain the nonlinear interaction effects. This reduction points to the redundancy of some of the covariates when interaction effects are directly included in the model. 
The prediction performance of nonlinear AMs seems to improve quite a lot when pairwise interactions are included in the model.

\noindent \textbf{Hierarchy vs no hierarchy.}
We observe that \modelcname, i.e., the additive model with hierarchical interactions, achieves the best out-of-sample RMSE -- this improves over \modelbname~(interactions with no hierarchy).
The improvement is statistically significant based on a paired t-test. 
We observe that \modelbname~selects fewer effects (33 Main + 201 Interactions) when compared to \modelcname~(133 Main + 255 Interactions). 
On the other hand, \modelcname~obeys the hierarchy constraint and selects fewer covariates than \modelbname -- this can aid in interpretability and be easier to operationalize from a practical standpoint.

\noindent {\bf Insights from visualizations.}  As illustrated by Figure~\ref{fig:Nonlinear predictors}, which displays some marginal nonlinear fits, an appealing aspect of nonlinear AMs is that they naturally allow the practitioner to gather insights into the associations between the response and a feature of interest, for example, by exploring the map $x_{j} \mapsto f_{j}(x_{j})$. Similarly, the map $(x_{j}, x_{k}) \mapsto f_{j,k}(x_{j}, x_{k})$ would shed light into how the interaction of $(x_j,x_k)$ influences the output.
Such interpretations are a by-product of our additive model framework, and may not be readily available via black-box ML methods such as Neural Networks and GBDT---see also~\cite{Lou2013} for related discussions advocating for the interpretability of AMs. 
These association plots, for example, can help stakeholders identify promising factors influencing self-response scores and potentially informing policy decisions (e.g., targeted investments and optimizing operational costs).

\begin{figure}[!t]
        \centering
        \includegraphics[width=0.75\textwidth]{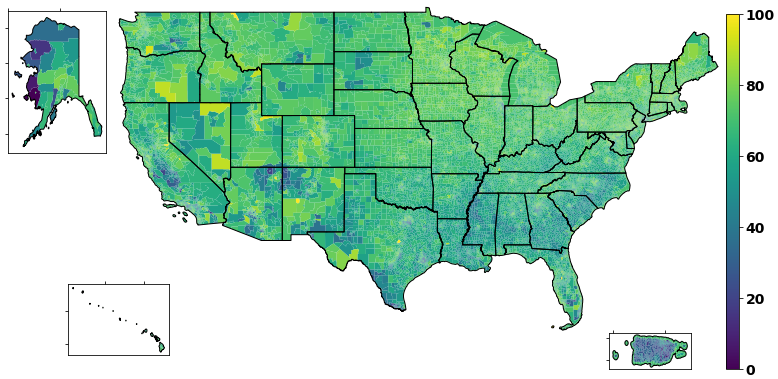}
        \caption{\textit{Predicted ACS self-response rates for all tracts in the United States.}}. 
        \label{fig:Maps-US-b-prediction}
\end{figure}
\begin{figure}[!t]
        \centering
        \includegraphics[width=0.75\textwidth]{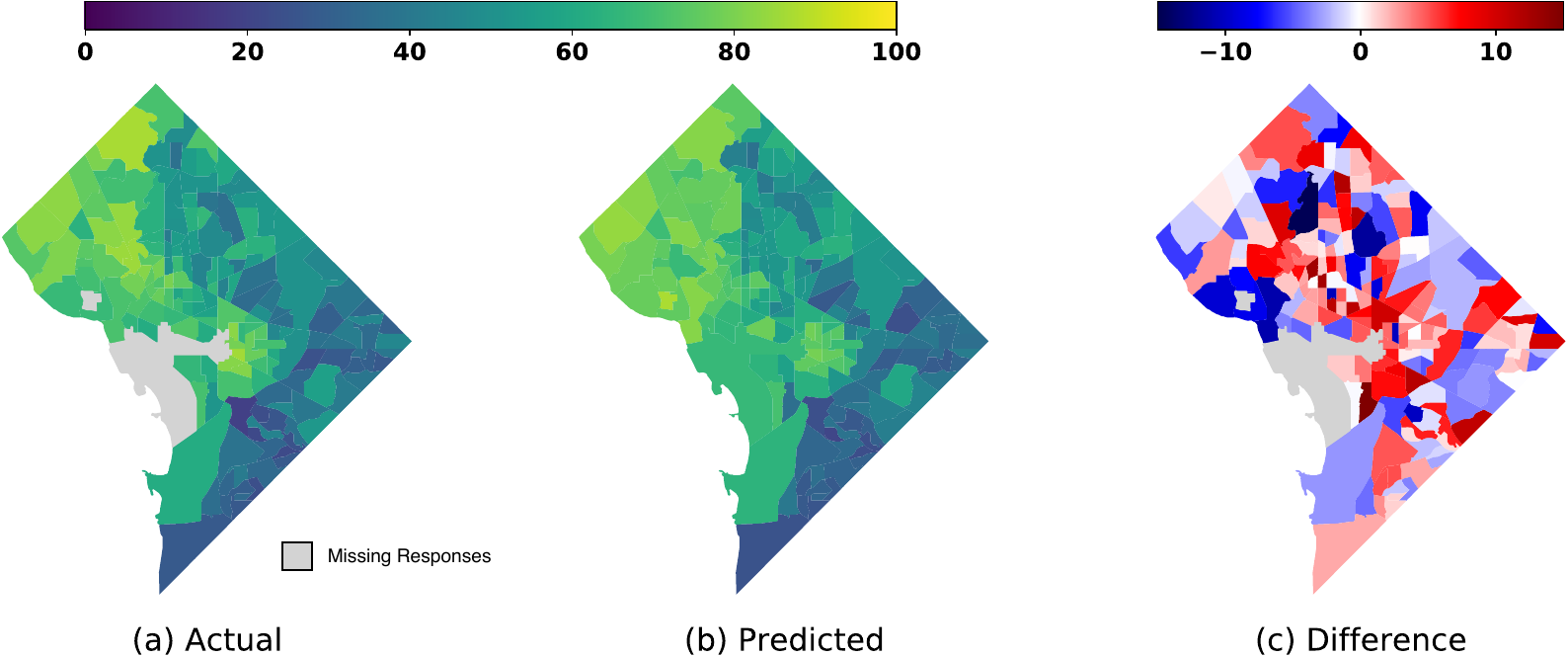}
        \caption{\textit{ACS self-response rates for all tracts in the District of Columbia. a) Actual ACS self-response rates. b) Predicted ACS self-response rates for AMs with interactions \eqref{eq: GAM with interactions L0}. c) Difference between the actual and predicted self-response rate: difference = actual - predicted.}}
        \label{fig:Maps-DC}
\end{figure}

To obtain a finer understanding of the performance of our models, we use visualization tools inspired by earlier works from the US Census Bureau~\citep{LRS-internal-2017,2020ICC}.
Figure \ref{fig:Maps-US-b-prediction} illustrates the tract self-response rates predicted by \modelbname~i.e.,
model~\eqref{eq: GAM with interactions L0} on a map of the United States. Different from Figure~\ref{fig:Maps-US-b-actual}, which shows actual data with missing values for some of the tracts\footnote{Some responses are deliberately suppressed by the US Census Bureau for privacy considerations, to limit the disclosure of information about individual respondents and to reduce the number of estimates with unacceptable levels of statistical reliability \citep{2016DS}}, Figure~\ref{fig:Maps-US-b-prediction} provides predicted self-response rates for all tracts based on our proposed model. 
Predictive models in general and our models in particular, allow us to forecast the response rates for tracts where gathered data is incomplete. 
It is worth noting that our regression-based models may be preferable over commonly used nearest neighbor type methods: the latter models may not readily offer insights into factors associated with high/low response rates. Nearest neighbor-based methods may also be ill-posed for isolated regions such as Alaska, Hawaii, and Puerto Rico, each having at least $10\%$ of their tracts censored.

\begin{wrapfigure}[23]{r}{0.5\textwidth}
  \begin{center}
    \includegraphics[width=0.5\textwidth]{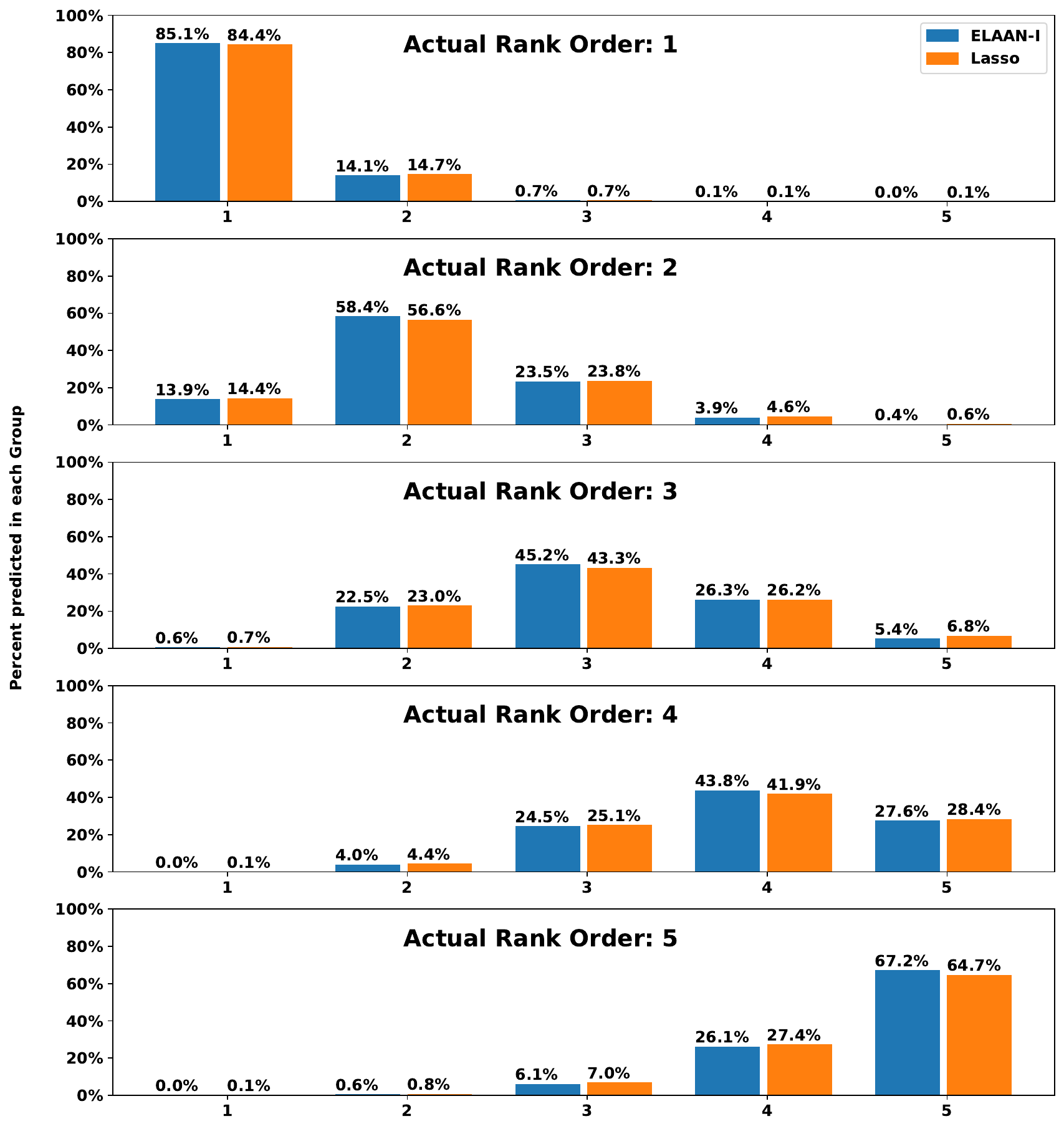}
    \caption{\textit{Comparison of quintile groups based on Lasso (linear model)~\citep{LRS-internal-2017,2020ICC} and our proposed~\modelbname. 
    }}.
    \label{fig:Maps-US-a}
  \end{center}
\end{wrapfigure}
Figure \ref{fig:Maps-DC} displays, side by side, the actual and the predicted self-response rates for the tracts in Washington DC, as well as the corresponding differences. Both the actual and the predicted rates are higher than average in most parts of the Northwest and a portion of the Northeast DC, and lower in the Southeast and most of the Northeast DC.

\looseness=-1 Figure \ref{fig:Maps-US-a} shows how the sorting of the Census tracts into quintiles~\citep{LRS-internal-2017,2020ICC} of the actual self-response rates compares with the corresponding sorting of the predictions made by: the Lasso (we use an $\ell_1$-penalized linear model with main effects and no interactions), and \modelbname.
For example, the top panel in Figure \ref{fig:Maps-US-a} shows that among the tracts in the first quintile of the actual self-response rates, $85.1\%$ are correctly predicted by \modelbname~to fall in the first quintile. The corresponding proportion for Lasso is~$84.4\%$. The same panel shows that among the tracts in the first quintile of the actual self-response rates, $14.1\%$ are incorrectly predicted to fall in the second quintile; this proportion is smaller than the one for the Lasso ($14.7\%$). Similarly, in the second quintile of the actual self-response rates, a higher proportion ($58.4\%$) is correctly identified by \modelbname~to be in the second quintile; this is an improvement over the~$56.6\%$ for the Lasso regression model. For all 5 quintiles of the actual self-response rates, \modelbname~identifies a higher proportion to be in the correct quintile than Lasso, suggesting that our approach has an edge over the Lasso.

\subsection{Interpreting important features}\label{subs:interpret}
\begin{figure}[!t]
       \begin{tabular}{cc}
        \includegraphics[width=0.475\textwidth,height=0.26\textheight]{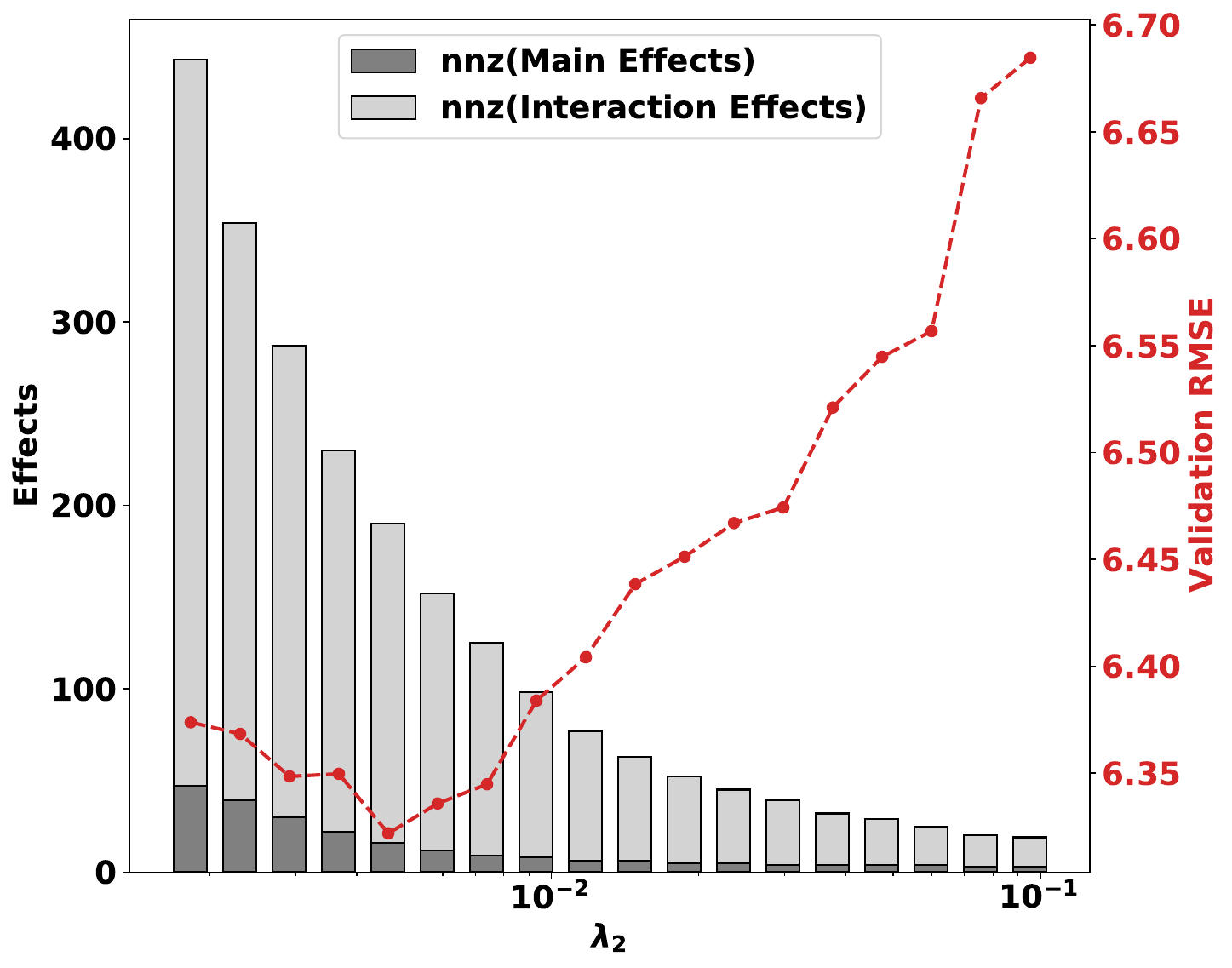}&
        \includegraphics[width=0.485\textwidth]{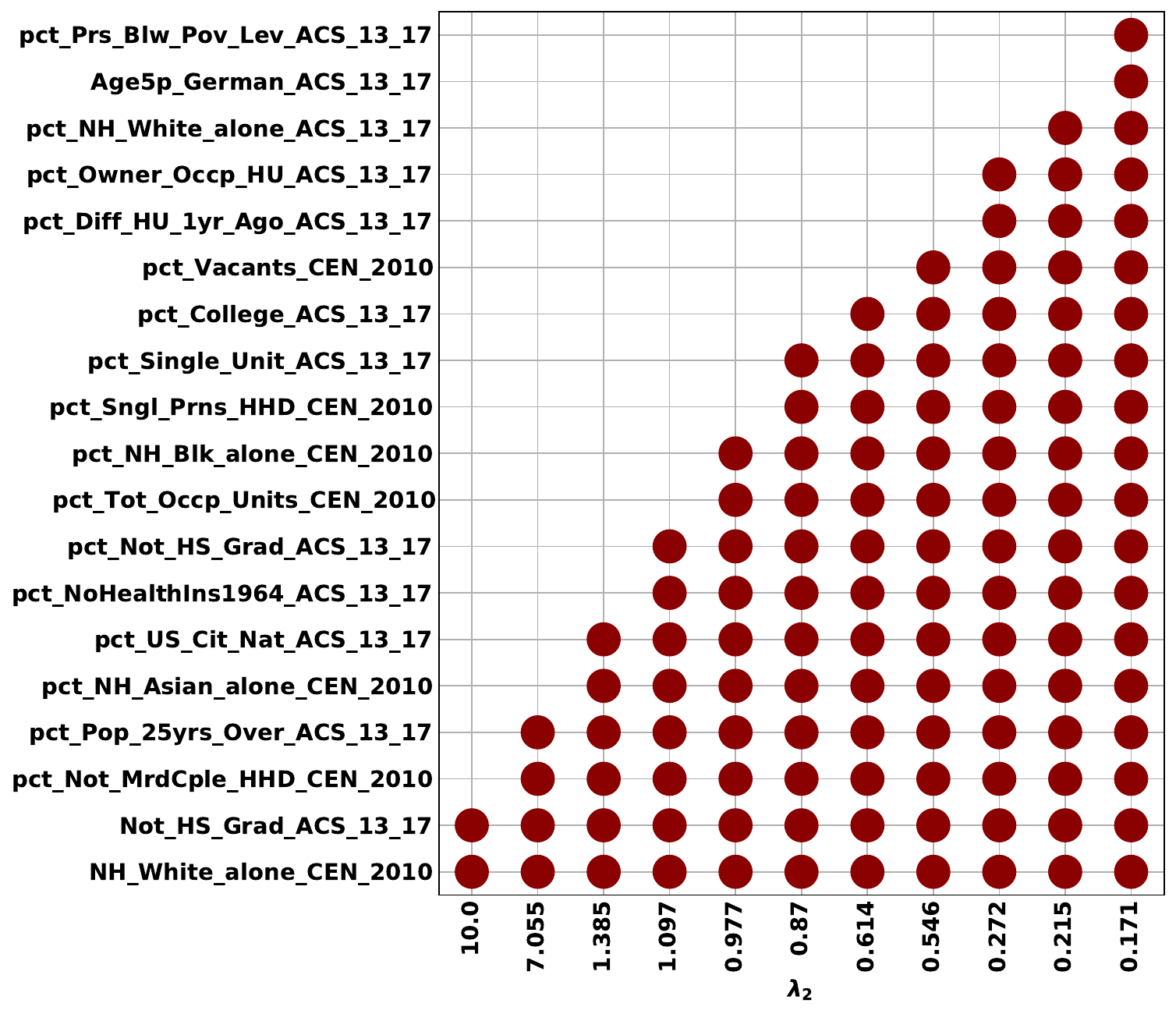}
        \end{tabular}
\caption{\textit{Estimates available from our model \modelbname~i.e., model~\eqref{eq: GAM with interactions L0}: $\ell_0$-sparse AMs with main effects and interactions.  [Left] number of nonzero effects vs the corresponding RMSE. [Right] Variables are corresponding to the main effects selected by the model along the regularization path. For visualization purposes, we focus on the top 19 main effects.}} \label{fig:GAMsWithInteractionsRegPath}
\end{figure}

We now illustrate how our methodology can guide the practitioner in obtaining a set of features and deriving associated actionable insights into the factors contributing to low response rates in surveys.
An important aspect of our regularized learning framework is that it provides an automated scheme to identify a collection of models, balancing out the complexity of the model and data-fidelity.
To see this, Figure \ref{fig:GAMsWithInteractionsRegPath}[left panel] plots the
number of main and interaction effects against the associated prediction error for model~\eqref{eq: GAM with interactions L0}. The plot shows that by trading off slight predictive performance (in terms of RMSE, which is shown by the red dashed line), we can limit the number of 
effects at any level, as desired by the practitioner who seeks a more parsimonious model.
Specifically, if we would like to limit the number of main effects to under~$20$, Figure~\ref{fig:GAMsWithInteractionsRegPath}(b) shows the top $19$ main effects in the order they enter the model along the regularization path. The definitions of the variables in Figure~\ref{fig:GAMsWithInteractionsRegPath}(b) are provided in the Supplement \citep{supplement}. Our investigation reveals that most (though not all) of these variables also appear in the models reported in prior studies by the Census Bureau -- see, for example ,~\cite {Erdman2016} and~\cite{2020ICC}. Interestingly, some features discovered by our framework can be interpreted in the context of clusters defined in~\cite{Bates2010} or mindset solutions in~\cite{2020ICC}, as discussed below. 

Using cluster analysis on the 12 covariates from the HTC score, \cite{Bates2010} categorize the tracts into eight distinct clusters: All Around Average I and II, Economically Disadvantaged I and II, Ethnic Enclave I and II, Single Unattached Mobiles, and Advantaged Homeowners. 
From the cluster analysis in \cite{Bates2010}, we see that the reported clusters have a wide variation in response rate (the response rate is not used to identify the clusters). 
Each cluster has some key characteristics (pertaining to a subset of variables) that seem to  drive the response rate.  
From the description of these clusters in \cite{Bates2010}, we see that the covariates automatically discovered by our framework, as listed in 
Figure~\ref{fig:GAMsWithInteractionsRegPath}[right panel], broadly cover the important characteristics upon which the clusters are based. 
For example, there is high variability across clusters with respect to the occupancy rate and home ownership status, which are captured by covariates ``Owner-occupied housing units'' and ``Total occupied units''.
Racial diversity also appears to play a role in cluster formation -- in particular, the covariates related to ethnic groups characterize the Ethnic Enclave clusters. 
Similarly, covariates related to poverty, for example, ``Persons below the poverty line'', lack of education (``No high school diploma''), and lack of health insurance (``No health insurance''), determine the Economically Disadvantaged clusters. 
Covariates related to high mobility, for example, ``Different housing unit in prior year'', high education (``College or higher degree holders''), marital status (``Single person households''), and nonspousal occupants (``Unmarried couples''), characterize the Single Unattached Mobiles cluster. 

The above discussion suggests that there may be important interaction effects within the variable groups that characterize each cluster. We observe that our AMs with interactions model \eqref{eq: GAM with interactions L0} 
automatically selects many of these effects. For instance, our model recovers several interaction effects between important covariates characterizing the Economically Disadvantaged I cluster. \cite{Bates2010} notes that this cluster has high percentages of people in poverty, people receiving public assistance, and adults without a high school education. Our model identifies an interaction effect between ``No high school graduation'' and ``No health insurance''. The same cluster also has a large African American population and above average number of children under 18. Our model potentially captures this characteristic by selecting an interaction effect between ``Non-Hispanic blacks'' and ``Population with ages 5-17''. We also recover an interaction effect between ``Non-Hispanic blacks'' and ``No high school graduation'', which seems to be reflective of this cluster. 
Consider also the Advantaged Homeowners cluster characterized by stable homeowners who are predominantly white and live in single-unit spousal-households. Our model automatically selects a closely related interaction effect between ``Non-Hispanic whites'' and ``Single units''. Another example is the Single Unattached Mobiles cluster, which is characterized by young, highly educated population living in multi-units structures with high mobility \citep{Bates2010}; the tracts are almost exclusively urban and have an above-average Asian population relative to the national average. Our model selects multiple interaction effects within this group of variables, for example: i) ``Moved in 2010 or later'' and ``Multi-unit structures''; ii) ``College or higher degree holders'' and ``Single person households''; iii) "People speaking Asian languages" and ``Multi-unit structures''; iv) ``People speaking Asian languages'' and ``Urban clusters''. 

Some of the covariates in Figure~\ref{fig:GAMsWithInteractionsRegPath}[right] can also be interpreted in terms of the mindset solutions in \cite{2020ICC}. In a mindset solution, survey respondents are categorized using factor analysis into 
six mindsets based on their predicted self-response patterns and demographic characteristics. These mindsets are determined by demographic characteristics such as age, race, income, home-ownership, presence of children in the household, marital status, internet use, English proficiency, and country of birth. The six mindsets are: Eager Engagers, Fence Sitters, Individuals with Confidentiality, Head Nodders, Wary Skeptics and Disconnected Doubters. The covariates can be also interpreted in terms of eight segmented tracts: Responsive Suburbia, Main Street Middle, Country Roads, Downtown Dynamic, Student and Military Communities, Sparse Spaces, Multicultural Mosaic, and Rural Delta and Urban Enclaves -- see \cite{2020ICC} for additional details.

\subsection{Conclusions}
We propose a framework to obtain models with good predictive accuracy
while simultaneously delivering parsimonious and interpretable statistical models. We hope that our framework will help practitioners identify some key factors that influence survey response rates. 
Insights gathered from our models may be used for targeted follow-up surveys and resource allocation for advertisements. We hope that our models will help inform the goal of achieving improved census coverage in a timely fashion while optimizing operational risks and costs. Our framework can complement and potentially offer an alternative to the LRS metric currently used in the US Census Bureau ROAM application.

\section*{Acknowledgements}
The authors would like to thank Nancy Bates, Tommy Wright and Joanna Fane Lineback from the U.S. Census Bureau for helpful discussions. Thanks also to other researchers from the Centers for Statistical Research and Methodology and Behavioral Science Methods (U.S Census Bureau) for helpful feedback in earlier versions of this work. Any opinions and conclusions expressed herein are those of the authors and do not necessarily represent the views of the U.S. Census Bureau. This research was supported in part, by grants from the Office of Naval Research: N000141812298, N000142112841 and
N000142212665 and the National Science Foundation: NSF-IIS-1718258 awarded to Rahul Mazumder. 

\begin{supplement}
\stitle{Additional Algorithmic Details, Proofs for Statistical Theory and Ablation Studies}
\sdescription{We present a detailed discussion of the univariate and bivariate smooth function estimation with splines, scalability considerations for optimizing \modelbname, and technical details and proofs for Theorem \ref{add.thm}. We also present an algorithm for optimizing \modelcname. Additional ablation studies and Census variable descriptions are also included.}
\end{supplement}
\begin{supplement}
\stitle{Data and Code}
\sdescription{We provide code and data for simulations and large-scale case study.}
\end{supplement}

\bibliographystyle{imsart-nameyear}
\bibliography{references}
\clearpage

\begin{center}
{\Large\bf{Supplementary Material: Predicting Census Survey Response Rates with Parsimonious
Additive Models and Structured Interactions}} \smallskip \\
by Shibal Ibrahim, Peter Radchenko, Emanuel Ben-David, and Rahul Mazumder\\~\\
\appendix
\section*{Supplementary Material}
\end{center}
This supplemental document contains some additional results not included in the main body of the paper. In particular, we present
\begin{enumerate}[-]
    \item a detailed discussion of the univariate and bivariate smooth function estimation with splines. These serve as building blocks for modelling main and interaction effects in Section \ref{sec:additive-nln-pwise-int}. The discussion is outlined in Section \ref{sec:computational-details-univariate-bivariate-estimation}.
    \item scalability considerations for optimizing \modelbname~i.e., optimization problem \eqref{eq: GAM with interactions L0} in Section \ref{sec: Parsimonious interactions}.
    \item statistical theory for \modelbname: technical details and proofs for Theorem \ref{add.thm} in Section \ref{sec:theory}.
    \item algorithm for optimizing \modelcname.
    \item additional ablation studies.
    \item Census variable descriptions for Figure \ref{fig:Spy} in Section \ref{sec: Parsimonious interactions} and Figure \ref{fig:GAMsWithInteractionsRegPath}b in Section \ref{subs:interpret}.
\end{enumerate}
This supplement is not entirely self-contained, so readers might need to refer to the main paper.

\etocdepthtag.toc{mtappendix}
\etocsettagdepth{mtchapter}{none}
\etocsettagdepth{mtappendix}{section}
\etocsettagdepth{mtappendix}{subsection}

\setcounter{page}{1}

\setcounter{table}{0}
\renewcommand{\thetable}{S\arabic{table}}%
\setcounter{figure}{0}
\renewcommand{\thefigure}{S\arabic{figure}}%
\setcounter{equation}{0}
\renewcommand{\theequation}{S\arabic{equation}}
\setcounter{section}{0}
\renewcommand{\thesection}{S\arabic{section}}%
\setcounter{footnote}{0}

\clearpage
\renewcommand\contentsname{} 
\tableofcontents

\section{Computational details for optimizing \modelbname}
\label{sec:computational-details-univariate-bivariate-estimation}

\subsection{Univariate and bivariate smooth function estimation with splines}
\label{sec-supp:univariate-and-bivariate-smooth-function-estimation-with-splines}
We discuss how the univariate and bivariate smooth functions are used to model main effects and interaction effects in Section \ref{sec:additive-nln-pwise-int}. The computational details for estimating these building blocks via quadratic optimization are outlined below.

\subsubsection{One dimensional nonparametric function estimation}

Suppose that we have observations $\{({y}_{i}, u_{i})\}_{1}^{n}$ corresponding to a univariate response ${y}_{i}$ and a univariate predictor $u_{i} \in [0,1]$. If the underlying relationship between~$y$ and~${u}$ can be modeled via a twice continuously differentiable (i.e., smooth) function $m(u)$, we can estimate $m(u)$ as a function that minimizes the following objective: $\sum_{i=1}^n (y_i - m (u_i))^2 + \lambda \int (m^{\prime \prime}(u))^2 du$.
This functional optimization problem can be reduced to a finite dimensional quadratic program by using a suitable basis representation for $m(u)$. For example, if all of the $u_{i}$'s are distinct and~$b_1(u),\dots,b_n(u)$ are the basis functions \cite[for example, cubic splines with knots at the observations $u_{i},$][]{Wahba1990,Hastie2015},
then we can write $m(u) = \sum_{i=1}^{n} \gamma_{i} b_{i}(u)$.
In this case, 
$\int (m^{\prime \prime}(u))^2 du = \B{\gamma}^\top\M{D} \B{\gamma}$, where $\M{D}$
is a positive semidefinite matrix with 
the $(i,l)$-th entry given by $\int  b^{''}_{i}(u)b^{''}_{l}(u) du$.

Instead of using an $n$-dimensional basis representation for $m(u)$, which is computationally expensive and perhaps statistically redundant,
one can 
choose a smaller number of basis functions, such as $K = O(n^{1/5})$; this leads to a low-rank representation of $\M{D}$~\citep{Hall2005,Ruppert2008} and hence an improved computational performance. 
Reducing the number of basis elements from $n$ to $K$ leads to fewer parameters at the expense of a marginal increase in bias~\citep{Wahba1990}, which is often negligible in practice.

The evaluations of the function $m(u)$ at the $n$ data points can be stacked together in the form of a vector $\M{B}\B\beta$, where $\M{B}$ is an $n\times K$ matrix of basis functions evaluated at the observations, and~$\B\beta \in \mathbb{R}^K$ contains the corresponding basis coefficients that need to be estimated from the data. The univariate function fitting problem then reduces to a regularized least squares problem given by  $\min_{\B{\beta}\in\mathbb{R}^K}  \{\norm{\B{y} - \B{B}\B{\beta}}_2^2 + \lambda \B{\beta}^T \B{A} \B{\beta}\}.$

While many choices of spline bases are available~\citep{Wood2006,Hastie2001}, we use penalized B-splines for their appealing statistical and computational properties.
Following~\cite{Eilers1996}, a second-order finite difference penalty on the coefficients of adjacent B-splines serves as a good discrete approximation to the integral of the squared second-order derivative penalty $\Omega$ discussed above. The regularized least squares problem takes the form
\begin{align}
    \min_{\B{\beta}}~~\norm{\B{y} - \B{B}\B{\beta}}_2^2 + \lambda \sum_{l=1}^{K-2}(\Delta^{2}\beta_l)^2,
\end{align}
where $K$ is the number of the B-spline basis functions and $\Delta^2 \beta_l = \beta_l - 2\beta_{l+1} + \beta_{l+2}$ is the second-order finite difference of the coefficients of adjacent B-splines with equidistant knots. We note that the regularization term can be represented in matrix form as $\lambda \norm{\B{D}\B{\beta}}_2^2$, where $\B{D} \in \mathbb{R}^{(K-2) \times K}$ is a banded matrix with nonzero entries given by $d_{l,l}=1$, $d_{l,l+1}=-2$ and $d_{l,l+2}=1$ for $l \in [K-2]$.

\subsubsection{Two dimensional nonparametric function estimation}
To model the response as a function of two covariates one can again use reduced rank parameterizations, in the form of multivariate splines \citep{Wahba1990}, thin-plate splines \citep{Kammann2003}, or tensor products of B-splines \citep{Eilers2003,  Wood2006}, for example. We use tensor products of B-splines to model the two-dimensional smooth functions of the form $m(u,v)$, with $(u,v) \in [0,1]^2$ for concreteness. 
We start from a low-rank B-spline basis representation for the marginal smooth functions as $m_1(u) = \sum_{k=1}^K \beta_k b_k(u)$ and $m_2(v) = \sum_{l=1}^L \delta_l c_l(v)$, where $\{b_k\}$ and~$\{c_l\}$ are B-spline basis functions, and $\{\beta_k\}, \{\delta_l\}$ are unknown basis coefficients. We convert the marginal smooth function~$m_1$ into a smooth function of covariates~$u$ and~$v$ by allowing the coefficients~$\beta_k$ to vary in a smooth fashion with respect to~$v$. Given that we have an available basis for representing smooth functions of~$v$, we can write $\beta_k(v)=\sum_{l=1}^L \delta_{k,l} c_l(v)$ and then arrive at the tensor smooth given by $m(u,v) = \sum_{k=1}^K \sum_{l=1}^L \delta_{k,l} b_k(u) c_l(v)$. We note that the evaluations of the function $m(u, v)$ at the $n$ data points can be stacked together in a vector written as~$\B{R}\B{\gamma}$. 
We denote the vectors evaluating the two marginal functional bases,  $\{b_k(\cdot)\}$ and $\{c_l(\cdot)\}$, at the $i$-th data point as $\B B_i \in \mathbb{R}^K$ and $\B C_i \in \mathbb{R}^L$, respectively. 
In matrix notation, the model matrix~$\B{R} \in \mathbb{R}^{n \times KL}$ can be expressed as $\B{R}=(\B{B}\otimes\B{1}_L)\odot(\B{1}_K\otimes\B{C})$, where operations~$\otimes$ and ~$\odot$ denote Kronecker product and element-wise multiplication respectively. The basis coefficients~$\delta_{k,l}$ are appropriately ordered into the vector~$\B{\gamma} \in \mathbb{R}^{KL}$ via the vectorization operation $  \B{\gamma}=\text{vec}(\B{\delta})=[\delta_{1,1}, \cdots, \delta_{K,1}, \delta_{1,2}, \cdots, \delta_{K,2}, \cdots, \delta_{1,L}, \cdots, \delta_{K,L}]^\top$, where  $\B{\delta} = [\delta_{k,l}]_{k \in [K], l \in [L]}$ stores the basis coefficients in matrix form.

The smooth 2D estimate can be obtained by making use of tensor products and a discretized version of the smoothness penalty~\citep{Eilers2003}, given by:
\begin{align}
    \min_{\B{\gamma}} \norm{\B{y} - \B{R}\B{\gamma}}_2^2 + \lambda_b \sum_{k=1}^K \sum_{l=1}^{L-2} (\Delta_{(1)}^2\delta_{k,l})^2 + \lambda_c \sum_{l=1}^L \sum_{k=1}^{K-2}  (\Delta_{(2)}^2\delta_{k,l})^2,
\end{align}
where $\Delta_{(1)}^2 \delta_{k,l} = \delta_{k,l} - 2 \delta_{k,l+1} + \delta_{k,l+2}$ and $\Delta_{(2)}^2 \delta_{k,l} = \delta_{k,l} - 2 \delta_{k+1,l} + \delta_{k+2,l}$. The regularization terms above can be compactly represented as 
quadratic forms in $\B\gamma$ as follows:
\begin{align}
\label{eq: two-dimensional PSR}
    \min_{\B{\gamma}} \norm{\B{y} - \B{R}\B{\gamma}}_2^2 + \lambda_b \B{\gamma}^T \B{P}_b \B{\gamma} + \lambda_c \B{\gamma}^T \B{P}_c \B{\gamma},
\end{align}
where $\B{P}_b = (\B{D}^T \B{D})\otimes \B{I}_L$ and $\B{P}_c=\B{I}_K \otimes (\B{D}^T \B{D})$. 
The regularizer ensures that the coefficients in the same row (or column) of $\B{\delta}$ vary regularly, leading to a smooth 2D surface.

\subsection{A finite dimensional quadratic program}
Problem~\eqref{eqn:base-GAM-with interactions} can be written as a finite-dimensional quadratic program. 
Using splines to model each of the main-effects and the interaction-effects, one can write the main-effects and interaction-effects as a linear combination of suitable bases elements. 
More specifically, $\M{f}_{j} = \M{B}_{j} \B{\beta}_{j}$, where $\M{B}_{j} \in \mathbb{R}^{n \times K_j}$ is the model matrix and $\B{\beta}_{j} \in \mathbb{R}^{K_{j}}$ is the vector of coefficients for each main effect component. Similarly, $\M{f}_{j,k} = \M{B}_{j,k} \B{\theta}_{j,k}$, where $\M{B}_{j,k} \in \mathbb{R}^{n \times K_{j,k}}$ is the model matrix and $\B{\theta}_{j,k} \in \mathbb{R}^{K_{j,k}}$ is the vector of coefficients for each interaction effect component.
Here, $K_{j}$ and $K_{j,k}$ denote the corresponding dimensions of the bases
-- in our implementation, all the values $K_{j}$ are taken to be the same and all the $K_{j,k}$ are taken to be the same as well.
We use a penalty function to control the smoothness (i.e., integral of the squared second derivative) of the 1D and 2D components. 
Writing~$\B{\beta}$ for the vector obtained by stacking together the coefficients $\B{\beta}_j$, $j\in[p]$ for the main-effects, and defining the vector~$\B{\theta}$ for the interaction effects analogously, we express the optimization problem \eqref{eqn:base-GAM-with interactions} as follows: 
\begin{equation*}
\small
   \min_{\B{\beta}, \B{\theta}} \; \frac{1}{n} \Big\|\B{y} - \big[\sum_{j \in [p]} \B{B}_j \B{\beta}_i + \sum_{j<k} \B{B}_{j,k} \B{\theta}_{j,k} \big]\Big\|_2^2 + \lambda_1 \big[\sum_{j \in [p]} \B{\beta}_j^T \B{S}_j \B{\beta}_j + \sum_{j<k} \B{\theta}_{j,k}^T \B{S}_{j,k} \B{\theta}_{j,k} \big].
\end{equation*}
Here, $\B{S}_{j} = \B{D}_j^T \B{D}_j$ and $\B{S}_{j,k} = (\B{D}_j^T \B{D}_j)\otimes \B{I}_k + \B{I}_j \otimes (\B{D}_k^T \B{D}_k)$ are the smoothness penalty matrices for the main effects and the interaction components, respectively. For convenience, we use the same smoothness penalty~$\lambda_{1}$ for both the main and the interaction effects, though in general they may be taken to be different.

\subsection{Block Coordinate Descent for solving \eqref{eq: GAM with interactions L0}}
\label{sec-supp:block-coordinate-descent-for-solving-am-with-interactions}
In our block CD method, the blocks correspond to the basis coefficients for either the main effects $\{\B\beta_{j}\}$ or the interaction effects $\{\B\theta_{j,k}\}$. Given an initialization 
$(\B \beta_1^{(0)}, \cdots, \B \beta_{p}^{(0)}, \B \theta_{1,2}^{(0)}, \cdots, \B \theta_{p-1,p}^{(0)})$, at every cycle, we sequentially sweep across the main effects and the interaction effects.  If we denote the solution after $t$ cycles by $(\B \beta_1^{(t)}, \cdots, \B \beta_{p}^{(t)}, \B \theta_{1,2}^{(t)}, \cdots, \B \theta_{p-1,p}^{(t)})$, then the block of coefficients for $j$-th main effect $\B \beta_{j}^{(t+1)}$ at the cycle $t+1$ is obtained by optimizing~\eqref{eq: GAM with interactions L0} with respect to~$\B \beta_{j}$, with other variables held fixed:
\begin{align}
\label{eq-supp:bcd-main}
    \B \beta_{j}^{(t+1)} \in \underset{\B{\beta}_j \in \mathbb{R}^{K_j}}{\argmin}~G(\B \beta_1^{(t+1)}, \cdots, \B \beta_{j-1}^{(t+1)}, \B \beta_j^{}, \B \beta_{j+1}^{(t)}, \cdots, \B \beta_{p}^{(t)}, ~~ \B \theta_{1,2}^{(t)}, \cdots, \B \theta_{p-1,p}^{(t)}).
\end{align}
Similarly, $\B \theta_{j,k}^{(t+1)}$, the coefficients for the $(j,k)$-th interaction effect at cycle $t+1$ are updated as
\begin{align}
\label{eq-supp:bcd-interaction}
    \B \theta_{j,k}^{(t+1)} \in \underset{\B{\theta}_{j,k} \in \mathbb{R}^{K_{j,k}}}{\argmin}~G(\B \beta_1^{(t+1)}, \cdots, \B \beta_{p}^{(t+1)}, ~~ \B \theta_{1,2}^{(t+1)}, \cdots, \B \theta_{(j,k)-1}^{(t+1)}, \B \theta_{j,k}^{}, \B \theta_{(j,k)+1}^{(t)} \cdots, \B \theta_{p-1,p}^{(t)}).
\end{align}
The block minimization problem~\eqref{eq-supp:bcd-main} reduces to:
\begin{align} \label{eq-supp: Subproblem main}
    \B{\beta}_j^{(t+1)} = \underset{\B{\beta}_j \in \mathbb{R}^{K_j}}{\argmin}  
    ~\psi_{j} (\B{r}^{(t)}; \B\beta_{j}):= \frac{1}{n} \big\|\B{r}^{(t)} - \B{B}_j \B{\beta}_j\big\|_2^2 + \lambda_1~\B{\beta}_j^T \B{S}_j \B{\beta}_j + \lambda_2 \mathbb{1} [\B{\beta}_j \neq \B 0 ] ,
\end{align}
where $\B{r}^{(t)} = \B{y}-(\sum_{j^{\prime}=1}^{j-1} \B{B}_{j^{\prime}} \B{\beta}_{j^{\prime}}^{(t+1)} + \sum_{j^{\prime}=j+1}^p \B{B}_{j^{\prime}} \B{\beta}_{j^{\prime}}^{(t)} + \sum_{j^{\prime}<k^{\prime}} \B{B}_{j^{\prime},k^{\prime}} \B{\theta}_{j^{\prime},k^{\prime}}^{(t)})$ 
denotes the residual. 
A solution to~\eqref{eq-supp: Subproblem main} can be computed in closed form via the following thresholding operator:
\begin{align}
    \label{eq-supp: Subproblem main solve}
    \B{\beta}_j^{(t+1)} = \begin{cases}
    \B{0} &\text{if}~~\psi_{j} (\B{r}^{(t)}; \B 0 ) \leq \min_{{\B\beta}_{j} \neq \B 0} \psi_{j} (\B{r}^{(t)}; \B\beta_{j}) \\
    \big(\B{B}_j^T \B{B}_j+ n\lambda_1 \B{S}_j\big)^{-1}\B{B}_j^T \B{r}^{(t)} &\text{otherwise}.
    \end{cases}
\end{align}
Similarly, the sub-problem for the interaction effects is
\begin{align} \label{eq-supp: Subproblem interaction}
    \B{\theta}_{j,k}^{(t+1)} = \underset{\B{\theta}_{j,k} \in \mathbb{R}^{K_{j,k}}}{\argmin}  ~\psi_{j,k} (\B{r}^{(t)}; \B\theta_{j,k}):= \frac{1}{n} \big\|\B{r}^{(t)} - \B{B}_{j,k} \B{\theta}_{j,k}\big\|_2^2 + \lambda_1~\B{\theta}_{j,k}^T \B{S}_{j,k} \B{\theta}_{j,k} +  \alpha \lambda_2 \mathbb{1} [\B{\theta}_{j,k} \neq \B 0 ],
\end{align}
where $\B{r}^{(t)} = \B{y}- (\sum_{j^{\prime}=1}^{p} \B{B}_{j^{\prime}} \B{\beta}_{j^{\prime}}^{(t+1)} + \sum_{(j^{\prime},k^{\prime})=1,2}^{(j,k)-1} \B{B}_{j^{\prime},k^{\prime}} \B{\theta}_{j^{\prime},k^{\prime}}^{(t+1)} + \sum_{(j^{\prime},k^{\prime}) = (j,k)+1}^{p-1,p} \B{B}_{j^{\prime},k^{\prime}} \B{\theta}_{j^{\prime},k^{\prime}}^{(t)})$.
A solution to this problem is given by 
\begin{align} \label{eq-supp: Subproblem interaction solve}
    \B{\theta}_{j,k}^{(t+1)} = \begin{cases}
 \B{0} &\text{if}\; ~~\psi_{j,k} (\B{r}^{(t)}; \B 0 ) \leq \min_{{\B\theta}_{j,k} \neq \B 0} \psi_{j,k} (\B{r}^{(t)}; \B\theta_{j,k}) \\
    \big(\B{B}_{j,k}^T \B{B}_{j,k}+ n\lambda_1 \B{S}_{j,k}\big)^{-1}\B{B}_{j,k}^T \B{r}^{(t)} &\text{otherwise}.
    \end{cases}
\end{align}
These block CD updates need to be paired with several computational devices in the form of active set updates, cached matrix factorizations,  and warm-starts, among others. We draw inspiration from similar strategies used in CD-based procedures for sparse linear  regression~\citep{HazimehL0Learn,Friedman2010}, and adapt them to our problem. 
These devices are discussed in detail in the Supplement Section \ref{sec-supp:scalability-considerations-for-solving-am-with-interactions}. 

\subsection{Scalability considerations for solving \eqref{eq: GAM with interactions L0}}
\label{sec-supp:scalability-considerations-for-solving-am-with-interactions}
We draw inspiration from strategies used in CD-based procedures for sparse linear regression~\citep{HazimehL0Learn,Friedman2010}, and adapt them to scale cyclic block coordinate descent in Section~\ref{sec: Efficient Computations at scale} to large problem instances of \eqref{eq: GAM with interactions L0} . These include active set updates, cached matrix factorizations and warm-starts. They are explained in more detail below:  

\subsubsection{Active set updates}
A main computational bottleneck in the block CD approach is the number of passes across the $O(p^2)$ blocks. However, as we anticipate a solution that is sparse (with few nonzero main and interaction effects), we use an active set strategy. We 
restrict our block CD procedure to a small subset~$\mathcal Q$ of the~$O(p^2)$ variables, with all blocks outside the active set being set to zero. Once the CD algorithm converges on the active set, we check if all blocks outside the active set satisfy the coordinate-wise optimality conditions\footnote{That is, we check if optimal solutions to~\eqref{eq:bcd-main} and~\eqref{eq:bcd-interaction} are zero for all blocks outside ${\mathcal Q}$.}. 
If there are any violations, we select the corresponding blocks, append them to the current active set, and then rerun our block CD procedure. As there are finitely many active sets, the algorithm is guaranteed to converge -- in practice, with warm-start continuation (discussed below), the number of active set updates is quite small, and the algorithm is found to converge quite quickly.

  \subsubsection{Cached matrix factorizations} The updates~\eqref{eq-supp: Subproblem main solve} and ~\eqref{eq-supp: Subproblem interaction solve} require computing a matrix inverse. In particular, if a block is nonzero, we need to compute a linear system solution of the form: $\B{A}_l^{-1}\B{b}_l$ where $\B{A}_l = \B{B}_l^T \B{B}_l+ n\lambda_1 \B{S}_l$.
We note that matrix  $\B{A}_l$ is fixed throughout the CD updates and is independent  of the choice of the sparsity regularization parameter $\lambda_2$ -- hence, we pre-compute a matrix factorization for $\B{A}_{l}$ (for example, an $LU$ decomposition) and use it to compute the solution to the linear system. 
As the dimension of $\B{A}_{l}$ equals the number of basis coefficients, which is small, this can be done quite efficiently.

  \subsubsection{Warm-starts} 
  \label{supp-sec:warm-starts}
  We use our CD procedure to compute a path of solutions to~\eqref{eq: GAM with interactions L0} for a 2D grid of tuning parameters $(\lambda_{1}, \lambda_{2}) \in \{\lambda_1^{(l)}\}_{l=0}^L \times \{\lambda_2^{(m)}\}_{m=0}^M$,  where $\lambda_{1}$ corresponds to the smoothness parameter and $\lambda_{2}$ the sparsity parameter. Here, 
  $\lambda_1^{(l)} > \lambda_1^{(l+1)}$ for all $l$, and  $\lambda_2^{(m)} > \lambda_2^{(m+1)}$ for all $m$.
  When $\lambda_{1}=\lambda_{1}^{(0)}$ (most regularized), we compute a sequence of solutions across the $\lambda_{2}$-values (from large to small values): 
 a solution obtained at $(\lambda_1^{(0)}, \lambda_2^{(m)})$ is used to initialize our CD procedure for the value $(\lambda_1^{(0)}, \lambda_2^{(m+1)})$. As the number of nonzeros in a solution to~\eqref{eq: GAM with interactions L0} generally increases with $\lambda_{2}$-values, our CD procedure for $(\lambda_1^{(0)}, \lambda_2^{(m+1)})$ uses an active set that is slightly larger than the active set\footnote{This is usually taken to be 1-10\% larger than the current active set, and is chosen in a greedy fashion from among the main and interaction effects lying outside the current active set.}
corresponding to the solution at $(\lambda_1^{(0)}, \lambda_2^{(m)})$.
Once we have traced a full path over~$\lambda_2$, we use warm-starts in the lateral direction across the space of $\lambda_1$. For all $l \geq 0$, to obtain a solution to~\eqref{eq: GAM with interactions L0} at $(\lambda_1^{(l+1)}, \lambda_2^{(m)})$, we use the solution at $(\lambda_1^{(l)}, \lambda_2^{(m)})$ as a warm-start.

\section{Statistical Theory for \modelbname: technical details and proofs}
\label{sec-supp:statistics-theory-proofs}
\label{sec:proofs}

\subsection{Notation and assumptions for Theorem~\ref{add.thm}}

Given a candidate regression function~$f:[0,1]^p\mapsto \mathbb{R}$ of the form $f(\M{x})=\sum_{j\in[p]} f_j(x_j)+\sum_{j<k} f_{j,k}(x_j,x_k)$, we define its sparsity level as
\begin{equation*}
G(f)=\sum_{j\in[p]} \mathbb{1} [ \bff_j  \neq \M{0}]+\sum_{j<k} \mathbb{1} [ \bff_{j,k}  \neq \M{0}].
\end{equation*}
To ensure identifiability of the additive representation for $f(\M{x})$ additional restrictions need to be imposed on the components~$f_j$ and~$f_{j,k}$. For the simplicity of the presentation, we avoid specifying a particular set of restrictions and treat every
representation of $f$ as equivalent, with the understanding that one particular representation is used when evaluating quantities such as $G(f)$.

We write $\|\cdot\|_{L_2}$ for the $L_2$ norm of a real-valued function on $[0,1]^2$ or $[0,1]$. We focus on the case where~$\mathcal{C}_1$ and~$\mathcal{C}_2$ are $L_2$-Sobolev spaces. More specifically, we let
\begin{equation*}
\mathcal{C}_2=\left\{g(u,v):[0,1]^2\mapsto \mathbb{R}, \;  \|g\|_{L_2}+\left\|\frac{\partial^2 g}{\partial u^2}\right\|_{L_2}+\left\|\frac{\partial^2 g}{\partial u\partial v}\right\|_{L_2}+\left\|\frac{\partial^2 g}{\partial v^2}\right\|_{L_2}<\infty\right\}
\end{equation*}
\begin{equation*}
\Omega(g)=\left\|\frac{\partial^2 g}{\partial u^2}\right\|_{L_2}^2+\left\|\frac{\partial^2 g}{\partial u\partial v}\right\|_{L_2}^2+\left\|\frac{\partial^2 g}{\partial v^2}\right\|^2_{L_2}, \smallskip
\end{equation*}
and also let ${\mathcal{C}_1}=\left\{h:[0,1]\mapsto \mathbb{R}, \;  \|h\|_{L_2}+\|h^{\prime\prime}\|_{L_2}<\infty\right\}$, \; ${\Omega}(h)=\|h^{\prime\prime}\|^2_{L_2}$. We define the corresponding space of additive functions with interactions as
\begin{equation*}
\mathcal{C}_\text{gr}=\{f:[0,1]^p\mapsto \mathbb{R}, \; f(\M{x})=\sum_{j\in[p]} f_j(x_j)+\sum_{j<k} f_{j,k}(x_j,x_k),\; f_j\in\mathcal{C}_1,\;f_{j,k}\in\mathcal{C}_2,\; G(f)\le K\},
\end{equation*}
where~$K$ is some arbitrarily large universal constant, and let
\begin{equation*}
\Omega_{\text{gr}}(f)=\sum_{j\in[p]} {\Omega}(f_j)+\sum_{j<k} \Omega(f_{j,k}).
\end{equation*}
We associate each $f\in \mathcal{C}_\text{gr}$ with the vector $\bff=\sum_{j\in[p]} \bff_j+\sum_{j<k} \bff_{j,k}$, where $\bff_j=\big(f_j(x_{1j}),...,\\ f_j(x_{nj})\big)$ and
$\bff_{j,k}=\big(f_{j,k}(x_{1j},x_{1k}),...,f_{j,k}(x_{nj},x_{nk})\big)$.

We assume that the observed data follows the model $\M{y}=\bff^*+\bepsilon$, where $\bff^*=\big(f(\M{x}_{1}),...,\\ f(\M{x}_{n})\big)$ is the vector representation of a function $f^*:[0,1]^p\mapsto \mathbb{R}$, and the elements of~$\bepsilon$ are independent $N(0,\sigma^2)$ with $\sigma>0$.

\subsection{Preliminaries and supplementary results}

We will prove a more general result by replacing the $2$-nd derivative in the definition of~$\mathcal{C}_\text{gr}$ and~$\Omega_\text{gr}$ with an $m$-th derivative, and redefining~$r_n$ as $n^{-m/(2m+2)}$. Theorem~\ref{add.thm} will then follow as a special case corresponding to $m=2$. The more general prediction error bound extends the result in Theorem~2 of \cite{lin2}, which is established for the fixed~$p$ setting, from the main effects models to the interaction models. When~$\log(p)\lesssim n^{1/(m+1)}$, the prediction error rate matches the optimal bivariate rate of~$n^{-m/(m+1)}$. 

We extend the domain of~$\|\cdot\|_n$ from vectors in~$\mathbb{R}^n$ to real-valued functions on~$[0,1]^p$ by letting $\|\cdot\|_n$ be the empirical $L_2$-norm.  Thus, given a function~$h$, we let $\|h\|_n=[\sum_{i=1}^n h(\M{x}_i)^2/n]^{1/2}$.  This extension is consistent in the sense that $\|f\|_n=\|\bff\|_n$ for $f\in\mathcal{C}_{\text{\rm gr}}$.
By analogy with the $\|\cdot\|_n$ notation, we define $(\bepsilon,\bv)_n=(1/n)\sum_{i=1}^n\epsilon_i v_i$ for each $\bv\in\mathbb{R}^n$.
For the simplicity of the presentation, we use the double-index notation $f_{jj}$ for the main effect components~$f_j$ of functions $f\in\mathcal{C}_\text{gr}$ and then take advantage of the additive representation
\begin{equation*}
\sum_{j\in[p]} f_j(x_j)+\sum_{j<k} f_{j,k}(x_j,x_k) = \sum_{j\le k} f_{j,k}(x_j,x_k).
\end{equation*}
We also define $\tilde{\Omega}(f_{j,k})=\sqrt{\Omega(f_{j,k})}$ and $\tilde{\Omega}_{\text{gr}}(f)=\sum_{j\le k} \sqrt{\Omega(f_{j,k})}$.

We let $\mathcal{J}=\{(j,k)\in [p]\times[p], \;\text{s.t.}\; j\le k \}$. Given $J\subseteq \mathcal{J}$, we define functional classes $\mathcal{F}(J)=\{f:\; f(\M{x})=\sum_{(j,k)\in J} f_{j,k}(x_j,x_k),\;f_{j,k}\in\mathcal{C}_2\}$ and~$\mathcal{H}({J})=\{h: \; h\in\mathcal{F}(J),\; \|h\|_n/(r_n+\tilde{\Omega}_{\text{\rm gr}}(h)\le 1\}$; we also let $\mathcal{F}_{s}=\{f: \; f\in\mathcal{C}_{\text{\rm gr}}, \; G(f)\le s\}$ for every natural~$s$. Given a positive constant~$\delta$ and a metric space~$\mathcal{H}$ endowed with the norm $\|\cdot\|$, we use the standard notation and write $H(\delta,\mathcal{H},\|\cdot\|)$ for the $\delta$-entropy of $\mathcal{H}$ with respect to~$\|\cdot\|$.  More specifically, $H(\delta,\mathcal{H},\|\cdot\|)$ is the natural logarithm of the smallest number of balls with radius~$\delta$ needed to cover~$\mathcal{H}$. The following result is proved in Section~\ref{prf.lem.entr.bnd}, bounds the entropy of  $\mathcal{H}({J})$.

\begin{lemma}\label{lem.entr.bnd}
$H(u,\mathcal{H}({J}),\|\cdot\|_n)\lesssim (1/u)^{2/m}$ for~$u\in(0,1)$.
\end{lemma}
Let $M=K(K+1)/2$. We will need the following maximal inequalities, which are proved in Sections~\ref{prf.lem.max.ineq} and~\ref{prf.lem.max.ineq.unif}, respectively.

\begin{lemma}\label{lem.max.ineq}
Suppose that $J\subseteq\mathcal{J}$ and $|J|\le M$. Then, with probability at least $1-e^{-t}$, inequality
\begin{equation*}
\label{hp.ineq.add}
(\bepsilon/\sigma,\bff)_n\lesssim \Big[ r_n + \sqrt{t/n} \Big] \|\bff\|_n + \Big[ r_n^2 + r_n\sqrt{t/n} \Big] \tilde{\Omega}_{\text{\rm gr}}(f)
\end{equation*}
holds uniformly over ${f\in\mathcal{F}(J)}$.  
\end{lemma}

\begin{lemma}
\label{lem.max.ineq.unif}
With probability at least $1-\epsilon$, inequality
\begin{eqnarray*}
(\bepsilon/\sigma,\bff)_n&\lesssim& \Big[r_n + \sqrt{\frac{\log(ep)}{n}} + \sqrt{\frac{\log(1/\epsilon)}{n}} \Big] \|\bff\|_n \\
&&+\Big[r_n^2 + r_n\sqrt{\frac{\log(ep)}{n}} + r_n\sqrt{\frac{\log(1/\epsilon)}{n}} \Big] \tilde{\Omega}_{\text{\rm gr}}(f)
\end{eqnarray*}
holds uniformly over ${f\in\mathcal{F}_{M}}$.
\end{lemma}

\subsection{Proof of Theorem~\ref{add.thm}}

Consider an arbitrary function $f\in \mathcal{C}_{\text{gr}}$. For the remainder of the proof, all universal constants will be chosen independently of~$f$. Because~$f$ is feasible for optimization problem~(\ref{crit.add1}), we have the following inequality:
\begin{equation}
\label{basic.ineq.fn}
\|\widehat \bff_n-\bff^*\|_n^2+\lambda_n  \Omega_{\text{gr}}(\widehat f_n) +\mu_nG(\widehat f_n)\le \|\bff-\bff^*\|_n^2+  2(\bepsilon,\widehat\bff_n - \bff)_n+\lambda_n  \Omega_{\text{gr}}(f)+\mu_nG(f).
\end{equation}

Applying Lemma~\ref{lem.max.ineq.unif} with $f=\widehat f_n-f$ and $\epsilon=(ep)^{-1}$, we derive that, with probability at least $1-1/p$,
\begin{eqnarray}
(\bepsilon/\sigma,\widehat\bff_n - \bff)_n&\le& {c}_1\Big[ r_n + \sqrt{\frac{\log(ep)}{n}} \Big] \|\widehat\bff_n - \bff\|_n \nonumber\\
\label{max.ineq.fhat} && + a_2\Big[ r_n^2 + r_n\sqrt{\frac{\log(ep)}{n}}  \Big] \tilde{\Omega}_{\text{\rm gr}}(\widehat f_n-f)
\end{eqnarray}
for some universal constants~$a_1$ and~$a_2$. For the remainder of the proof we restrict our attention to the random event on which~(\ref{max.ineq.fhat}) holds.  Multiplying inequality~(\ref{basic.ineq.fn}) by two, using~(\ref{max.ineq.fhat}), and letting
\begin{equation*}
\tau_n:=2{c}_1\sigma \Big[ r_n + \sqrt{\frac{\log(ep)}{n}} \Big], \;
\tilde{\lambda}_n:= 4a_2\sigma\Big[ r_n^2 + r_n\sqrt{\frac{\log(ep)}{n}}\Big], \; \text{and} \; {\lambda}_n\ge\tilde{\lambda}_n,
\end{equation*}
we derive
{\small 
\begin{equation*}
2\|\widehat \bff_n-\bff^*\|_n^2+2\lambda_n  \Omega_{\text{gr}}(\widehat f_n) +2\mu_nG(\widehat f_n) \le
2\|\bff - \bff^*\|_n^2+ 2\tau_n\|\widehat \bff_n - \bff\|_n+\tilde{\lambda}_n \tilde{\Omega}_{\text{gr}}(\widehat f_n-f)+2\lambda_n  \Omega_{\text{gr}}(f)+2\mu_nG(f).
\end{equation*}
}
Applying inequalities $2\tau_n\|\widehat \bff_n - \bff^*\|_n\le\tau_n^2+\|\widehat \bff_n - \bff^*\|_n^2$ and $2\tau_n\|\bff^* - \bff\|_n\le\tau_n^2+\|\bff - \bff^*\|_n^2$, we arrive at
\begin{eqnarray*}
2\|\widehat \bff_n-\bff^*\|_n^2+2\lambda_n  \Omega_{\text{gr}}(\widehat f_n) +2\mu_nG(\widehat f_n)&\le&\|\widehat \bff_n-\bff^*\|_n^2+3\|\bff - \bff^*\|_n^2 +2 \tau_n^2\\
\\
&& + \tilde{\lambda}_n\big[ \tilde{\Omega}_{\text{gr}}(\widehat f_n)+\tilde{\Omega}_{\text{gr}}(f)\big]+2\lambda_n  \Omega_{\text{gr}}(f)+2\mu_nG(f).
\end{eqnarray*}
Taking into account inequality $\tilde{\Omega}_{\text{gr}}(\tilde{f})\lesssim G(\tilde{f})+ \Omega_{\text{gr}}(\tilde{f})$, which holds for any $\tilde{f}\in \mathcal{C}_{\text{gr}}$, we conclude that
\begin{equation}
\label{prf.final.bnd}
\|\widehat \bff_n-\bff^*\|_n^2 +\lambda_n  \Omega_{\text{gr}}(\widehat f_n) +2\mu_nG(\widehat f_n) \le 3\|\bff - \bff^*\|_n^2+ a_3\tau_n^2 +2\lambda_n \Omega_{\text{gr}}(f)+2\mu_nG(f) 
\end{equation}
for some universal constant~$a_3$. \qed

\subsection{Proof of Corollary~\ref{add.cor}}
The stated prediction error bound is a direct consequence of the bound in Theorem~\ref{add.thm}, hence we only need to establish the sparsity bound. Taking $f=f^*$, we rewrite inequality~(\ref{prf.final.bnd}) as follows:
\begin{equation}
\label{prf.cor.bnd}
\|\widehat \bff_n-\bff^*\|_n^2 +\lambda_n  \Omega_{\text{gr}}(\widehat f_n) +2\mu_nG(\widehat f_n) \le a_3\tau_n^2 +2\lambda_n \Omega_{\text{gr}}(f^*)+2\mu_nG(f^*). \end{equation}
Provided that constant~$c_2$ in the lower bound for~$\mu_n$ is sufficiently large, we then have
$G(\widehat f_n)-G(f^*) < 1$, which implies $G(\widehat f_n)\le G(f^*)$.

We now focus on the classical asymptotic setting, where~$p$ is fixed and~$n$ tends to infinity, under the conditions of Corollary~\ref{add.cor}. Noting that $1/n=o(r_n^2)$ and repeating the arguments in the proofs of Theorem~\ref{add.thm} and Corollary~\ref{add.cor} with $\epsilon=\exp(-n r_n^2)$, we conclude that inequality~(\ref{prf.cor.bnd}) and the accompanying sparsity bound $G(\widehat f_n)\le G(f^*)$ hold with probability tending to one. Imposing an additional requirement that $\lambda_n=O\big(r_n^2+\log(ep)/n\big)$ and $\mu_n=O\big(r_n^2+\log(ep)/n\big)$ then leads to $\|\widehat \bff_n-\bff^*\|_n^2=O\big(r_n^2+\log(ep)/n\big)$. \qed

\subsection{Proof of Lemma~\ref{lem.entr.bnd}}
\label{prf.lem.entr.bnd}
We will establish the stated entropy bound for a somewhat larger functional space ${\mathcal{H}}_{J}'=\{h: \; h\in\mathcal{F}(J),\; \|h\|_{n}+\Tilde{\Omega}_{\text{\rm gr}}(h)\le 1\}$.  We treat~$m$ as a fixed universal constant in all inequalities that follow.

Consider an arbitrary~$g\in\mathcal{C}_2$. By the Sobolev embedding theorem \cite[for example,][Theorem 3.13]{oden1976reddy}, we can write $g$ as a sum of a polynomial of degree~$m-1$ and a function~$\tilde{g}$ that satisfies $\|\tilde{g}\|_{L_2}\lesssim \tilde{\Omega}(g)$, where we note that $\Tilde{\Omega}(g)=\Tilde{\Omega}(\tilde{g})$. Applying Lemma 10.9 in \cite{vdg2000applications}, which builds on the interpolation inequality of \cite{agmon1965lectures}, we derive $\|\tilde{g}\|_{\infty}\lesssim \Tilde{\Omega}(\tilde{g})$. Thus, ${\mathcal{H}}_{J}'\subseteq \{p+\tilde{h}: \; p\in\mathcal{P}_{J},\; \tilde{h}\in\tilde{\mathcal{H}}_{J}\}$, where
\begin{eqnarray*}
\mathcal{P}_{J}&=&\{p: \; p(\M{x})=\sum_{(j,k)\in J}\sum_{l=0}^{m-1}\sum_{q=0}^{m-1} \alpha_{j,k,l,q} x_{j}^l x_{k}^q, \; \alpha_{j,k,l,q}\in\mathbb{R},\;\|p\|_{n}\le 2 \}\\
\tilde{\mathcal{H}}_{J}&=&\{\tilde{h}:\; \tilde{h}\in \mathcal{F}(J), \; \tilde{\Omega}_{\text{\rm gr}}(\tilde{h})\le 1, \;  \|\tilde{h}_{j,k}\|_{\infty}\lesssim \Tilde{\Omega}(\tilde{h}_{j,k})\; \forall (j,k)\in J \}.
\end{eqnarray*}
Bound $\|p\|_{n}\le 2$ in the definition of $\mathcal{P}_{J}$ holds because if $h=p+\tilde{h}$ for $h\in {\mathcal{H}}_{J}'$ and $\tilde{h}\in\tilde{\mathcal{H}}_{J}$, then $\|p+\tilde{h}\|_n\le 1$ and $\|\tilde{h}\|_n\le \Tilde{\Omega}_{\text{\rm gr}}(\tilde{h})\le 1$. Consequently,
\begin{equation}
\label{entr.sum.bnd}
H(u,\mathcal{H}({J}),\|\cdot\|_{n})  \le H(u,\mathcal{H}_{J}',\|\cdot\|_{n})\le H(u/2,\mathcal{P}_{J},\|\cdot\|_{n}) + H(u/2,\tilde{\mathcal{H}}_{J},\|\cdot\|_{\infty}),
\end{equation}
where we used the fact that the unit ball with respect to the $\|\cdot\|_{\infty}$-norm is contained within the corresponding ball with respect to the $\|\cdot\|_{n}$-norm. We note that $\mathcal{P}_{J}$ is a ball of radis~$2$, with respect to the $\|\cdot\|_{n}$-norm, in a linear functional space of dimension $\lesssim|J|+1$. Hence, $H(u/2,\mathcal{P}_{J},\|\cdot\|_{n})\lesssim |J|+|J|\log(1/u)$ by, for example, Corollary 2.6 in \cite{vdg2000applications}.  Thus, the result of Lemma~\ref{lem.entr.bnd} follows from~\ref{entr.sum.bnd} if we also establish that $H(\delta,\tilde{\mathcal{H}}_{J},\|\cdot\|_{\infty})\lesssim |J|(1/\delta)^{2/m}$ for $\delta\in(0,1)$.

We complete the proof by deriving the stated bound on $H(\delta,\tilde{\mathcal{H}}_{J},\|\cdot\|_{\infty})$.  We represent~$\tilde{\mathcal{H}}_{J}$ as
{\small 
\begin{equation*}
\Big\{\tilde{h}:\; \tilde{h}(\M{x})=\sum_{{(j,k)}\in J} \lambda_{(j,k)} g_{(j,k)}(x_{(j,k)}), \, \sum_{{(j,k)}\in J} |\lambda_{(j,k)}| \le 1, \, g_{(j,k)}\in \mathcal{C}_2, \, \tilde{\Omega}(g_{(j,k)})\le 1, \, \|g_{(j,k)}\|_{\infty}\le 1 \Big\}.
\end{equation*}
}
Given functions $\tilde{h}(\M{x})=\sum_{{(j,k)}\in J} \lambda_{(j,k)} g_{(j,k)}(x_j,x_k))$ and $\tilde{h}'(\M{x})=\sum_{{(j,k)}\in J} \lambda_{(j,k)}' g_{(j,k)}'(x_j,x_k))$ in~$\tilde{\mathcal{H}}_{J}$, we have
{\small \begin{eqnarray*}
\|\tilde{h} - \tilde{h}'\|_{\infty} &\le&  \|\sum_{{(j,k)}\in J} \lambda_{(j,k)} g_{(j,k)} - \sum_{{(j,k)}\in J} \lambda_{(j,k)} g_{(j,k)}'\|_{\infty} + \|\sum_{{(j,k)}\in J} \lambda_{(j,k)} g_{(j,k)}' - \sum_{{(j,k)}\in J} \lambda_{(j,k)}' g_{(j,k)}'\|_{\infty} \\
&\le&  \max_{{(j,k)}\in J}\|g_{(j,k)} - g_{(j,k)}'\|_{\infty} \sum_{{(j,k)}\in J} |\lambda_{(j,k)}|  + \max_{{(j,k)}\in J}\|g_{(j,k)}'\|_{\infty} \sum_{{(j,k)}\in J} |\lambda_{(j,k)} - \lambda_{(j,k)}'| \\
&\le&   \max_{{(j,k)}\in J}\|g_{(j,k)} - g_{(j,k)}'\|_{\infty} + \sum_{{(j,k)}\in J} |\lambda_{(j,k)} - \lambda_{(j,k)}'|.
\end{eqnarray*}
}
Consequently, if we let $\mathcal{G}=\{g: \; g\in \mathcal{C}_2, \; \Tilde{\Omega}(g)\le 1, \; \|g\|_{\infty}\le 1\}$, let $\|\cdot\|_1$ denote the~$\ell_1$-norm, and let $B^{d}_1$ denote a unit $\ell_1$-ball in~$\mathbb{R}^d$, then
\begin{equation*}
H(\delta,\tilde{\mathcal{H}}_{J},\|\cdot\|_{\infty})\le |J|H(\delta/2,\mathcal{G},\|\cdot\|_{\infty}) + H(\delta/2,B^{|J|}_1,\|\cdot\|_1).
\end{equation*}
By the results in \cite{birman1}, $H(\delta/2,\mathcal{G},\|\cdot\|_{\infty})\lesssim (1/\delta)^{2/m}$.  By the standard bounds on the covering numbers of a norm ball, $H(\delta/2,B^{|J|}_1,\|\cdot\|_1)\lesssim|J|+|J|\log(1/\delta)$.  Consequently, $H(\delta,\tilde{\mathcal{H}}_{J},\|\cdot\|_{\infty})\lesssim |J|(1/\delta)^{2/m}$ for $\delta\in(0,1)$. \qed

\subsection{Proof of Lemma~\ref{lem.max.ineq}}
\label{prf.lem.max.ineq}

We note that $\|h\|_n\le r_n$ and $\Tilde{\Omega}_{\text{\rm gr}}(h)\le 1$ for every $h\in \mathcal{H}({J})$.  By Lemma~12 in the supplementary material for \cite{tan2019doubly} (cf. Corollary 8.3 in \citealp{vdg2000applications}),
\begin{equation}
\label{tail.bnd.entr}
\sup_{h\in\mathcal{H}({J})}(\bepsilon/\sigma,\bh)_n\lesssim  n^{-1/2}\int_0^{r_n}\sqrt{H(u,\mathcal{H}({J}),\|\cdot\|_n)}du + r_n\sqrt{t/n}
\end{equation}
with probability at least $1-e^{-t}$.

We note that $r_n=n^{-m/(2m+2)}$ and, thus, $n^{-1/2}r_n^{(m-1)/m}=r_n^2$. Using Lemma~\ref{lem.entr.bnd} to bound the entropy, we derive
\begin{eqnarray*}
n^{-1/2}\int_0^{r_n}\sqrt{H(u,\mathcal{H}({J}),\|\cdot\|_n)}du&\lesssim& n^{-1/2}\int_0^{r_n}u^{-1/m}du
\lesssim n^{-1/2}r_n^{(m-1)/m}= r_n^2.
\end{eqnarray*}
Applying bound~(\ref{tail.bnd.entr}), we conclude that
\begin{equation*}
\sup_{h\in\mathcal{H}({J})}(\bepsilon/\sigma,\bh)_n\lesssim  r_n^2 + r_n\sqrt{t/n}
\end{equation*}
with probability at least $1-e^{-t}$.  The statement of the lemma is then a consequence of the fact that for every $f\in\mathcal{F}({J})$, function
$f/\big[\|f\|_n r_n^{-1}+\Tilde{\Omega}_{\text{\rm gr}}(f)\big]$ falls in the class~$\mathcal{H}({J})$.
\qed

\subsection{Proof of Lemma~\ref{lem.max.ineq.unif}}
\label{prf.lem.max.ineq.unif}

Let~$M_s$ denote the number of distinct subsets of $\mathcal{J}$ that have size~$M$ or smaller. We note that $\log(M_s)\le 4M\log(ep)$ and, thus, $M_s e^{-t} \le e^{4M\log(ep)-t}$.  Applying Lemma~\ref{lem.max.ineq} together with the union bound, we derive that, with probability at least $1 - e^{4M\log(ep)-t}$, inequality
\begin{equation*}
(\bepsilon/\sigma,\bff)_n\lesssim \Big[ r_n + \sqrt{t/n} \Big] \|\bff\|_n + \Big[ r_n^2 + r_n\sqrt{t/n} \Big] \Tilde{\Omega}_{\text{\rm gr}}(f)
\end{equation*}
holds uniformly over ${f\in\mathcal{F}_{M}}$.   We complete the proof by noting that for $t=4M\log(ep)+\log(1/\epsilon)$ the above inequality becomes
\begin{eqnarray*}
(\bepsilon/\sigma,\bff)_n&\lesssim& \Big[ r_n + \sqrt{\frac{\log(ep)}{n}}+ \sqrt{\frac{\log(1/\epsilon)}{n}} \Big]\|\bff\|_n \\
&& + \Big[ r_n^2 + r_n\sqrt{\frac{\log(ep)}{n}}+ r_n\sqrt{\frac{\log(1/\epsilon)}{n}} \Big] \Tilde{\Omega}_{\text{\rm gr}}(f),
\end{eqnarray*}
and the corresponding lower-bound on the probability simplifies to $1-\epsilon$. \qed

\section{Algorithms for \modelcname~with sparse hierarchical interactions: Problem~\eqref{eq: GAM with interactions L0 with hierarchy MIP}}
\label{sec-supp:algorithms-for-sparse-hierarchical-interactions}
Similar to the case of~\eqref{eq: GAM with interactions L0},
and with the computational speed in mind, we present approximate methods to obtain good solutions to~\eqref{eq: GAM with interactions L0 with hierarchy MIP}.
As mentioned earlier, these approximate solutions can be used to initialize MIP-based approaches for~\eqref{eq: GAM with interactions L0 with hierarchy MIP} to improve the solution, and/or to certify the quality of these approximate solutions following~\citet{bertsimas2015best,HazimehL0Learn}.

Our first step is to reduce the number of main and interaction effects in~\eqref{eq: GAM with interactions L0 with hierarchy MIP} by making use
of the family of solutions available from~\eqref{eq: GAM with interactions L0}. We consider the union of supports available from the family of solutions obtained from~\eqref{eq: GAM with interactions L0}, across the 2D grid of tuning parameters $\lambda_1, \lambda_{2}$.
Let $\mathcal{M} \subset [p]$ and $\mathcal{I} \subset [p(p-1)/2]$ denote the sets of all nonzero main effects and interactions effects, respectively\footnote{If $\mathcal I$ includes an interaction $(j,k)$ where main-effect $j$ is not included in $\mathcal M$, we expand $\mathcal M$ to include $j$. This way, we make sure that all interaction effects have the corresponding main effects included in $\mathcal M$.}, encountered along the 2D path of solutions to~\eqref{eq: GAM with interactions L0}.
We form a reduced version of problem~\eqref{eq: GAM with interactions L0 with hierarchy MIP} with $z_{i} = 0, i \notin {\mathcal M}$ and $z_{j,k} =0$ for all $(j,k) \notin {\mathcal I}$. Let us denote the reduced problem by ${\mathcal P}(\mathcal M, \mathcal I)$. 
We consider a convex relaxation of ${\mathcal P}(\mathcal M, \mathcal I)$, denoted by ${\mathcal P}^{R}(\mathcal M, \mathcal I)$, where all binary variables $\{z_{i}\}$, $\{z_{j,k}\}$ in ${\mathcal P}(\mathcal M, \mathcal I)$ are relaxed to the interval $[0,1]$. As the sizes of $\mathcal M$ and $\mathcal I$ are generally small, it is computationally feasible to solve the relaxation ${\mathcal P}^{R}(\mathcal M, \mathcal I)$ -- we let $\{z^R_{i}\}$ and $\{z^R_{j,k}\}$ denote a solution to this relaxation. Following~\cite{HazimehHS}, it can be shown that this solution satisfies the strong hierarchy constraint (almost surely).
To obtain a feasible solution to Problem~\eqref{eq: GAM with interactions L0 with hierarchy MIP}, we apply a relax-and-round procedure. For the rounding step, we consider a threshold $\tau \in (0,1)$ and obtain $\tilde{z}_{i}= \mathbb{1}[z^R_{i} > \tau]$ for all $i \in \mathcal M$ and $\tilde{z}_{j,k} = \mathbb{1}[z^R_{j,k} > \tau]$ for all $(j,k) \in \mathcal I$. We set $\tilde{z}_{i}=0$, $i \notin {\mathcal M}$; and $\tilde{z}_{j,k}=0$ for all $(j,k) \notin {\mathcal I}$.
It can be verified that this rounding procedure maintains strong hierarchy. Finally, we perform a `polishing' step where we solve~\eqref{eqn:base-GAM-with interactions} restricted to the support defined by $\{\tilde{z}_{i}\}_{i}$ and $\{\tilde{z}_{j,k}\}_{j,k}$.

\noindent {\bf Related Work.} In contrast to Problem~\eqref{eq: GAM with interactions L0}, the regularization penalty in problem~\eqref{eq: GAM with interactions L0 with hierarchy MIP} is not separable across the blocks due to the overlapping groups created by the strong hierarchy constraint. Hence, the CD-based procedures discussed for~\eqref{eq: GAM with interactions L0} do not apply to the hierarchical setting. 
To the best of our knowledge, there are no prior specialized algorithms for~\eqref{eq: GAM with interactions L0 with hierarchy MIP} that apply to the scale that we consider here.
In fact, even in the linear regression setting, current algorithms for 
problems with a hierarchy constraint are somewhat limited in terms of the problem-scales they can address. The sole exceptions appear to be the convex optimization based approaches of~\cite{Lim2015,HazimehHS}, which can address sparse linear regression problems with a large number of features and a small number of observations.

\section{Simulations}

\subsection{Definition of F1-score}
\label{supp-sec:f1-definition}
F1-score is a harmonic mean of precision and recall:
\begin{align}
    \text{F1-score} = 2*\frac{\text{Precision} \times \text{Recall}}{\text{Precision} + \text{Recall}},
\end{align}
where 
\begin{align}
    \text{Precision} &= \frac{\#\text{ of True Positives}}{\#\text{ of True Positives} + \#\text{ of False Positives}}, \nonumber \\
    \text{Recall} &= \frac{\#\text{ of True Positives}}{\#\text{ of True Positives} + \#\text{ of False Negatives}} \nonumber
    \end{align}
For example, if we have 10 main effects and the true support is given by $\{1,1,1,1,0,0,0,0,0,0\}$ and the recovered support is $\{1,1,1,0,0,1,1,0,0,0\}$, the F1-score is $66.67\%$.

\begin{table}[!b]
\caption{Average integrated squared error for different configurations of basis elements. ``No'' in Smoothing column indicates $\lambda_1=0$. ``Yes'' in Smoothing column indicates $\lambda_1$ is tuned via cross-validation. Note that for all cases $\lambda_2$ is tuned via cross-validation.}
\label{tab:number-of-basis}
\resizebox{\textwidth}{!}{\begin{tabular}{|lccc|c|c|c|c|c|}
\hline
\multicolumn{4}{|c|}{} & \multicolumn{1}{r|}{$\mathbf{N_{train}}$} &  &  &  &  \\ \cline{1-5}
\multicolumn{1}{|l|}{\textbf{Model}} & \multicolumn{1}{c|}{$\mathbf{\alpha}$} & \multicolumn{1}{c|}{\textbf{Smoothing}} & $\mathbf{K_i}$ & $\mathbf{K_{ij}}$ & \multirow{-2}{*}{\textbf{100}} & \multirow{-2}{*}{\textbf{200}} & \multirow{-2}{*}{\textbf{400}} & \multirow{-2}{*}{\textbf{1000}} \\ \hline
\multicolumn{1}{|l|}{} & \multicolumn{1}{c|}{} & \multicolumn{1}{c|}{Yes} & { 20} & 8x8 & 0.269 (0.014) & 0.100 (0.004) & 0.0444 (0.0016) & 0.0161 (0.0005) \\
\multicolumn{1}{|l|}{} & \multicolumn{1}{c|}{} & \multicolumn{1}{c|}{Yes} & 10 & 5x5 & 0.193 (0.012) &  {0.075} (0.002) &  {0.0412} (0.0016) & {0.0135} (0.0004) \\
\multicolumn{1}{|l|}{} & \multicolumn{1}{c|}{} & \multicolumn{1}{c|}{Yes} & 5 & 3x3 &  {0.162} (0.005) & 0.088 (0.002) & 0.0579 (0.0007) & 0.0445 (0.0003) \\ \multicolumn{1}{|l|}{} & \multicolumn{1}{c|}{\multirow{-4}{*}{1.0}} & \multicolumn{1}{c|}{No} & 5 & 3x3 & 0.177 (0.005) & 0.100 (0.003) & 0.0609 (0.0008) & 0.0463 (0.0004) \\ \cline{2-9} 
\multicolumn{1}{|l|}{} & \multicolumn{1}{c|}{ } & \multicolumn{1}{c|}{Yes} & 20 & 8x8 & 0.270 (0.013) & 0.116 (0.006) & 0.0405 (0.0013) & 0.0157 (0.0004) \\
\multicolumn{1}{|l|}{} & \multicolumn{1}{c|}{ } & \multicolumn{1}{c|}{Yes} & 10 & 5x5 & 0.220 (0.013) &  {0.077} (0.002) &  {0.0353} (0.0011) & {0.0135} (0.0004) \\
\multicolumn{1}{|l|}{} & \multicolumn{1}{c|}{} & \multicolumn{1}{c|}{Yes} & 5 & 3x3 &  {0.174} (0.006) & 0.087 (0.002) & 0.0577 (0.0005) & 0.0448 (0.0003) \\ 
\multicolumn{1}{|l|}{\multirow{-8}{*}{\modelbname}} & \multicolumn{1}{c|}{\multirow{-4}{*}{{ 1.5}}} & \multicolumn{1}{c|}{No} & 5 & 3x3 & 0.188 (0.007) & 0.095 (0.002) & 0.0603 (0.0007) & 0.0455 (0.0003) \\ \hline
\end{tabular}}
\end{table}
\subsection{Ablation Studies}
\label{supp-section:additional-ablation-studies}
In this section, we perform ablation studies to investigate (i) the effect of number of basis elements on out-of-sample generalization (ii) the performance of our estimator when the error distribution of $\epsilon$ is not normally distributed.  
\subsubsection{Effect of number of basis functions} 
\label{supp-sec:knots}
We study the effect of number of basis functions on the model performance. We consider 3 different configurations: $\{(K_i=5, K_{ij}=3\times3), (K_i=10, K_{ij}=5\times5), (K_i=20, K_{ij}=8\times8)\}$. The results are reported in Table~\ref{tab:number-of-basis}. As expected, we observe that the performance generally degrades when the number of basis elements is too small (underfitting)  or too large (overfitting). 
When the number of bases elements is large, it is important to use smoothing for regularization.

\subsubsection{Skewed or heteroskedastic error distributions (Synthetic data)}
\label{supp-sec:skewed-dist}

We study how our model performs when the error distribution is (i) skewed, (ii) heteroskedastic. For (i),
we consider a log-normal distribution for the error: $\epsilon \sim \mathcal{LN}(-1,0.2546)$. For (ii), we consider a normal distribution for the error: $\epsilon \sim \mathcal{N}(0,0.5092 g_2(x_5))$, where the standard deviation is a function of the 5-th covariate in $\M{x}$. We compare average ISE for \modelbname~with EBM and GAMI-Net in these scenarios in Table \ref{tab:synthetic-i-nonnormal}. We can observe that \modelbname~consistently has favorable performance over both EBM and GAMI-Net even when the error distributions are not normally distributed. As expected, we observe a small drop in performance in both skewed and heteroskedastic settings compared to the normal distribution setting. We do not observe any degradation in support recovery metric in Table \ref{tab:synthetic-i-nonnormal-support-recovery} under skewed setting; however, we do observe a drop in support recovery F1-scores in the heteroskedastic setting. For example, the F1-scores for main and interaction effects for \modelbname~are $96.4\%$ and $81.7\%$ in the heteroskedastic setting for $N_{train}=400$ compared to $97.4\%$ and $87.3\%$ in the normal distribution setting.

\begin{table}[!t]
\caption{Integrated squared error for \modelbname, EBM and GAMI-Net when the error distribution is skewed or heteroskedastic.}
\label{tab:synthetic-i-nonnormal}
\begin{tabular}{|l|l|c|c|c|}
\hline
 & \multicolumn{1}{r|}{$\mathbf{N_{train}}$} & {} &  &  \\ \cline{1-2}
\textbf{$\epsilon$-distribution} & \textbf{Model} & \multirow{-2}{*}{$\mathbf{100}$} & \multirow{-2}{*}{$\mathbf{200}$} & \multirow{-2}{*}{$\mathbf{400}$} \\ \hline
 & EBM & $0.275\pm0.004$ & $0.171\pm0.002$ & $0.103\pm0.001$ \\
 & GAMI-Net & $0.292\pm0.007$ & $0.123\pm0.004$ &  $0.060\pm0.004$ \\
\multirow{-3}{*}{Skewed} & {\modelbname} & $\mathbf{0.214}\pm0.008$ & $\mathbf{0.084}\pm0.003$ & $\mathbf{0.037}\pm0.001$ \\ \hline
 & EBM & $0.284\pm0.004$ & $0.176\pm0.002$ & $0.105\pm0.001$ \\
 & GAMI-Net & $0.304\pm0.010$ &  $0.133\pm0.005$ &  $0.078\pm0.005$ \\
\multirow{-3}{*}{Heteroskedastic} & {\modelbname} & $\mathbf{0.215}\pm0.008$ & $\mathbf{0.087}\pm0.002$ & $\mathbf{0.042}\pm0.001$ \\ \hline
\end{tabular}
\end{table}

\begin{table}[!h]
\caption{Support recovery metrics for \modelbname, EBM and GAMI-Net when the error distribution is normal, skewed or heteroskedastic.}
\label{tab:synthetic-i-nonnormal-support-recovery}
\begin{tabular}{|l|c|l|c|c|}
\hline
\textbf{$\epsilon$-distribution} & $\mathbf{N_{train}}$ & \textbf{Model} & \textbf{F1(main)} & \textbf{F1(interactions)} \\ \hline
\multirow{9}{*}{Skewed} & \multirow{3}{*}{100} & EBM & $57.14\pm0.00$ & $\textbf{50.81}\pm1.86$ \\
 &  & GAMI-Net & $61.72\pm0.84$ & $21.39\pm1.32$ \\ 
 &  & ELAAN-I & $\mathbf{86.17}\pm1.11$ & $38.80\pm2.95$ \\ \cline{2-5} 
 & \multirow{3}{*}{200} & EBM & $57.14\pm0.00$ & $\mathbf{73.91}\pm2.12$ \\
 &  & GAMI-Net & $59.21\pm0.36$ & $31.84\pm1.52$ \\ 
 &  & ELAAN-I & $\mathbf{92.12}\pm0.83$ & $71.63\pm1.91$ \\ \cline{2-5} 
 & \multirow{3}{*}{400} & EBM & $57.14\pm0.00$ & $83.18\pm1.92$ \\
 &  & GAMI-Net & $57.63\pm0.20$ &  $43.33\pm2.57$ \\ 
 &  & ELAAN-I & $\mathbf{97.01}\pm0.60$ & $\mathbf{86.66}\pm 1.09$ \\ \hline
\multirow{9}{*}{Heteroskedastic} & \multirow{3}{*}{100} & EBM & $57.14\pm0.00$ & $\mathbf{49.96}\pm1.88$ \\
 &  & GAMI-Net & $61.55\pm1.40$ &  $19.71\pm2.03$ \\ 
 &  & ELAAN-I & $\mathbf{85.53}\pm1.11$ & $40.40\pm2.86$ \\ \cline{2-5} 
 & \multirow{3}{*}{200} & EBM & $57.14\pm0.00$ & $67.01\pm 2.09$ \\
 &  & GAMI-Net & $59.61\pm0.58$ &  $32.89\pm1.69$ \\ 
 &  & ELAAN-I & $\mathbf{90.72}\pm0.96$ & $\mathbf{71.71}\pm1.54$ \\ \cline{2-5} 
 & \multirow{3}{*}{400} & EBM & $57.14\pm0.00$ & $81.22\pm1.92$ \\
 &  & GAMI-Net & $58.42\pm0.54$ & $35.82\pm2.93$ \\ 
 &  & ELAAN-I & $\mathbf{96.41}\pm0.65$ & $\mathbf{81.75}\pm 1.20$ \\ \hline
\end{tabular}
\end{table}

\subsection{Tuning Details}
\subsubsection{Tuning Details for the Simulation in Section \ref{sec:simulation-1}}
We use 100 replications for each simulation setting, i.e., settings with different training set sizes, error distributions etc.
For each replication, we select the best hyperparameters via 5-fold cross-validation on the training set. We use mean squared error as the tuning criterion.
Next, we run the model on the full training set with the best hyperparameters to compute ISE performance on the test set.

For EBM and GAMI-Net, we tune the number of interactions in the range $[1,45]$ for 50 trials. For GAMI-Net, we used $32$ batch-size, $0.0001$ learning rate, $500$ epochs (for each stage) and $0.0$ loss threshold.

For \modelbname~and \modelcname, we used 10 knots for the main effects and 5 knots in each coordinate for the interaction effects (leading to $5\times5=25$ knots).
For \modelbname, we tuned $\lambda_1 \in \{10^{-6},\cdots,10^{-2}\}$ and $\lambda_2 \in \{10^{-5},\cdots,10^{-1}\}$ with warm-starts. 
For \modelcname, we tuned $\lambda_1 \in \{10^{-6},\cdots,10^{-2}\}$ and $\lambda_2 \in \{10^{-4},\cdots,1\}$, $\tau \in \{0.01,\cdots,1\}$ with warm-starts.

\subsubsection{Tuning Details for the Simulation in Section \ref{sec:simulation-2}}
We use 25 replications for each simulation setting.
For each replication, we select the best hyperparameters via validation tuning, using a validation set of size 2,000 with the mean squared error as the tuning criterion.
Next, we run the model on the full training set with the best hyperparameters to compute ISE performance on the test set.

For EBM, we tune the number of interactions in the range $[1,100]$.
For GAMI-Net, we set the number of interactions in to $500$. For GAMI-Net, we used $200$ batch-size, $0.001$ learning rate, $500$ epochs (for each stage). Loss threshold was set to $0.001$, which controls how many effects are selected after training and pruning. For GAMI-Net, we considered 2-layered NNs with 10 neurons per layer for each effect as suggested by the referee. The default basis with 5-layered NN with 40-neurons per layer was much slower.  

For \modelbname~and \modelcname, we used 10 knots for the main effects and 6 knots in each coordinate for the interaction effects (leading to $6\times6=36$ knots).
For \modelbname, we tuned $\lambda_1 \in \{10^{-7},\cdots,10^{-3}\}$ and $\lambda_2 \in \{10^{-4},\cdots,10^{0}\}$ with warm-starts.
The $\ell_0$ regularization path was terminated when the number of interactions reached $50$.
For \modelcname, we fix the smoothing parameter~$\lambda_1$ to the optimal value available from \modelbname~ and tuned $\lambda_2 \in \{10^{-2},\cdots,10^{2}\}$ and $\tau \in \{0.01,\cdots,1\}$ with warm-starts. For \modelbname~and \modelcname, we also considered three choices for $\alpha$ in the set $\{1.0, 1.5, 2.0\}$.

For \modelcname, our algorithm, outlined in \ref{sec-supp:algorithms-for-sparse-hierarchical-interactions}, approximately solves the problem under hierarchy with $\ell_0$ by solving the convex relaxation for computational reasons.
From variable section perspective, especially in the correlated setting considered in this simulation, the solution to the convex relaxation can overestimate the number of effects in the model e.g., the number of interactions. Hence, it can be beneficial to consider a solution  along the regularization path with smaller number of selected components, but lies within a standard error of the best solution. The numbers reported for prediction error and support recovery for \modelcname~in the Table \ref{tab:synthetic-large-supp} correspond to this choice of solution. This practice is common when using convex relaxations such as lasso in linear models.

\section{Data Analysis: Additional Details}

\subsection{Definition of important Census/American Community Survey variables}
\label{sec:app:B}

\subsubsection{Definition of the variables in Figure~\ref{fig:Spy}}
\label{sec-supp:definition-of-the-variables-in-Figure-spy}
\begin{itemize}
\item \texttt{Tot\_Population\_ACS\_13\_17}: U.S. resident population includes everyone who meets the ACS residence rules in the tract at the time of the ACS interview.

\item \texttt{pct\_Prs\_Blw\_Pov\_Lev\_ACS\_13\_17}: Percentage of the ACS eligible population that are classified as below the poverty level given their total family or household income within the last year.

\item \texttt{pct\_College\_ACS\_13\_17}: The percentage of the ACS population aged 25 years
and over that have a college degree or higher.

\item \texttt{pct\_Not\_HS\_Grad\_ACS\_13\_17}: The percentage of the ACS population aged 25 years
and over that are not high school graduates and have not received a diploma or the equivalent.

\item \texttt{pct\_Pop\_5\_17\_ACS\_13\_17}: The percentage of the ACS population that is between
5 and 17 years old.

\item \texttt{pct\_Pop\_18\_24\_ACS\_13\_17}: The percentage of the ACS population that is between
18 and 24 years old.

\item \texttt{pct\_Pop\_25\_44\_ACS\_13\_17}: The percentage of the ACS population that is between
25 and 44 years old.

\item \texttt{pct\_Pop\_45\_64\_ACS\_13\_17}: The percentage of the ACS population that is between
45 and 64 years old.

\item \texttt{pct\_Pop\_65plus\_ACS\_13\_17}: The percentage of the ACS population that is 65 years old or over.

\item \texttt{pct\_Hispanic\_ACS\_13\_17}: The percentage of the ACS population that identify as ``Mexican", ``Puerto Rican", ``Cuban", or ``another Hispanic, Latino, or Spanish origin".

\item \texttt{pct\_NH\_White\_alone\_ACS\_13\_17}: The percentage of the ACS population that indicate no
Hispanic origin and their only race as ``White" or report entries such as Irish, German, Italian, Lebanese, Arab, Moroccan, or Caucasian.

\item \texttt{pct\_NH\_Blk\_alone\_ACS\_13\_17}: The percentage of the ACS population that indicate no
Hispanic origin and their only race ``Black, African American, or Negro" or report entries such as African American, Kenyan, Nigerian, or Haitian.

\item \texttt{pct\_ENG\_VW\_ACS\_13\_17}: The percentage of all ACS occupied housing units where no one ages 14 years and over speaks English only or speaks English ``very well".

\item \texttt{pct\_Othr\_Lang\_ACS\_13\_17}: The percentage of the ACS population aged 5 years
and over that speaks a language other than English at
home.

\item \texttt{pct\_Diff\_HU\_1yr\_Ago\_ACS\_13\_17}: The percentage of the ACS population aged 1 year and over that moved from another residence in the U.S. or Puerto Rico within the last year.

\item \texttt{avg\_Tot\_Prns\_in\_HHD\_ACS\_13\_17}: The average number of persons per ACS occupied housing unit. This was calculated by dividing the total household population in the ACS by the total number of occupied housing units in the ACS.

\item \texttt{pct\_Sngl\_Prns\_HHD\_ACS\_13\_17}: The percentage of all ACS occupied housing units where a householder lives alone.

\item \texttt{pct\_Female\_No\_HB\_ACS\_13\_17}: The percentage of all ACS occupied housing units with
a female householder and no spouse of householder present.

\item \texttt{pct\_Rel\_Under\_6\_ACS\_13\_17}: The percentage of 2010 ACS family-occupied housing units with a related child under 6 years old; same-sex couple households with no relatives of the householder present are not included in the denominator.

\item \texttt{pct\_Vacant\_Units\_ACS\_13\_17}: The percentage of all ACS housing units where no one is living regularly at the time of interview; units occupied at the time of interview entirely by persons who are staying two months or less and who have a more permanent residence elsewhere are classified as vacant.

\item \texttt{pct\_Renter\_Occp\_HU\_ACS\_13\_17}: The percentage of ACS occupied housing units that are not owner occupied, whether they are rented or occupied without payment of rent.

\item \texttt{pct\_Owner\_Occp\_HU\_ACS\_13\_17}: The percentage of ACS occupied housing units with an
owner or co-owner living in it.

\item \texttt{pct\_Single\_Unit\_ACS\_13\_17}: The percentage of all ACS housing units that are in a structure that contains only that single unit.

\item \texttt{Med\_HHD\_Inc\_ACS\_13\_17}: Median ACS household income for the tract.

\item \texttt{Med\_House\_Value\_ACS\_13\_17}: Median of ACS respondents' house value estimates for the tract.

\item \texttt{pct\_HHD\_Moved\_in\_ACS\_13\_17}: The percentage of all ACS occupied housing units where the householder moved into the current unit in the year 2010 or later.

\item \texttt{pct\_NO\_PH\_SRVC\_ACS\_13\_17}: The percentage of ACS occupied housing units that do not have a working telephone and available service.

\item \texttt{pct\_HHD\_No\_Internet\_ACS\_13\_17}: Percentage of ACS households that have no Internet access.

\item \texttt{pct\_HHD\_w\_Broadband\_ACS\_13\_17}: Percentage of ACS households that have broadband Internet access.

\item \texttt{pct\_Pop\_w\_BroadComp\_ACS\_13\_17}: Percentage of people that live in households that have both broadband Internet access and a computing device of any kind in the ACS.

\item \texttt{pct\_URBANIZED\_AREA\_POP\_CEN\_2010}: The percentage of the 2010 Census total population that lives in a densely settled area containing 50,000 or more people.

\item \texttt{pct\_MrdCple\_HHD\_ACS\_13\_17}: The percentage of all ACS occupied housing units where the householder and his or her spouse are listed as members of the same household; does include same sex married couples.

\item \texttt{pct\_NonFamily\_HHD\_ACS\_13\_17}: The percentage of all ACS occupied housing units where a householder lives alone or with non relatives only; includes unmarried same-sex couples where no relatives of the householder are present.

\item \texttt{pct\_MLT\_U2\_9\_STRC\_ACS\_13\_17}: The percentage of all ACS housing units that are in a structure that contains two to nine housing units.

\item \texttt{pct\_MLT\_U10p\_ACS\_13\_17}: The percentage of all ACS housing units that are in a structure that contains 10 or more housing units.

\item \texttt{Civ\_labor\_16plus\_ACS\_13\_17}: Number of civilians ages 16 years and over at the time of the interview that are in the labor force in the ACS.

\item \texttt{pct\_Civ\_emp\_16plus\_ACS\_13\_17}: The percentage of ACS civilians ages 16 years and over in the labor force that are employed.

\item \texttt{pct\_One\_Health\_Ins\_ACS\_13\_17}: The percentage of the ACS population that have one type of health insurance coverage, including public or private.

\item \texttt{pct\_TwoPHealthIns\_ACS\_13\_17}: The percentage of the ACS population that have two or more types of health insurance.

\item \texttt{pct\_No\_Health\_Ins\_ACS\_13\_17}: The percentage of the ACS population that have no health insurance, public or private.

\end{itemize}
    
\subsubsection{Definition of the variables in Figure~\ref{fig:GAMsWithInteractionsRegPath}(b)}
\label{sec-supp:definition-of-the-variables-in-Figure-reg-path}

\begin{itemize}
\item \texttt{pct\_Prs\_Blw\_Pov\_Lev\_ACS\_13
\_17}: The percentage of the ACS eligible population that are classified as below the poverty level given their total family or household income within the last year, family size, and family composition.

\item \texttt{Age5p\_German\_ACS\_13\_17}: Number of people ages 5 years and over who speak English less than "very well" and speak German at home in the ACS. Examples include Luxembourgian.

\item \texttt{pct\_NH\_White\_alone\_ACS\_13\_
17}: The percentage of the ACS population that indicate no Hispanic origin and their only race as ``White" or report entries such as Irish, German, Italian, Lebanese, Arab, Moroccan, or Caucasian.

\item {pct\_Owner\_Occp\_HU\_ACS\_13
\_17}: The percentage of ACS occupied housing units with an owner or co-owner living in it.

\item \texttt{pct\_Diff\_HU\_1yr\_Ago\_ACS\_1
3\_17}: The percentage of the ACS population aged 1 year and over that moved from another residence in the U.S. or Puerto Rico within the last year.

\item \texttt{pct\_Vacant\_Units\_CEN\_2010}: The percentage of all 2010 Census housing units that have no regular occupants on Census Day; housing units with its usual occupants temporarily away (such as on vacation, a business trip, or in the hospital) are not considered vacant, but housing units temporarily occupied on Census Day by people who have a usual residence elsewhere are considered vacant. 

\item \texttt{pct\_College\_ACS\_13\_17}: The percentage of the ACS population aged 25 years and over that have a college degree or higher.

\item \texttt{pct\_Single\_Unit\_ACS\_13\_17}: The percentage of all ACS housing units that are in a structure that contains only that single unit.

\item \texttt{pct\_Sngl\_Prns\_HHD\_Cen\_2010}: The percentage of all 2010 Census occupied housing units where a householder lives alone.

\item \texttt{pct\_NH\_Blk\_alone\_CEN\_2010}: The percentage of the 2010 Census total population that indicate no Hispanic origin and their only race as ``Black, African American, or Negro" or report entries such as African American, Kenyan, Nigerian, or Haitian. 

\item \texttt{pct\_Tot\_Occp\_Units\_ACS\_13\_17}: The percentage of all ACS housing units that are classified as the usual place of residence of the individual or group living in it. 

\item \texttt{pct\_Not\_HS\_Grad\_ACS\_13\_17}: The percentage of the ACS population aged 25 years and over that are not high school graduates and have not received a diploma or the equivalent. 

\item \texttt{pct\_NoHealthIns1964\_ACS\_13\_17}: Percentage of people age 19 to 64 with no health insurance in the ACS.

\item \texttt{pct\_US\_Cit\_Nat\_ACS\_13\_17}: The percentage of the ACS population who are citizens of the United States through naturalization.

\item \texttt{pct\_NH\_Asian\_alone\_Cen\_2010}: The percentage of the 2010 Census total population that indicate no Hispanic origin and their only race as ``Asian Indian", ``Chinese", ``Filipino", ``Korean", ``Japanese", ``Vietnamese", or ``Other Asian".

\item \texttt{pct\_Pop\_25yrs\_Over\_ACS\_13\_17}: The percentage of the ACS population who are ages 25 years and over at time of interview.

\item \texttt{pct\_Not\_MrdCple\_HHD\_Cen\_2010}: The percentage of all 2010 Census occupied housing units where no spousal relationship is present.

\item \texttt{Not\_HS\_Grad\_ACS\_13\_17}: The percentage of the ACS population aged 25 years and over that are not high school graduates and have not received a diploma or the equivalent. 

\item \texttt{NH\_White\_alone\_CEN\_2010}: Number of people who indicate no Hispanic origin and their only race as ``White'' or report entries such as Irish, German, Italian, Lebanese, Arab, Moroccan, or Caucasian in the 2010 Census population. 
\end{itemize}

\end{document}